# Developing an ontology for the access to the contents of an archival fonds: the case of the Catasto Gregoriano.

LINA ANTONIETTA COPPOLA[1]

*ABSTRACT*: The research was proposed by adopting a computer-archival perspective, to exploit and extend the relational and contextual nature of the information assets of the Catasto Gregoriano, kept at the Archivio di Stato in Rome. Developed within the MODEUS project (Making Open Data Effectively Usable), this study originates from the following key ideas of MODEUS: to require Open Data to be expressed in terms of an ontology, and to include such an ontology as a documentation of the data themselves. Thus, Open Data are naturally "linked" by means of the ontology, which meets the requirements of the Linked Open Data vision. The aims are twofold: 1) to contribute to reuse data analyzed and extrapolated, making them machine readable and machine understandable; 2) to provide data access points combining both archival and content descriptions, in order to achieve higher quality and timely return of information. Combining the study of the Catatasto Gregoriano historical and political context, with the investigation of the information content of documents of the fonds, by abstracting and modeling, we came to the proposal of an explicit and formal conceptual model. Using an editor, called Eddy, for the development of ontologies expressed in the Graphol visual language, we realized an OWL ontology specifying an intensional conceptualization (TBox) of the domain of the Catasto Gregoriano. By fully exploiting the expressivty of OWL, the ontology captures the complexity of the domain through three main types of semantic relations: relations between the archival units, relations between elements of the domain described within the documents of the fonds, and, finally, the provenance relationships between those elements and the documents from which their extension can be derived. The development of the extensional level of the ontology (ABox) as well as the querying of the knowledge base obtained by merging the intensional and extensional levels of the ontology, show the potential of an approach that combines contextual information about the archival records with the information contained within such records, integrating - and enhancing – ontologies recently developed within the Cultural Heritage domain, and in particular, within the Archival domain.

## 1. Introduzione

L'uso e la crescita di interesse, in contesti e settori disciplinari differenti, delle tecnologie legate al Web Semantico, dalla pubblicazione di ontologie alla creazione di Linked Open Data, ha sollecitato e sollecita la ridefinizione di «modi e sistemi di valorizzazione del potere informativo espresso dalle descrizioni archivistiche»[2]. L'itinerario di esplorazione perseguito prende le mosse da un caso specifico, quello del Catasto gregoriano e dal parziale riordinamento e descrizione archivistica di uno dei fondi ad esso ascrivibili, *Catasti pontifici di Tivoli e suo territorio* (o *Cancelleria del censo di Tivoli*). Il fondo, in parte, riordinato e descritto con applicativo Xdams, scaturisce dall'iter attuativo e dall'apparato burocratico finalizzato all'esecuzione del primo catasto pontificio generale geometrico-particellare. Costituito di 428 registri collocabili nell'arco cronologico che intercorre tra il 1777 e il 1883, ha come soggetto produttore la Pontificia Cancelleria del censo o Cancelleria del censo di Tivoli, organo periferico dell'amministrazione del censo, cui era affidata, per la circoscrizione territoriale di competenza, la custodia e aggiornamento del catasto. Integrando i citati *corpora* documentari, già «ontologicamente individuati dal complesso di mutue relazioni che collegano i singoli documenti»[3], ai dati storico-politici - desunti dallo studio della normativa e dalla storia delle istuzioni - si è giunti, con l'ausilio di strumenti e editor di modeling, al secondo dei tre apporti previsti dal progetto: la concettualizzazione del dominio d'interesse. Terza e ultima proposta, il popolamento del livello estensionale della base di conoscenza e la generazione di interrogazioni, a titolo esemplificativo, che ne consentano l'accesso. Muovendo da siffatte premesse, non sembrerà inopportuna la puntuale disamina di quei risultati cui si è fatto cenno, elementi tangibili e al contempo sostrato ideale dell'indagine. Di questa, proponiamo qui di seguito, un estratto[4]. Si è scelto, infatti, di tralasciare il primo dei tre contributi - la schedatura e descrizione di 78 pezzi del fondo - e di concentrare l'attenzione sulla formalizzazione ontologica proposta.

[1] Università degli studi di Roma "La Sapienza". Dipartimento di Scienze documentarie, linguistico-filologiche e geografiche. Scuola di specializzazione in Beni archivistici e librari.

## 2. Il Catasto gregoriano: proposta di modello ontologico

La definizione di ontologia di Rudi Studer[5] - «An ontology is a formal, explicit specification of shared conceptualization» - se da un lato, riprende e condensa quanto visto finora, dall'altro fornisce l'appiglio, il gancio per esporre uno dei punti nodali, probabilmente, il punto nodale della presente ricerca: la proposta di uno schema concettuale espressivo, formale relativo ad un'area della conoscenza, segnatamente quella dell'impianto e del funzionamento del catasto Pio-Gregoriano, meglio, delle istituzioni coinvolte e della documentazione prodotta a tale scopo. Per consentire una adeguata interpretazione dei concetti e delle loro proprietà, occorre che la concettualizzazione - vale a dire l'astrazione del settore di riferimento effettuata - sia esplicitata nell'ontologia, facendo ricorso ad un livello intensionale e ad uno estensionale. «Nel primo caso, la concettualizzazione […] considera concetti e relazioni astratti, indipendentemente dagli specifici stati di riferimento e dal mutare delle relazioni nel tempo; nel secondo caso, […] mette in evidenza relazioni che descrivono un preciso "universo di discorso", dipendenti da uno specifico stato del mondo, un preciso insieme di riferimento».[6] Al livello intensionale - alle relazioni semantiche tra classi e proprietà e alla specifica degli elementi del dominio - viene dedicato, il presente capitolo, a quello estensionale - all' appartenenza di istanze a classi/proprietà descritte nel livello intensionale - il successivo. Piuttosto che proporre una sistematica elencazione degli assiomi OWL realizzati (vd. Appendice) si è ritenuta strategia migliore, tradurli in linguaggio naturale e grafico (Graphol)[7] scomponendo il diagramma dell'ontologia, prodotto con Eddy[8], e commentandone le singole porzioni. Le classi vengono segnalate, nell' esegesi, col corsivo. Con l'ausilio del colore si sono indicate, invece, quelle istanze e quelle classi - e di concerto, le loro data properties e object properties, sebbene lasciate in bianco - che trovano essenza e ragion d'essere solo nel contesto d' interesse: il catasto Pio-Gregoriano e, ancor meglio, nel modo in cui la sua documentazione e storia risultano sedimentati e organizzati all'interno dell'Archivio di Stato di Roma.

---

[2] F. Tomasi, M. Daquino, *Modellare ontologicamente il dominio archivistico in una prospettiva di integrazione disciplinare*, «JLIS.IT», 2015, 6, p. 13.
[3] G. Michetti, *Metodologie di analisi per l'automazione dei sistemi documentari*, in M. Guercio, *Archivistica informatica. I documenti in ambiente digitale*, Roma, Carocci, 2010, pp. 338.
[4] Le pagine che seguono sono desunte da L.A. Coppola, *La semantica come strumento d'apertura dei documenti d'archivio. Il caso del Catasto Pio-Gregoriano,* Tesi di Diploma (relatore prof.ssa A. Poggi, correlatore prof.ssa L. Giuva), discussa presso l'Università degli studi di Roma "La Sapienza", Dipartimento di Scienze documentarie, linguistico-filologiche e geografiche, Scuola di specializzazione in Beni archivistici e librari, a.a. 2015 - 2016.
[5] Rudi Studer, V. Richard Benjamins, Dieter Fensel, *Knowledge Engineering: Principles and Methods*, "Data & Knowledge Engineering", 25 (1998), n. 1-2, p. 161-198.

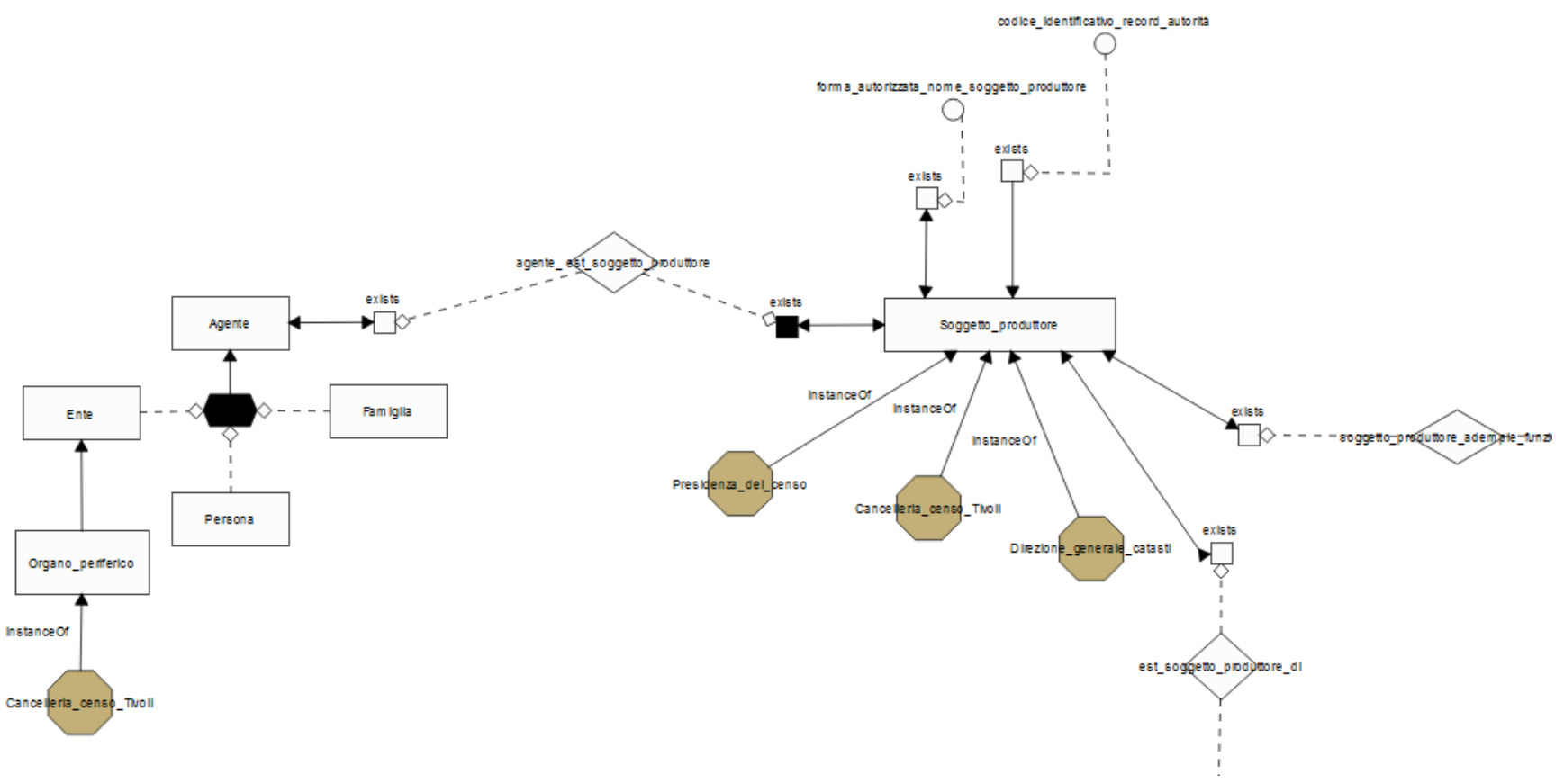

Fig.8

[6] M. T. Biagetti, *Le ontologie come strumenti per l'organizzazione della conoscenza in rete*, "AIDA Informazioni", 28, 2010, n.1/2, p. 14.

[7] D. Lembo, D. Pantaleone, V. Santarelli, D. F. Savo, *Easy OWL Drawing with the Graphol Visual Ontology Language*, Principles of Knowledge Representation and Reasoning: Proceedings of the Fifteenth International Conference, Cape Town, South Africa, April 25-29, 2016, AAAI Press 2016, pp. 573-576.

[8] D. Lembo, D. Pantaleone, V. Santarelli, D. F. Savo, *Eddy: A Graphical Editor for OWL 2 Ontologies*, Proceedings of the Twenty-Fifth International Joint Conference on Artificial Intelligence, New York, NY, USA, 9-15 July 2016, IJCAI/AAAI Press 2016, pp. 4252-4253.

Gli standard internazionali ISAD(G), ISAAR (CPF), ISDIAH e ISDF - in particolare i loro elementi obbligatori[9] - costituiscono uno dei riferimenti di questa pericope, e delle quattro successive, incentrate sulla rappresentazione, in chiave ontologica, delle entità di cui occorre dar conto nella descrizione archivistica. Come illustrato in Fig. 8, un *Agente* - di tipo *Ente, Persona, Famiglia* - potrebbe giocare il ruolo di *Soggetto produttore* di un'*Unità di descrizione*. Del *Soggetto produttore* interessano la forma o le forme autorizzate del nome - costituenti una chiave d'accesso per identificarlo univocamente - e il codice identificativo del record d'autorità. Gli ottagoni, in beige perché, in virtù di quanto detto, relativi all'area di conoscenza in esame, riportano le denominazioni di produttori particolari: *Presidenza del censo, Direzione generale del censo* e *Cancelleria del censo di Tivoli* - quest'ultima distinta dalle prime due istituzioni dal suo essere *Organo periferico* e non centrale dell'amministrazione del censo - preposti all'esecuzione di quanto necessario ai fini dell'attivazione del Gregoriano.

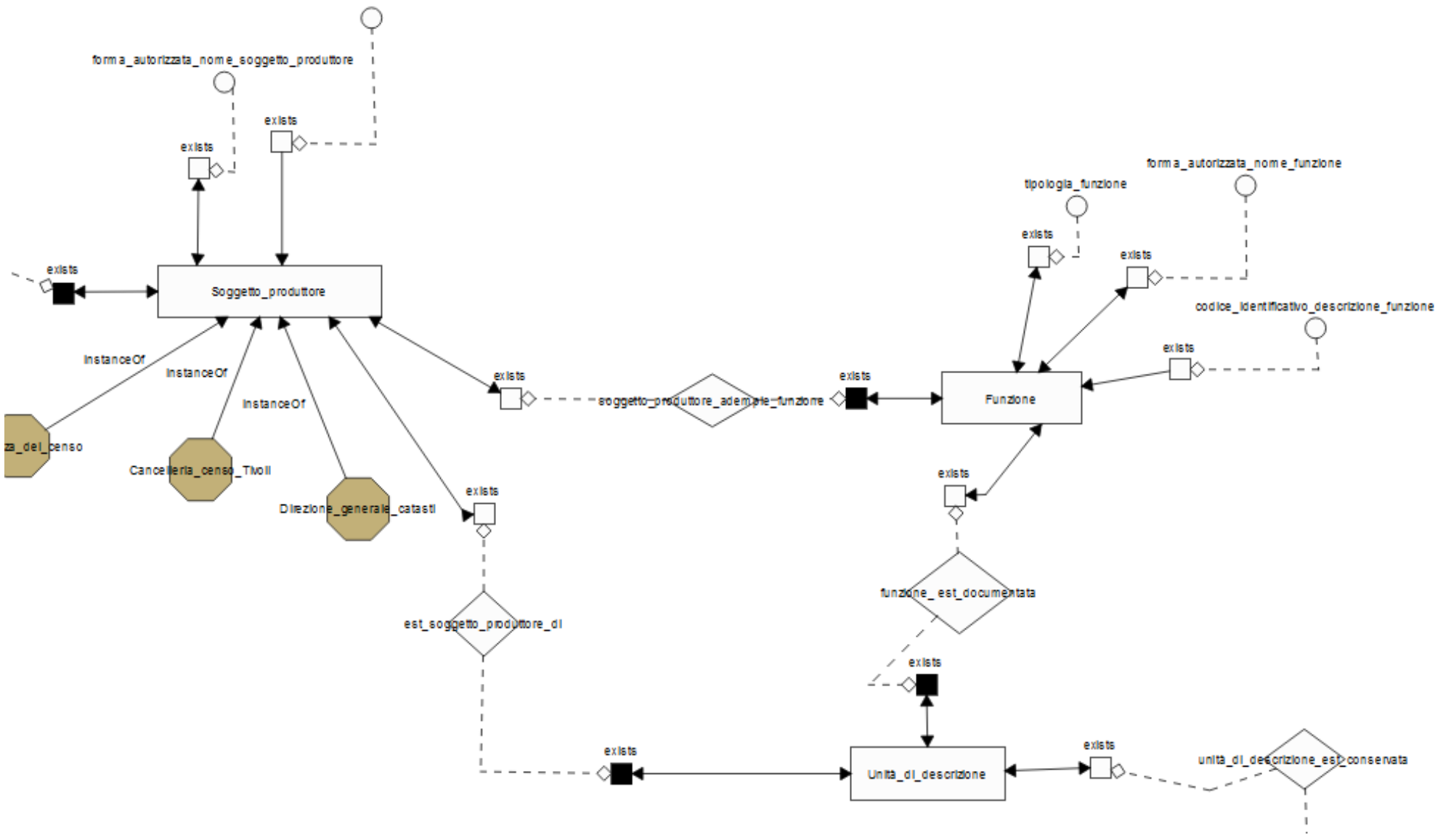

Fig.9

«La descrizione delle funzioni gioca un ruolo vitale nello spiegare la provenienza dei documenti. Le descrizioni delle funzioni possono essere d'aiuto nell'inquadrare i documenti nel loro contesto di produzione e uso. Possono aiutare a spiegare come e perché i documenti siano stati prodotti e in seguito usati, possono illustrare lo scopo e le funzioni da soddisfare per cui questi documenti furono prodotti all'interno dell'organizzazione e possono indicare come s'inseriscano e siano collegati ad altri documenti prodotti dalla stessa organizzazione [...] Mantenere le informazioni sulle funzioni separate sia dai record descrittivi, sia dai record di autorità, significa meno ridondanza delle informazioni e permette la costruzione di sistemi flessibili di descrizioni archivistiche»[10]. La condivisione dell'importanza di esplicitare la relazione che intercorre tra un *Soggetto Produttore*, la *Funzione* da questi svolta e l'*Unità di descrizione* da esso prodotta, sottolineata da ISDF, ha spinto alla modellazione proposta in Fig. 9. Della *Funzione* interessano: la tipologia, la forma autorizzata del nome della funzione e il codice identificativo della descrizione della stessa. Come in Fig. 8, così in Fig. 9 e parimenti in Fig. 10 si è ritenuto, forse a torto, che la data property "codice_identificativo" non fosse obbligatoria; scelta motivata dal fatto che esso potrebbe non essere posseduto da tutte le istanze della classe cui è, di volta in volta, riferito.

[9] Imprescindibili, per ISAD (G): Segnatura/e o codice/i identificativo/i. Denominazione o titolo. Soggetto produttore. Data/e. Consistenza dell'unità di descrizione. Livello di descrizione. Per ISAAR (CPF): Tipologia del soggetto produttore. Forma/e autorizzata/e del nome. Data/e. Codice identificativo del record d'autorità. Per ISDIAH: Forma/e autorizzata/e del nome. Ubicazione e indirizzi. Codice identificativo dell'istituto conservatore. Per ISDF: Tipologia della funzione. Forma/e autorizzata/e del nome. Codice identificativo della descrizione della funzione.

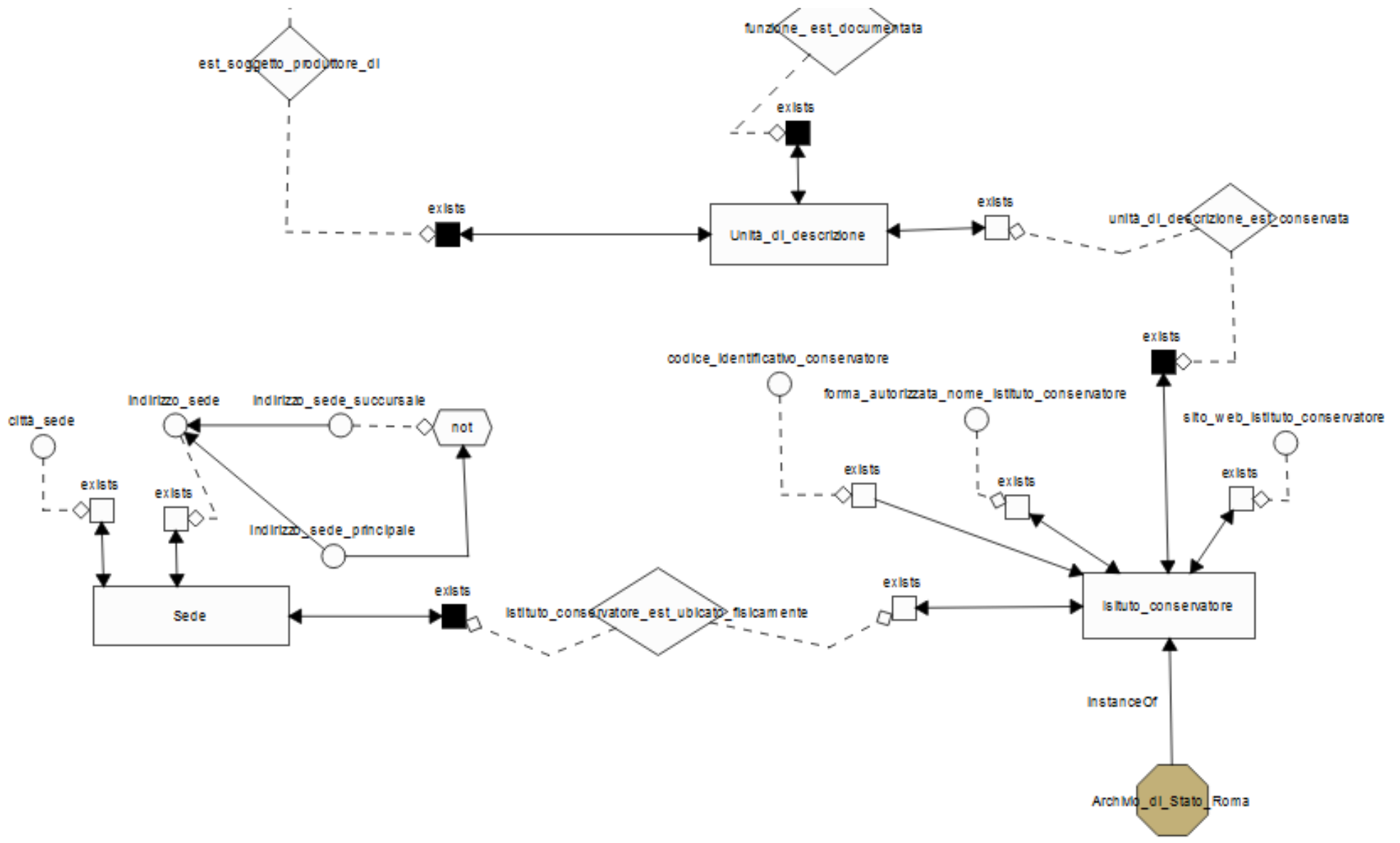

Fig.10

Agli Istituti conservatori d'archivio - fermo restando che la conservazione non venga, naturalmente, unicamente svolta da enti - e allo standard ISDIAH, si ispira la parte del modello in Fig. 10. Un'*Unità di descrizione* viene conservata da un *Istituto Conservatore* mediante una «coerente, coordinata e programmata attività di studio, prevenzione, manutenzione e restauro»[11], e aggiungerei, di fruizione e valorizzazione dello stesso. Tra gli istituti degni di nota, l'Archivio di Stato di Roma. Ogni Istituto avrà una forma autorizzata del nome; potrebbe non avere, viceversa, un codice identificativo. Anziché affidare ad una data property, il terzo degli elementi indispensabili previsti dallo standard, *Ubicazione e indirizzi*, si è scelto di espanderlo attraverso l'introduzione della classe

---

[10] International Standard for Describing Functions. http://media.regesta.com/dm_0/ANAI/anaiCMS/ANAI/000/0111/ANAI.000.0111.0005.pdf, p.8.

[11] Decreto Legislativo 22 gennaio 2004, n. 42, *Codice dei beni culturali e del paesaggio*, ai sensi dell'articolo 10 Legge 6 luglio 2002, n. 137, Sez. II, art. 29.

*Sede,* situata in una certa città e ad un certo indirizzo; l'indirizzo della sede succursale di un istituto differirà da quello della sede principale. Ad ogni istituto pertiene una sede fisica ed una logica, resa qui con la data property "sito_web_conservatore".

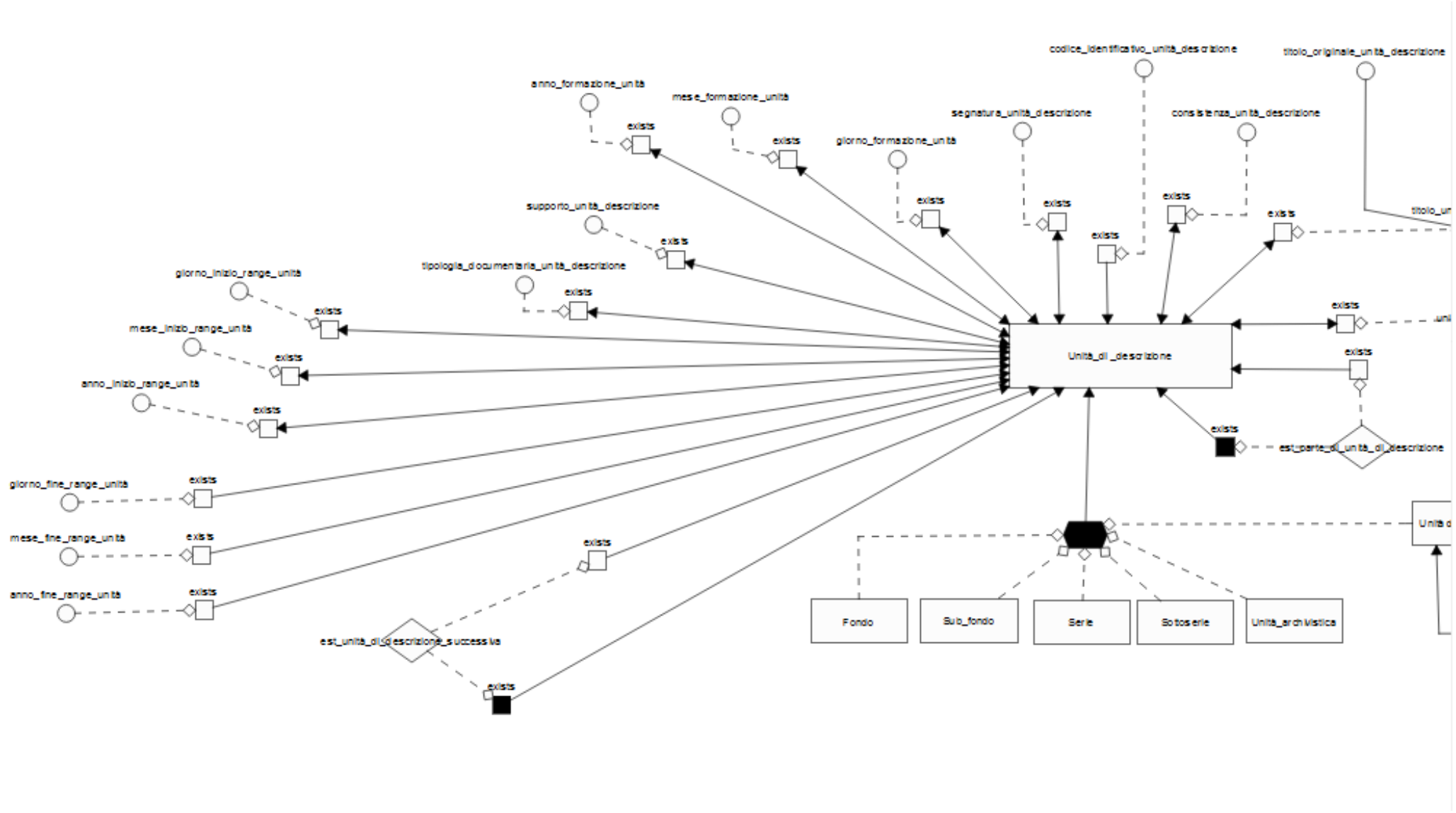

Fig.11

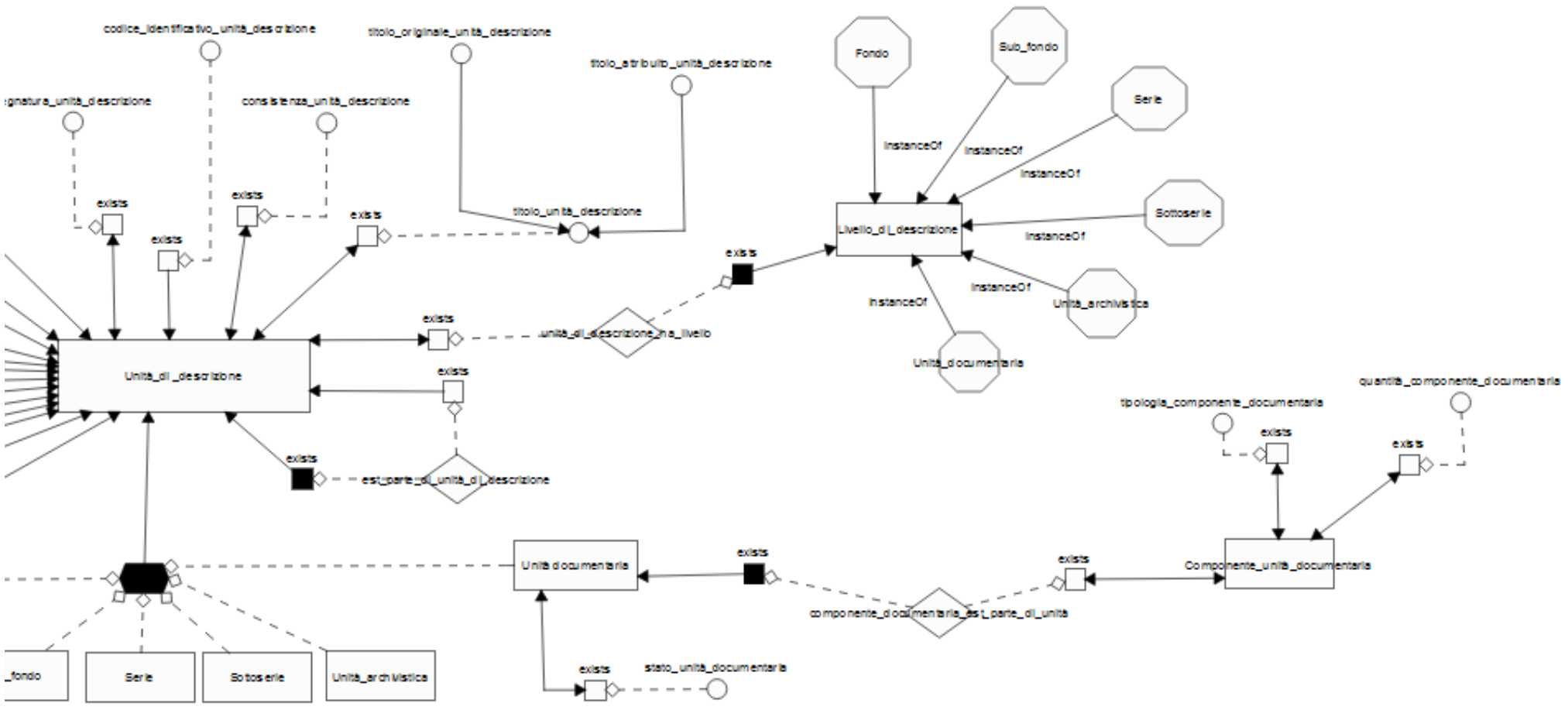

Fig.12

Le Figure 11 e 12, segmentate per ragioni di spazio ma deputate congiuntamente alla rappresentazione dell'*Unità di descrizione*, mostrano come data properties o object properties - alcune con obbligatorietà esplicita - gli elementi descrittivi che ISAD(G) ritiene essenziali allo scambio a livello internazionale, specificandoli ulteriormente e insinuandone di altri, nel tentativo di catturare maggiori informazioni. Un'*Unità di descrizione* sarà descritta per il tramite di: *Soggetto produttore* (vd.Fig.8) e delle sue funzioni e attività (vd. Fig. 9) - *Istituto Conservatore* (vd. Fig 10) - segnatura, codice identificativo, titolo (distinto in originale e attribuito), supporto, consistenza, tipologia documentaria, data di formazione singola (giorno, mese, anno), data iniziale (giorno, mese, anno) di un range cronologico. Non si è, invece, stabilita obbligatorietà per la data finale (giorno, mese, anno) dell'arco temporale, perché l'*Unità di descrizione* potrebbe assimilarsi ad un archivio, o parte di esso, ancora aperto, per esempio. Per quel che concerne il concetto di *Livello di descrizione*, cardine dello standard e della disciplina, si sono battute due piste differenti, probabilmente peccando in ridondanza ma guadagnando in potenziale espressività. In primo luogo, si sono esplicitate l'obbligatorietà e la funzionalità della classe *Unità di descrizione* nel partecipare alla relazione "unità_di_descrizione_ha_livello". I livelli sono molteplici, con grado di dettaglio differente, ma pur sempre circoscritti nell'alveo delle prescrizioni ISAD(G): fondo, sub-fondo, serie, sottoserie, unità archivistica, unità documentaria[12]. In secondo luogo, si sono classificate le *Unità di descrizione*, seguendo il criterio del livello, in: *Fondo, Sub_fondo, Serie, Sottoserie, Unità archivistica, Unità documentaria*. Le classi, tra loro disgiunte, a differenza della rappresentazione individuale visibile nella porzione in alto del diagramma, si prestano ad una verosimile esplosione. Se volessimo dire qualcosa in più, come d'altronde si è provato a fare, per esempio sull'essere *Unità documentaria* - potremmo predicare, in maniera analoga, su una o su tutte le altre entità - saremmo in grado di farlo; l'arricchimento proposto consiste, in questo caso, di una data property che indica lo stato di trasmissione del documento (corrispondente al grado di perfezione) - minuta, copia, originale - e di una object property, tale per cui ogni istanza della classe *Unità documentaria* potrebbe avere una o più componenti, con una loro tipologia e quantità (un sigillo aderente o plumbeo pendente, due fotografie allegate nell'ipotesi in cui si tratti di documenti digitali, nativi o digitalizzati, per esempio)[13]. Il collegamento gerarchico tra le *Unità di descrizione* si è modellato ammettendo che un'unità possa essere o meno compresa in un'altra. Di qui il non inserire l'obbligatorietà: se una sottoserie, infatti, può essere parte di una serie, un fondo potrebbe invece essere del tutto autonomo. Si è tentato, infine, attraverso la object property "est_unità_di_descrizione_successiva" di dare risposta al problema dell'individuazione nell'ambito di una sequenza di unità, e quindi di livelli di descrizione, di quelle successive e di quelle precedenti.

[12] Cfr. M. Morelli, *L'uso delle ontologie in ambito archivistico: un approccio pratico*, Tesi di Laurea (relatore prof.ssa Antonella Poggi) discussa presso l'Università degli studi di Roma "La Sapienza", Facoltà di Lettere e Filosofia, Corso di Laurea in Scienze Archivistiche e Librarie, a.a. 2013-2014.

[13] Vd. International Council on Archives Conseil International des Archives. Expert Group on Archival Description, *Record in Contexts. A conceptual model for archival description*, Consultation Draft v0.1, September 2016, p. 13 (*Record Component)* e p. 26 (*Record State*).

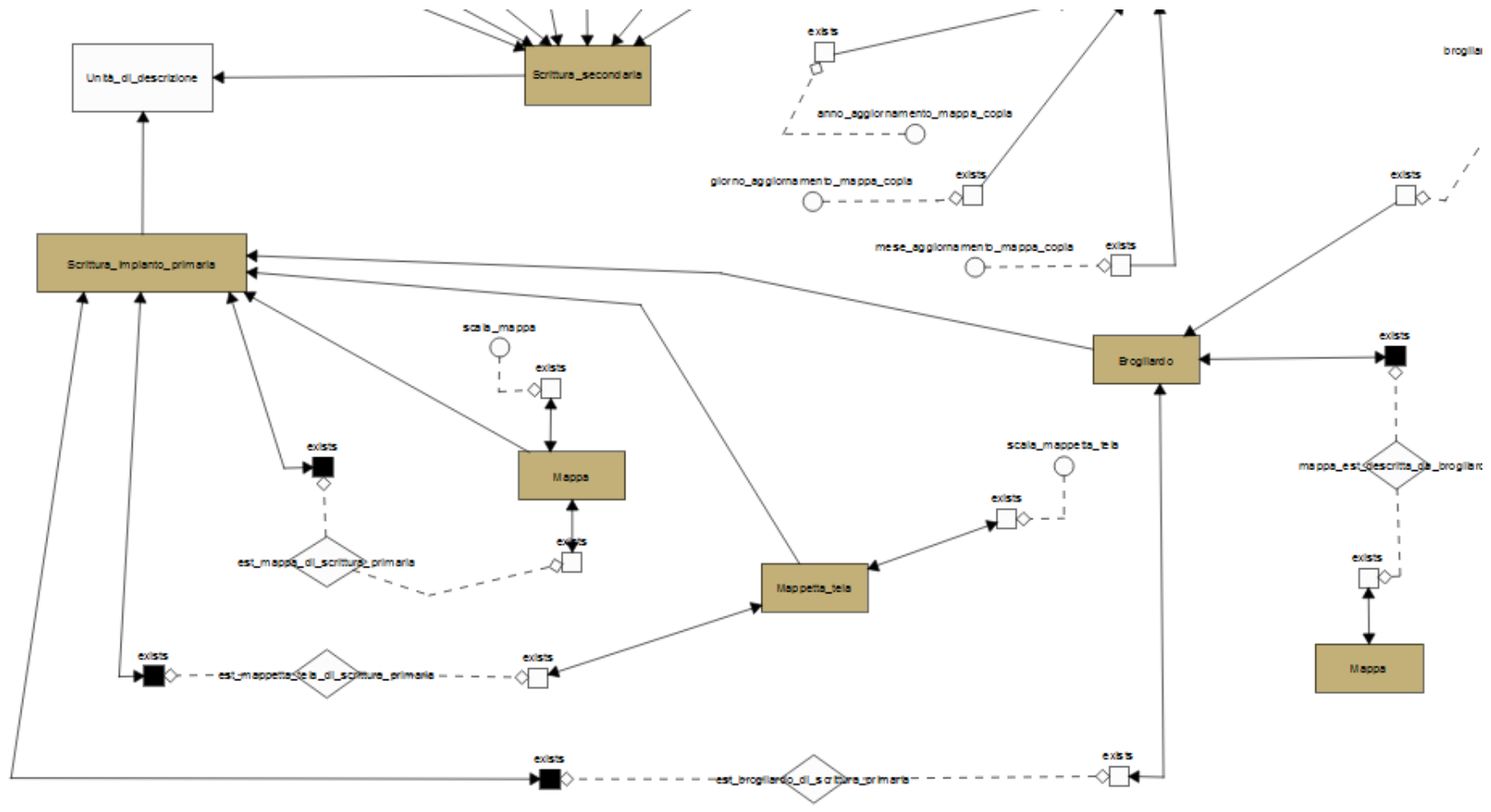

Fig.13

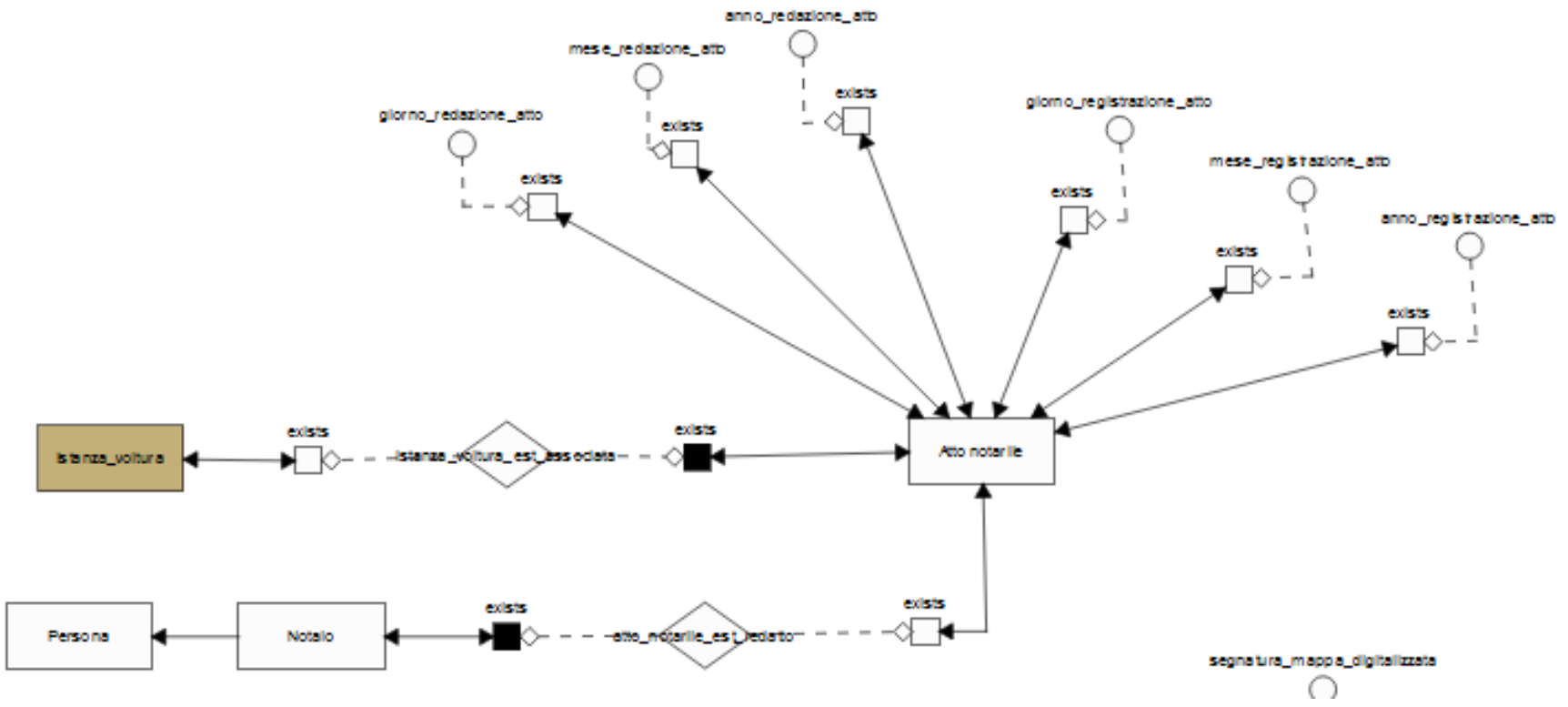

Fig.14

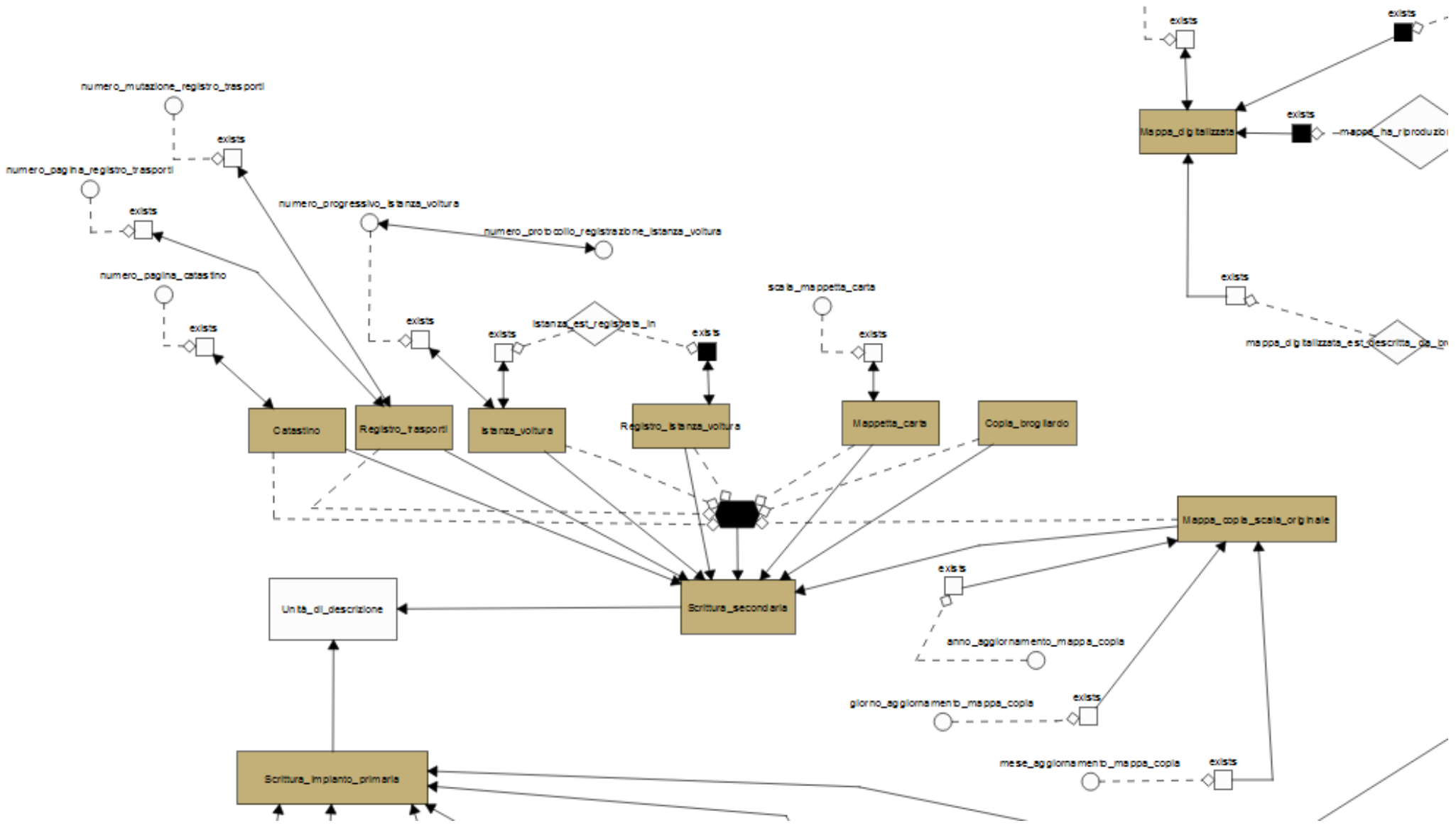

Fig.15

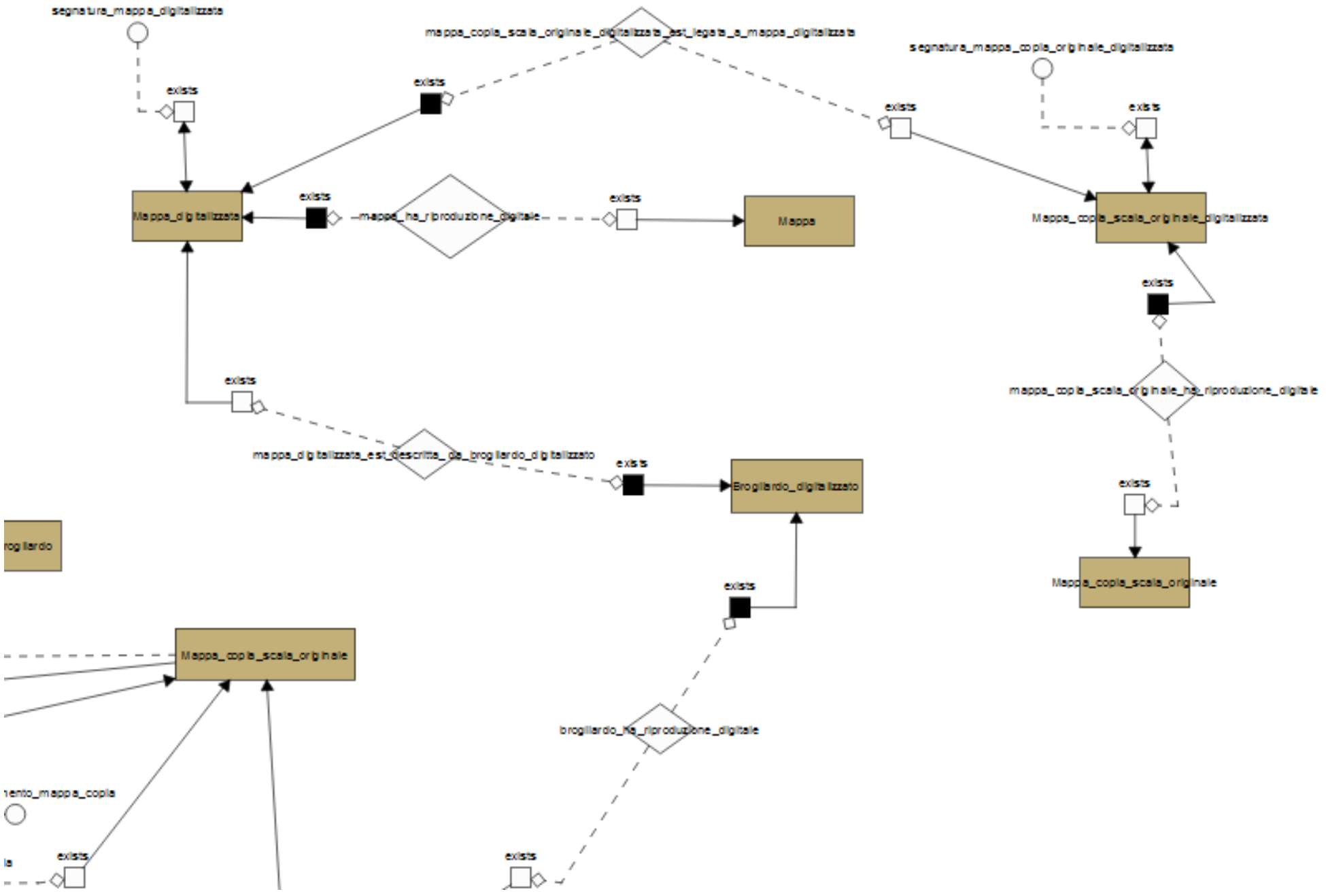

Fig.16

Le Figure 13-16, ben lo denota il fatto che le classi siano per la quasi totalità colorate, rappresentano un pezzetto di *mondo* (Catasto gregoriano), localizzato nello spazio (Archivio di Stato di Roma) e nel tempo (1816-1835), imperniato sul complesso archivistico cui il progetto pontificio diede luogo e sulle sue attuali conservazione e sistemazione. In base allo scopo per il quale furono create le scritture catastali possono essere ripartite in: primarie (*Mappa, Brogliardo, Mappetta tela*), destinate all'attivazione del censimento, e secondarie (*Catastino, Istanza di voltura, Registro dei trasporti, Registro istanza voltura, Mappetta carta, Copia Brogliardo, Mappa copia scala originale*), utili all'aggiornamento (e questo spiega, per esempio, il perché delle data properties anno, giorno, mese di aggiornamento per la classe *Mappa copia scala originale*) e alla custodia del catasto. Vedremo, più avanti, come le prime fossero versate alla *Presidenza del censo* - e di lì giunte a noi - le seconde alle *Cancellerie del censo*, deputate localmente alla revisione del catasto. Le data properties inserite in Fig. 15, riverberano la rete di legami necessari e naturali - il vincolo - che sussiste tra i documenti di un archivio: le Istanze di voltura (fornite di un numero progressivo che coincide con quello di protocollo, ovvero con quello che le connota all'atto della trascrizione in appositi registri) rinviano ai Catastini (segnatamente, al numero di pagina), i Catastini a loro volta ai Registri dei trasporti (al numero d'ordine della mutazione), i quali mettono in atto, in aggiunta, una serie di richiami interni servendosi del numero di pagina. Ancora, Catastini e Registri dei trasporti, fanno cenno al numero progressivo delle Istanze di voltura. Di estrema importanza, inoltre, le relazioni da queste intessute (Fig. 14) con gli atti e i registri dei notai, che lascerebbero prefigurare una stretta connessione con una fetta consistente del patrimonio dell'Istituto: gli archivi notarili. Analogamente connessi *Mappa, Brogliardo, Mappetta tela* (Fig.13). Le tre unità documentarie, sebbene custodite separatamente per ragioni logistiche, sono parte di un'unica unità archivistica. Le mappe, i brogliardi e le copie che delle mappe venivano generate risultano fruibili virtualmente, in numero ragguardevole ma non totale, grazie alle campagne di digitalizzazione portate avanti dall'Archivio di Stato di Roma. Come per il versante cartaceo, e ritorniamo al *fil rouge* delle relazioni, anche per quello non analogico si è cercato di preservare - e se ne tiene conto nella modellazione (Fig. 16) - il nesso esistente tra mappa e rispettivo brogliardo e tra mappa e copia che ne è stata tratta (la segnatura della *Mappa copia originale digitalizzata* fa riferimento infatti, per metà, a quella della *Mappa digitalizzata*: identificativo della *Mappa digitalizzata* COMARCA 140 - identificativo della *Mappa copia originale digitalizzata* corrispondente 322- Comarca 140). L'assenza di pezzi digitalizzati ha indotto tuttavia a non considerare obbligatoria la partecipazione delle classi *Mappa copia scala originale digitalizzata* e *Brogliardo digitalizzato* alle relazioni che li legano alla *Mappa digitalizzata* (Fig.16). La distinzione tra quel che si desume dallo studio della legislazione relativa al Catasto gregoriano, emanata prima e immediatamente dopo l'impianto, e quel che traspare dall'esame della documentazione così come strutturata in Archivio, tra l'ideale normativo (in bianco) e una prassi che ne rappresenta la peculiare messa in esecuzione (in beige), verrà ulteriormente marcata nelle porzioni di ontologia illustrate di seguito con l'ausilio non solo del colore ma anche della lettera R ("rappresentazione").

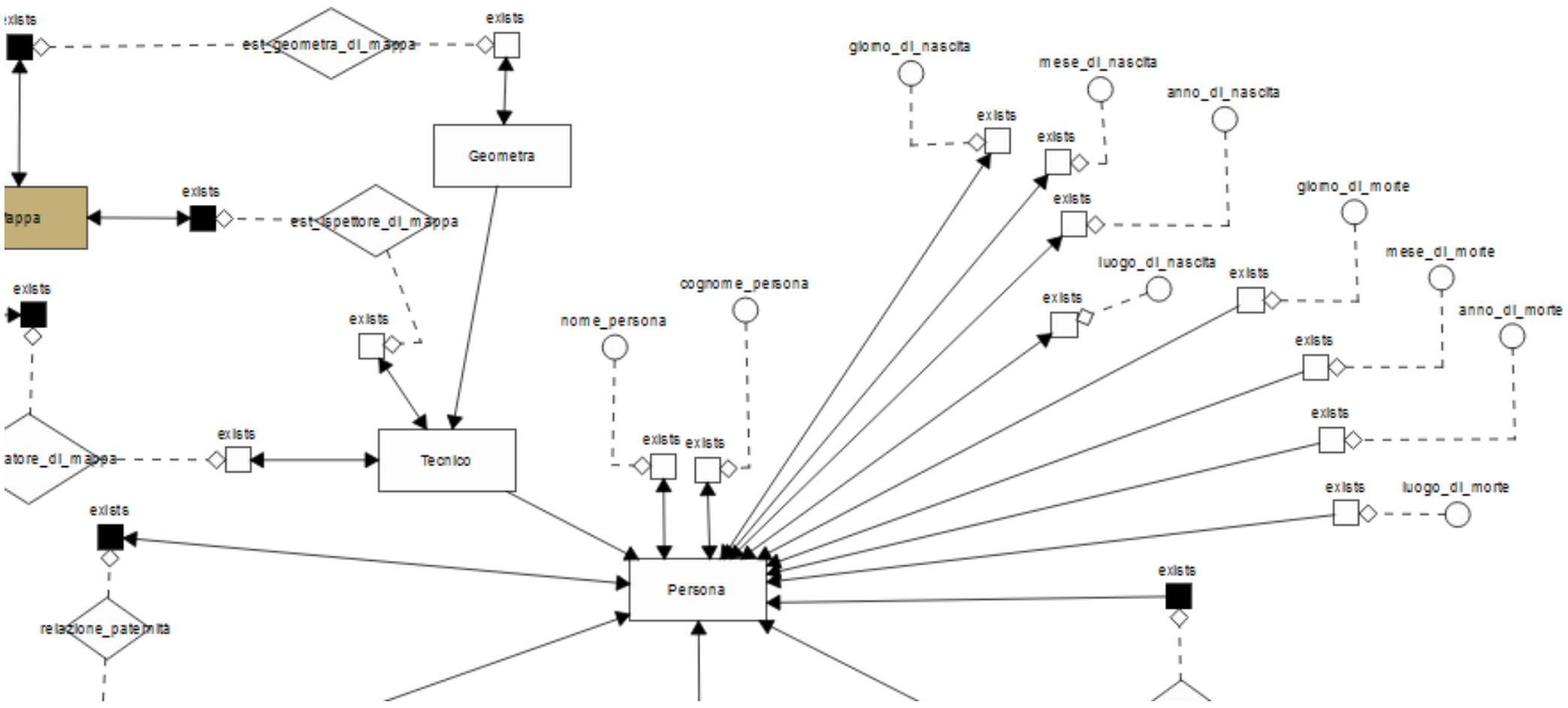

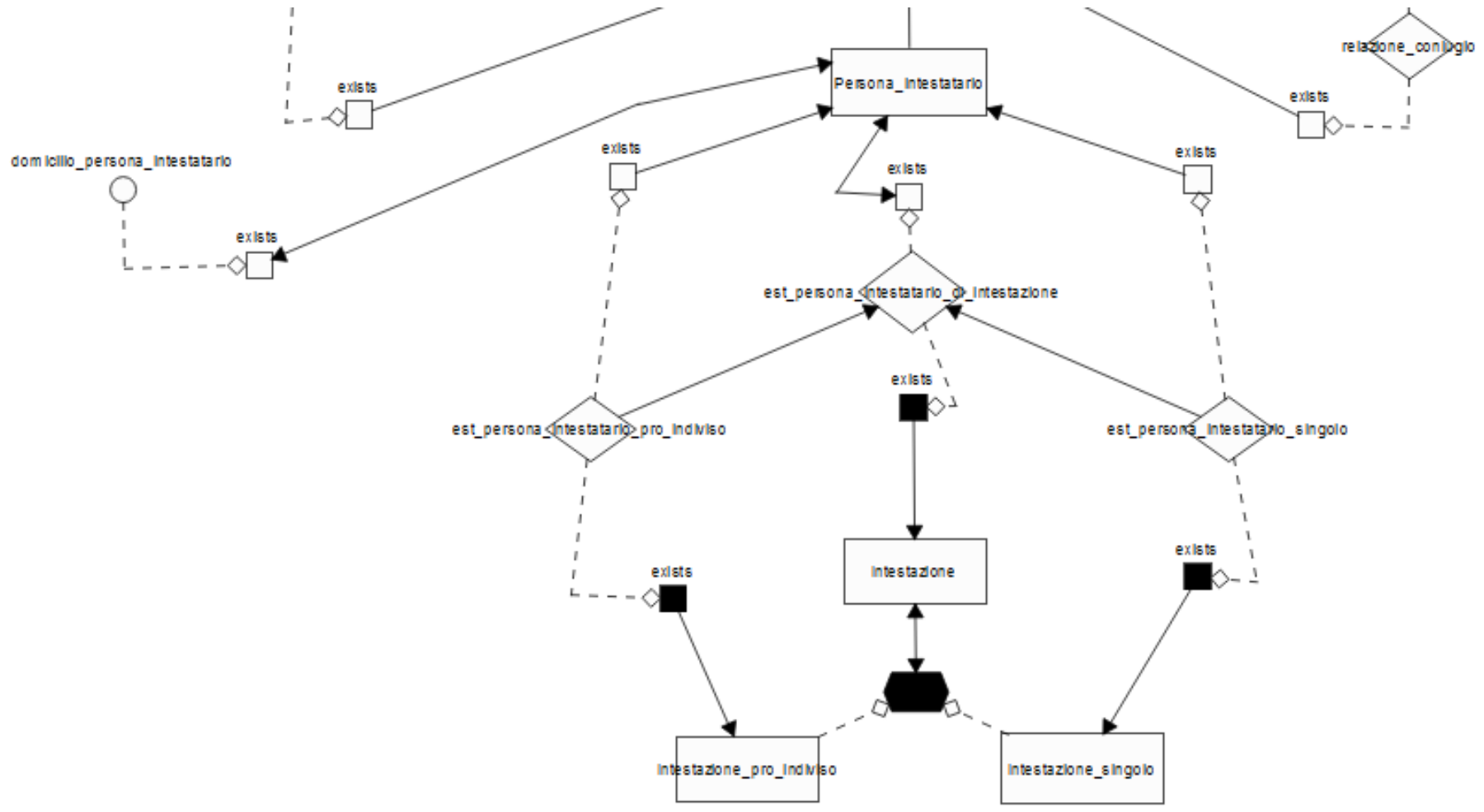

Fig.17

*Persona Intestatario*, sottoclasse di *Persona*, ne eredita, come si evince dalla Fig. 17, gli attributi. Oltre al domicilio dell'intestatario di un bene immobile potremmo pertanto conoscere: nome, cognome, luogo di nascita, data di nascita (giorno, mese, anno) ed eventualmente luogo e data di morte (giorno, mese, anno). Potrebbe non avere figli; parimenti discrezionale una relazione di coniugio. Viceversa, obbligatorio l'avere un padre. Una *Persona Intestatario* lo è per almeno un'*Intestazione*, singolarmente o congiuntamente ad altri (*pro indiviso*, con fratelli, nipoti per esempio) con cui detenga il possesso e/o la proprietà di uno più immobili.

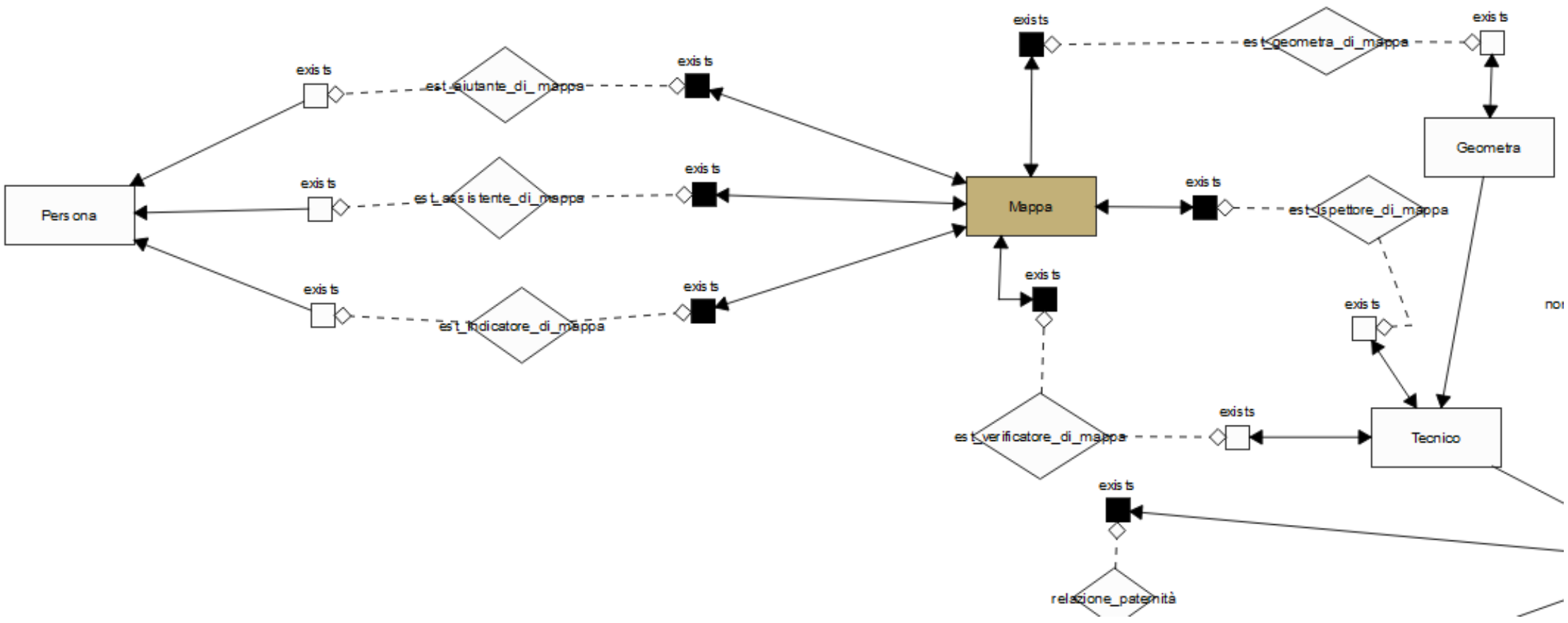

Fig.18

Una *Persona* potrebbe ricoprire il ruolo di assistente, aiutante o indicatore nell'ambito delle operazioni che portano all'elevazione di una *Mappa*; un'azione che necessita, per converso, di un *Geometra*, sottoclasse di *Tecnico*, e di un *Tecnico*, a sua volta sottoclasse di *Persona*, in veste di ingegnere ispettore o verificatore.

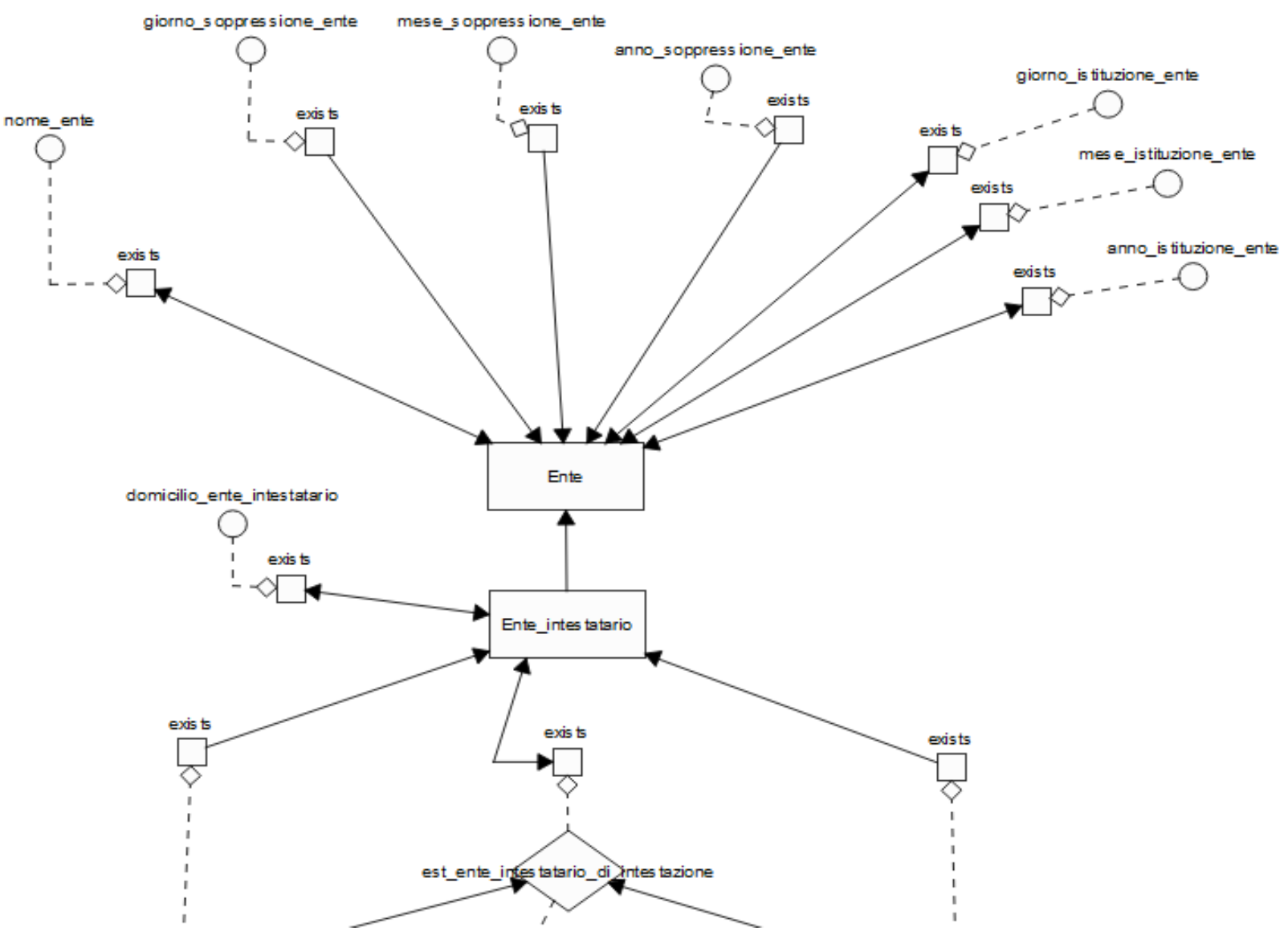

Fig.19

La Fig. 19, accresce e si ricollega al commento della Fig. 17. Un'*Intestazione*, sia essa goduta individualmente o cumulativamente, riguarda una *Persona Intestatario* o un *Ente Intestatario*. Al pari delle classi *Persona* e *Persona Intestatario*, *Ente Intestatario* in quanto sottoclasse di *Ente*, ne eredita le data properties - nome, data di istituzione (nome, mese, anno), eventuale data di soppressione (giorno, mese, anno) - aggiungendo il valore, domicilio.

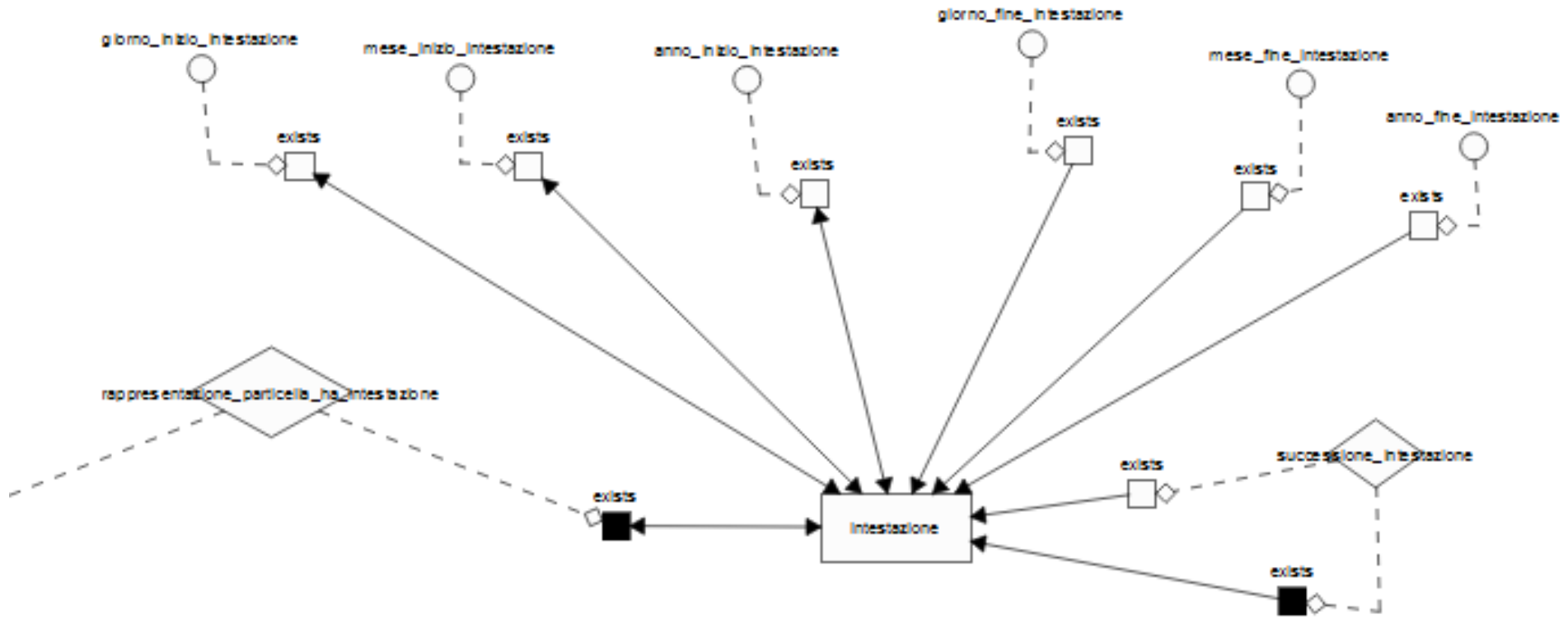

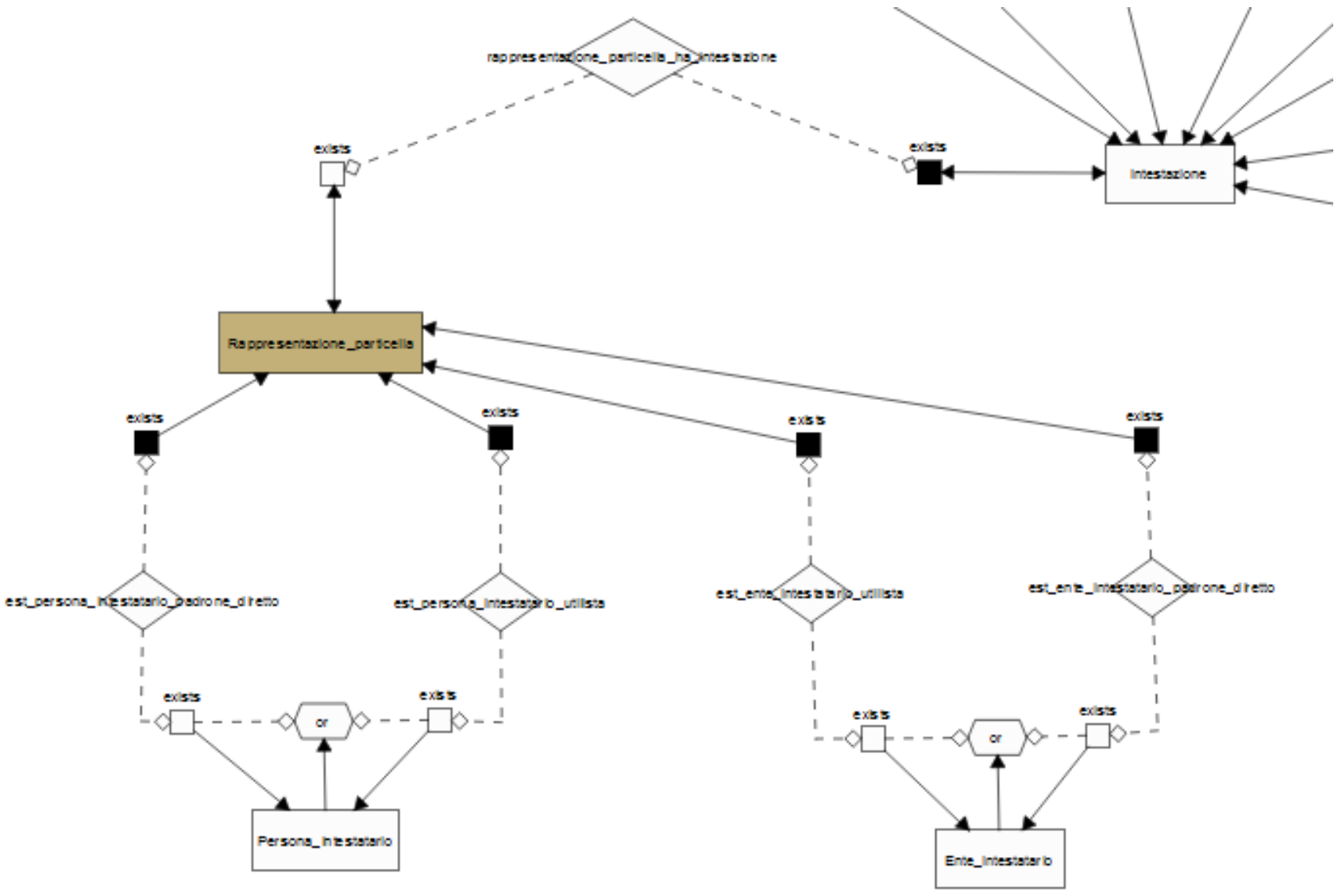

Fig.20

Della classe *Intestazione* interessano: la data d'inizio (giorno, mese, anno) e la, probabile, data di fine (giorno, mese, anno). La object property "rappresentazione_particella_ha intestazione" esprime il rapporto 1:1 che sussiste tra *Intestazione* e *Rappresentazione particella*: ogni I*ntestazione* è relativa ad una sola *Rappresentazione_particella*, e viceversa, ciascuna *Rappresentazione_particella* ha una sola *Intestazione*. Precisiamo: non esistono due *Intestazioni* di una medesima *Rappresentazione particella* che abbiano la stessa data d'inizio; tuttavia, l'*Intestazione* - e quindi l'intestatario - di un bene potrebbe, e di fatto, cambia nel corso del tempo. Di qui la scelta di tener conto della successione, eventuale, delle intestazioni. Infine, una persona o ente intestatario possono esercitare su un immobile due tipi diversi di diritti reali: la proprietà (padrone diretto) e il possesso (utilista). Una *Rappresentazione particella* avrà, naturalmente, o un padrone diretto o un utilista; questo non esclude, però, che chi detenga il dominio utile su un terreno o un edificio (per esempio a titolo di enfiteusi) non possa esserne proprietario, invece, di altri.

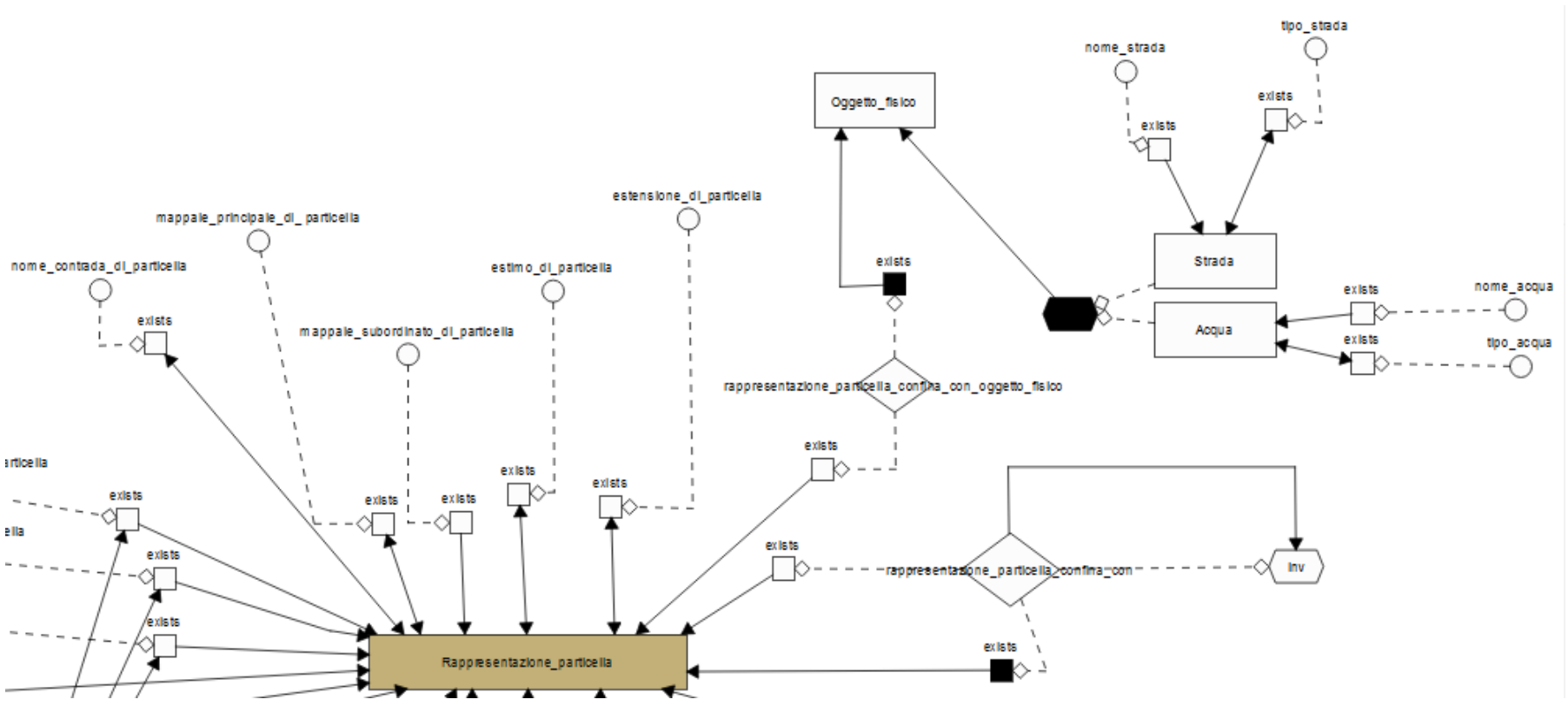

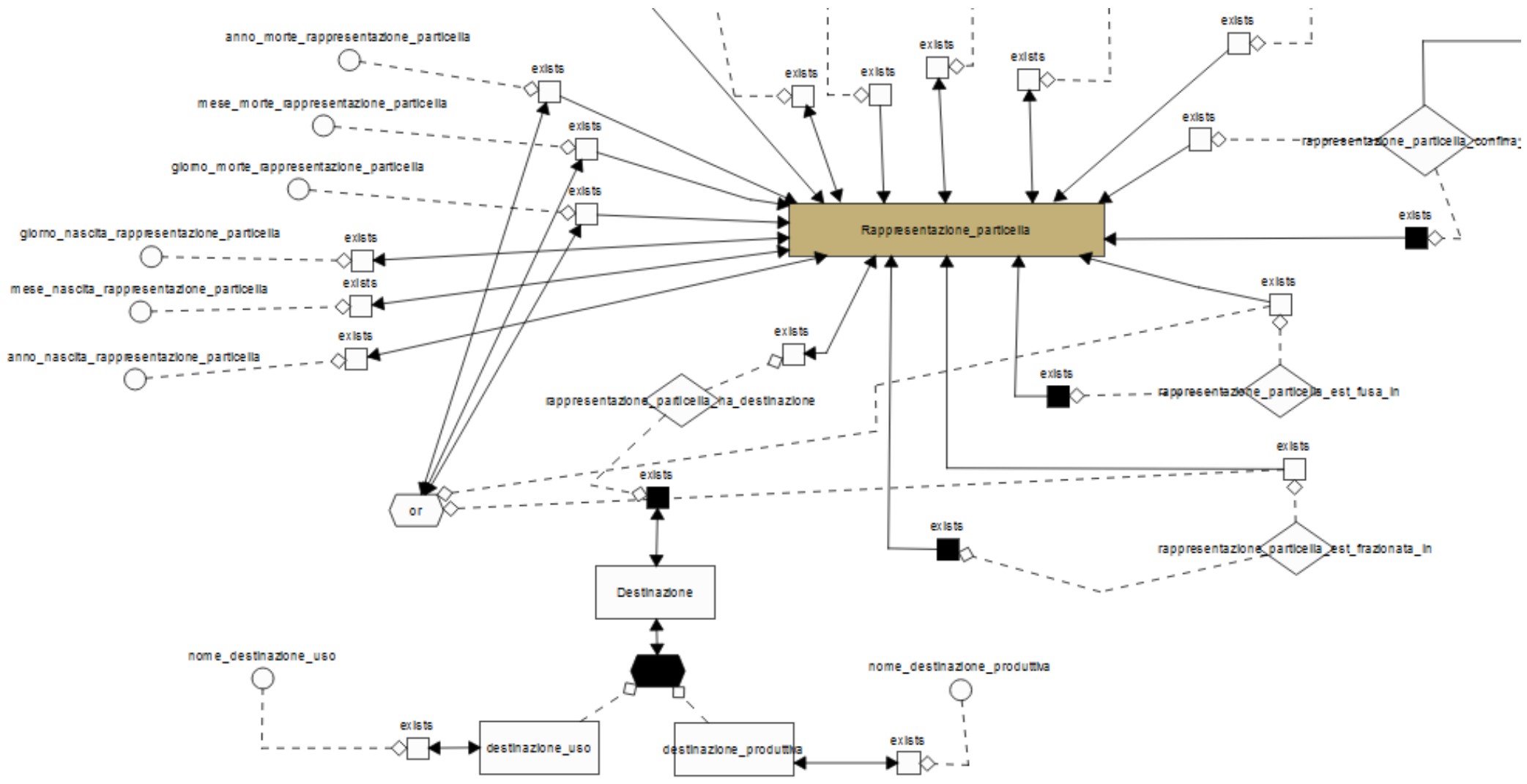

Fig.21

*Rappresentazione particella* contempla una quantità significativa, proporzionale al peso che il concetto possiede nell'economia del modello, di data properties e object properties; appartengono alla prima tipologia la denominazione della contrada (o vocabolo) in cui è ubicata, il mappale principale - numero progressivo per ciascuna *Mappa* che la identifica all'interno di quella e di ogni altra scrittura catastale - il mappale subalterno, l'estimo, l'estensione, la data di inizio rappresentazione (giorno, mese, anno) ovvero la quella della rappresentazione cartografica che per la prima l'abbia censita, una data di morte (giorno, mese, anno) laddove per morte si intende il venir meno della *Rappresentazione particella* nella configurazione originaria, perché oggetto di fusione e frazionamento. Non a caso, la modellazione stabilisce che la data di morte coinciderà o con il frazionamento o con la fusione. Una *Rappresentazione particella* avrà una destinazione d'uso o produttiva, di cui è utile fornire la denominazione; le denominazioni vengono, infatti, tratte da liste controllate, in cui ad ogni termine corrisponde una semantica definita, emanate con apposito regolamento. Altra questione essenziale: i confini. Una *Rappresentazione particella* potrebbe confinare con un'altra - per simmetria, la seconda a sua volta confinerà con la prima - ma potrebbe anche esser adiacente ad una *Strada* e/o ad un *Oggetto fisico* che si è chiamato *Acqua* e non corso d'acqua per motivo di opportunità: banalmente, per non limitare la classe ai soli torrenti, fiumi ma far in modo che includesse laghi, per esempio.

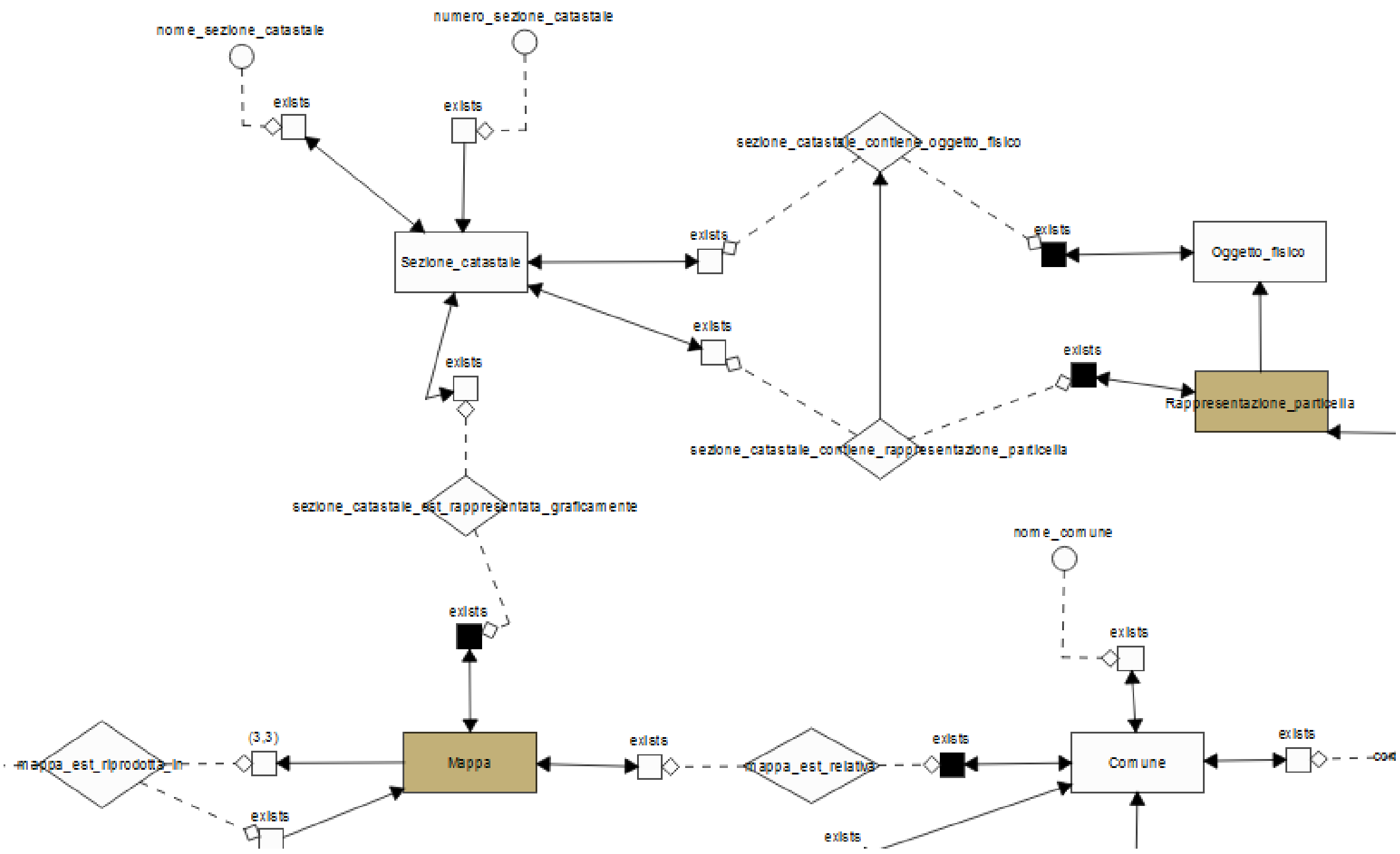

Fig.22

L'ontologia cattura la classe *Comune* nel duplice ruolo di entità fisico-geografica e amministrativa. La Fig. 22 incentrata sul primo dei due aspetti prende avvio dal concetto di *Sezione catastale* (individuata mediante il nome ed eventualmente il numero, tipicamente un ordinale) legato alla *Mappa* per il tramite della object property "sezione_catastale_est_rappresentata_graficamente"; ogni *Mappa* rappresenta graficamente, infatti, una ed una sola *Sezione catastale*. Ogni istanza di *Mappa*, inoltre, farà riferimento ad un sola istanza di *Comune*. Tuttavia, un Comune potrà essere, ed è, - basti pensare alla città di Tivoli - descritto in più rappresentazioni cartografiche.

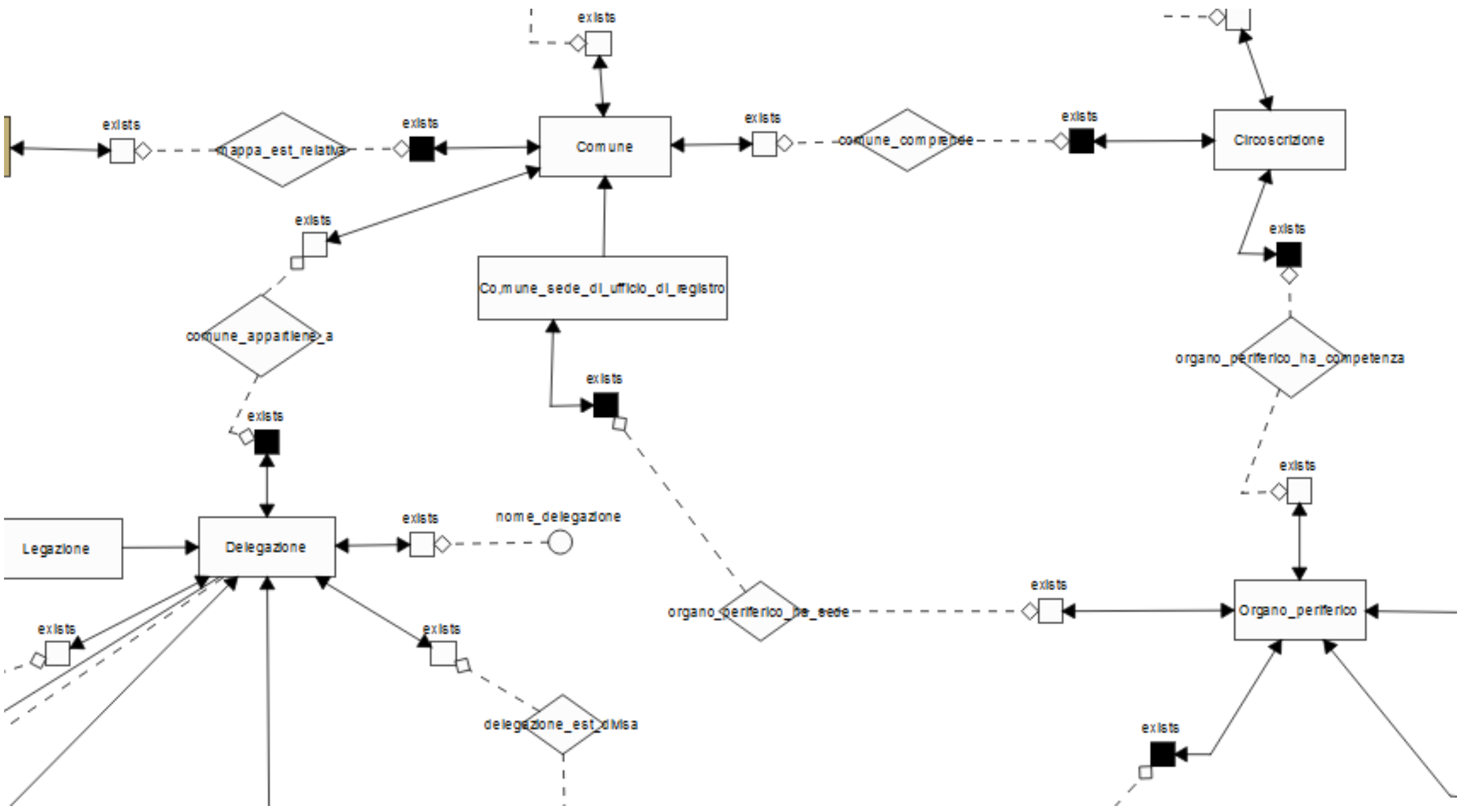

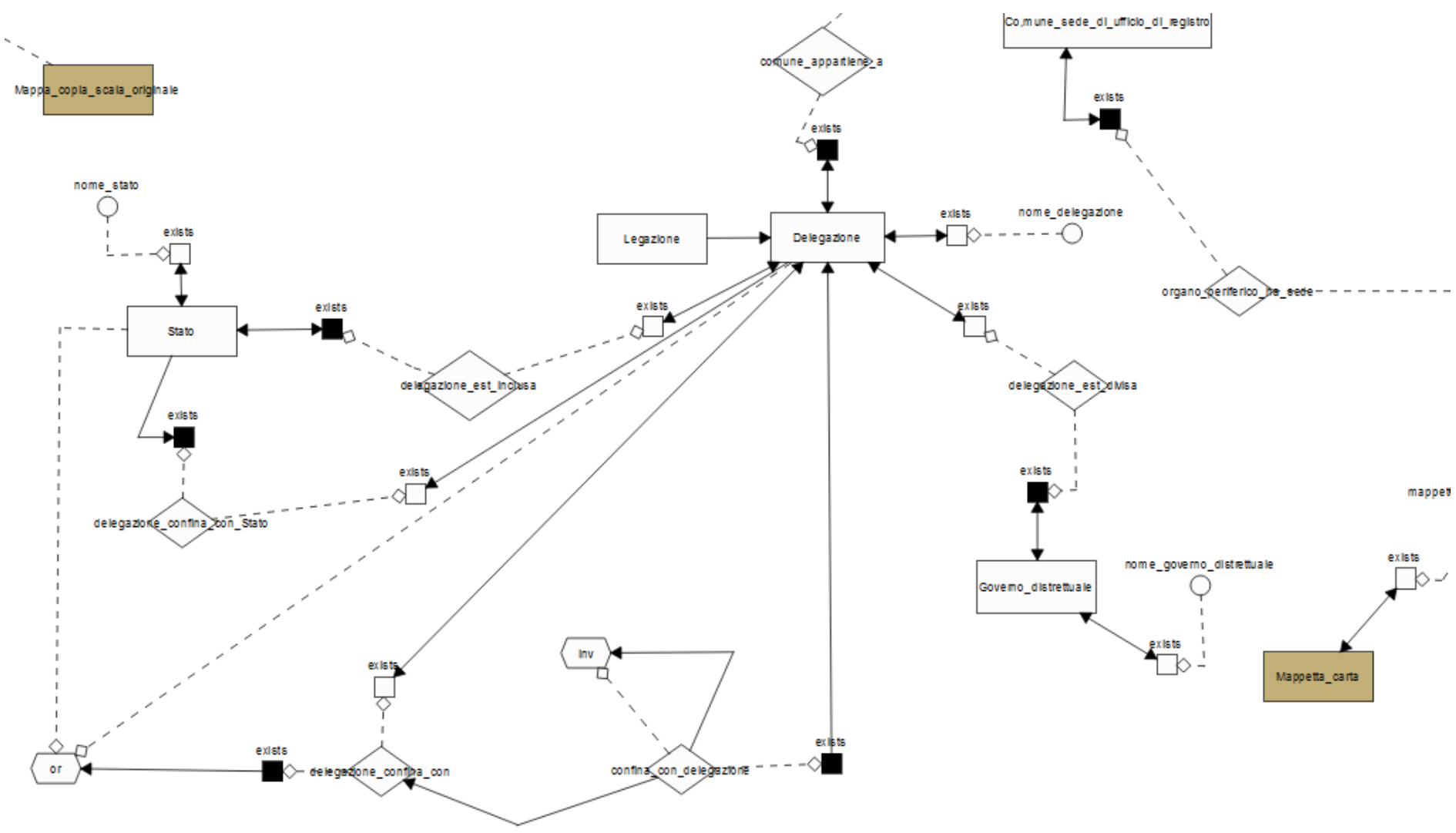

Fig.23

Quanto tratteggiato nelle due figure trae spunto dal riassetto politico-burocratico dello Stato Pontificio sancito con Motu proprio del 1816. Un *Comune* ingloba più circoscrizioni, sulle quali esercitano autorità le Cancellerie del censo (sottoclasse di *Organo periferico*), per territorio di competenza, ubicate presso le sedi di Ufficio di registro. Le delegazioni, incluse in uno Stato e ripartire in governi territoriali, assumono l'appellativo di *Legazione* (sottoclasse di *Delegazione*) se governate da un cardinale, includono più Comuni e possono confinare con un'altra *Delegazione* o con uno *Stato*.

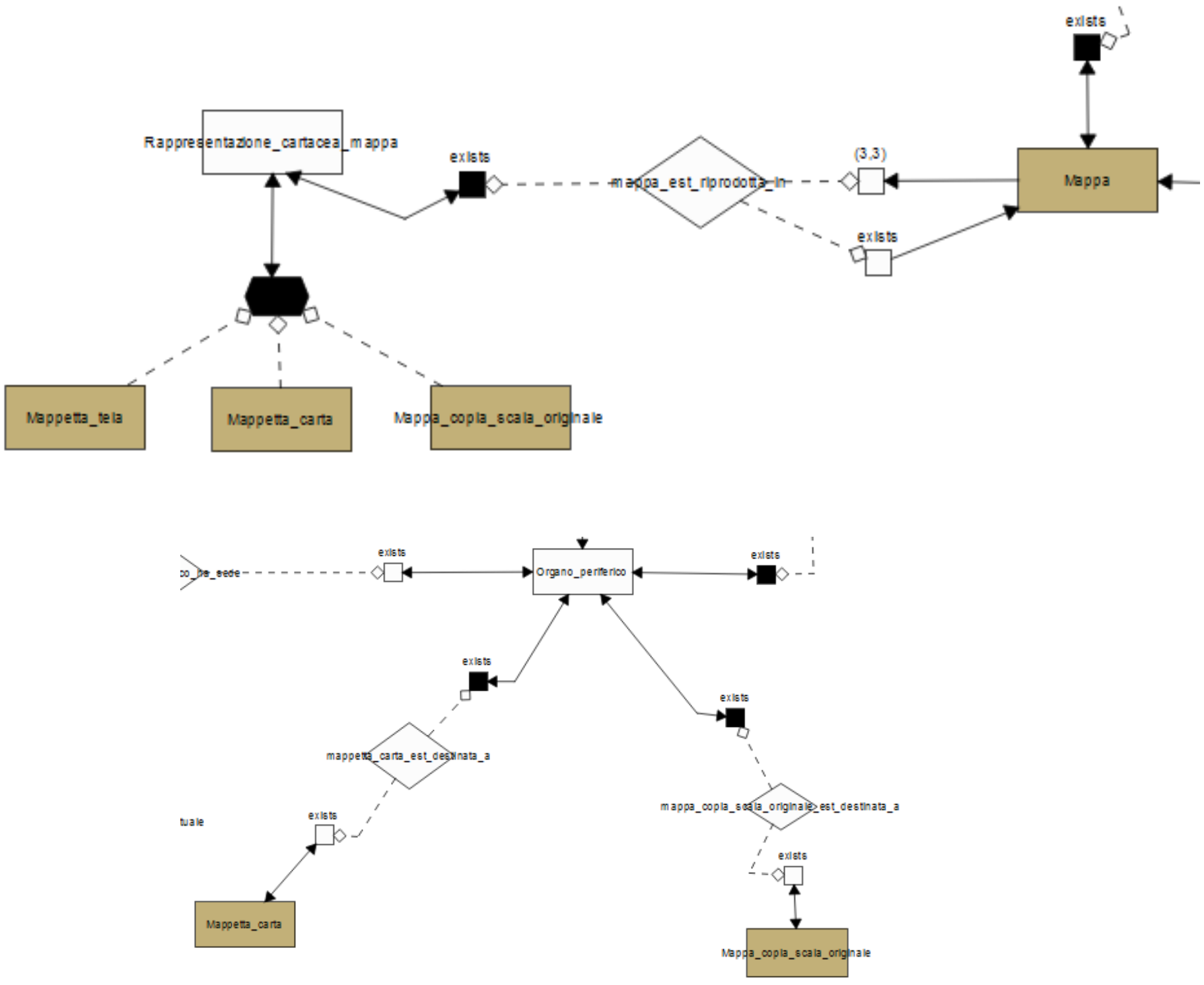

Fig.24

Delle mappe sarebbero state tratte tre copie: una, utilizzando la stessa scala della *Mappa* (1:2000) altre due - le mappette - in scala ridotta (1:4000 o 1:8000) su tela e su carta. La *Mappa Copia in Scala originale* e la *Mappetta su carta* vengono inviate alla Cancelleria del censo, territorialmente competente, corredate della *Copia del Brogliardo*, eseguita a partire dall'originale (Fig.25). La *Mappetta su tela,* la *Mappa* e il rispettivo *Brogliardo* - in cui trova posto la descrizione delle singole particelle di quella *Mappa* - vengono trasmesse al Dicastero del censo (gli istituti centrali cui ci si riferisce vengono catturati dalla base di conoscenza come istanze notevoli di *Soggetto produttore*, Fig.8).

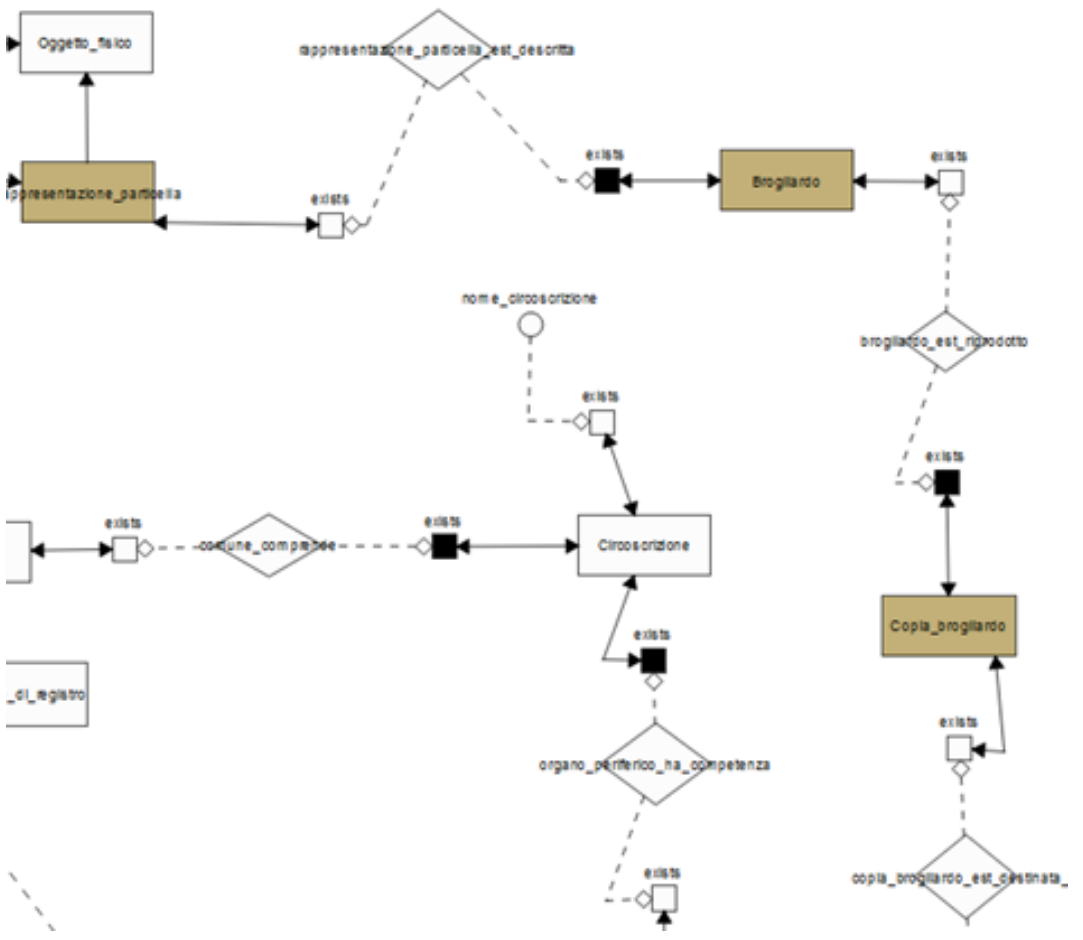

Fig.25

La formalizzazione ontologica del Catasto gregoriano così come raccontato dai documenti archivistici visionati (nota l'uso del colore e la pervasiva presenza della lettera R negli appellativi di classi e proprietà) si riflette pienamente, forse più che altrove, nei segmenti di diagramma che ci accinge ad illustrare.

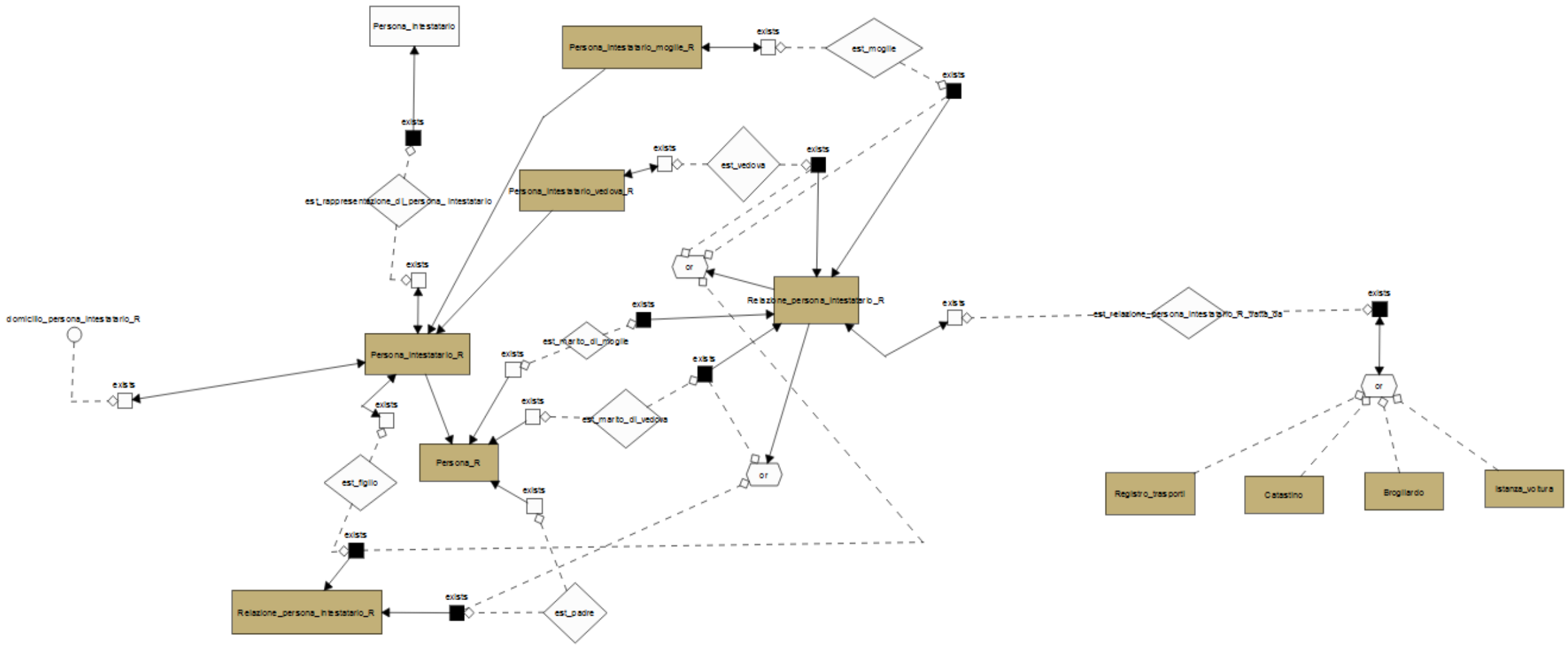

Fig.26

Degli individui intestatari di immobili si trova rappresentazione in una messe di tipologie documentarie; tuttavia, com'è noto, fattori disparati possono e hanno condotto al danneggiamento, alla dispersione di singoli documenti o nuclei documentari. Si spiega in questo modo la partecipazione non obbligatoria della classe *Persona Intestatario* alla "est_rappresentazione_di persona_intestatario". *Persona Intestatario R,* sottoclasse di *Persona* R, risulta identificata dal solo patronimico, nel caso di intestatario di genere maschile, e dalla formula di paternità addizionata al nome e cognome del coniuge, nel caso di intestatario di genere femminile, sia essa vedova o meno. A riprova di quanto detto poc'anzi, dunque del legame indissolubile che la modellazione ha qui con la realtà archivistica, questo segmento e quelli a venire dichiarano tutti la loro provenance. La classe *Relazione Persona Intestatario* R - ovvero i rapporti di paternità e/o di coniugio qui reificati, a differenza di quanto accadeva prima (Fig. 17) poiché consentono di identificare ciascuna *Persona Intestatario* R - si desume da quattro scritture catastali: istanze di voltura, brogliardi, registri dei trasporti e catastini.

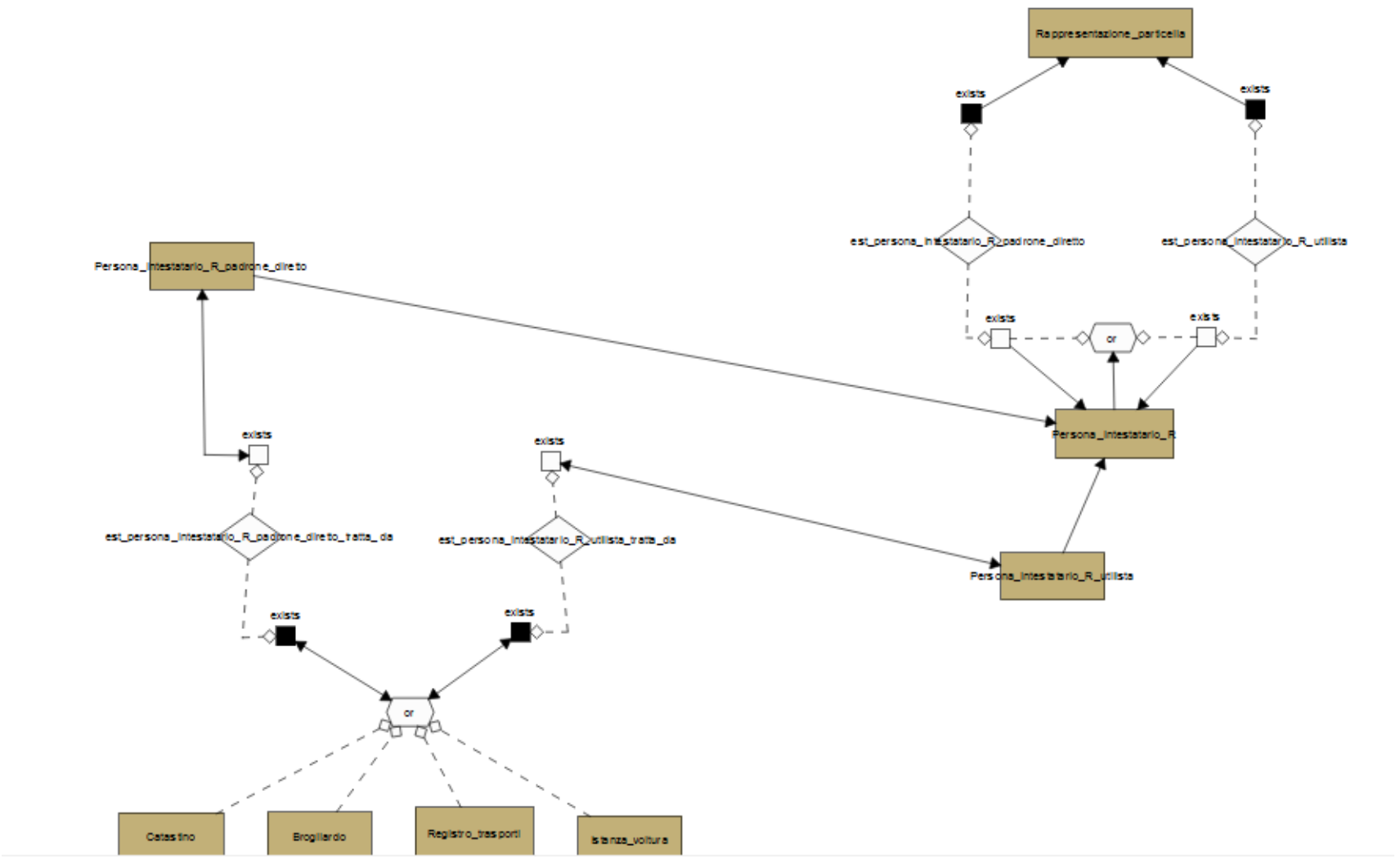

Fig. 27

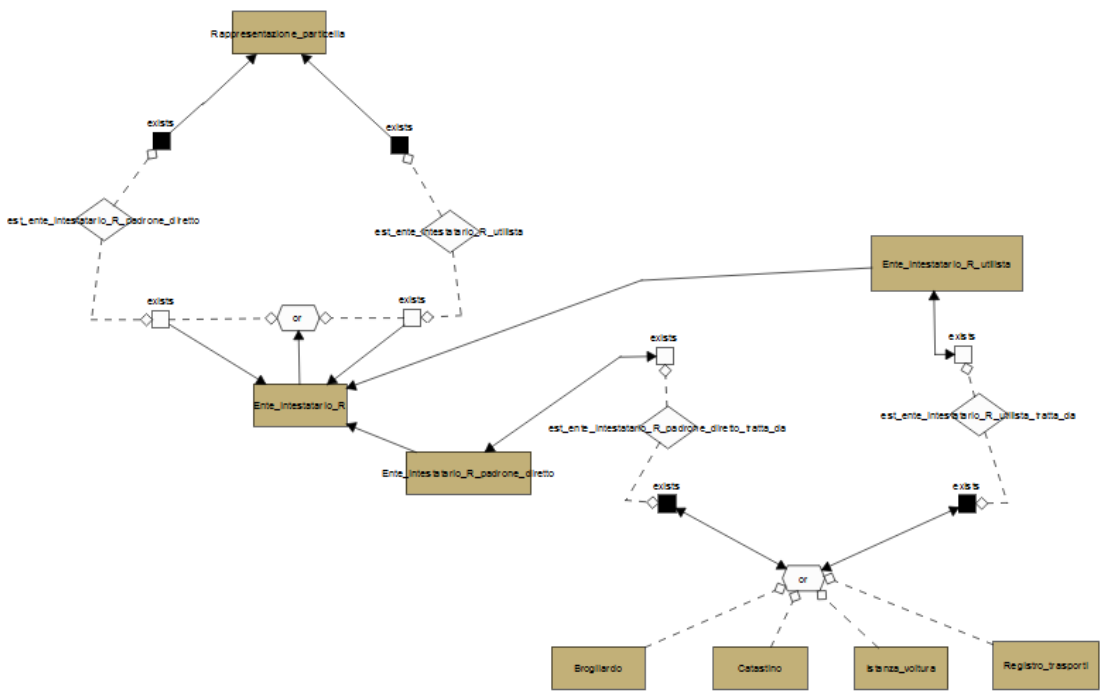

Fig.28

Una *Rappresentazione particella*, e lo si è già detto (Fig. 20), è legata ad un intestatario (ente o persona) che ne usufruisca o in veste di possessore o di proprietario. Si è scelto di reificare, vale a dire di rendere attraverso le classi *Persona Intestatario R utilista*, *Persona Intestatario R padrone diretto* (Fig. 27), *Ente Intestatario R utilista* e *Ente Intestatario R padrone diretto* (Fig. 28) quelle che in precedenza erano object property (Fig. 20), al fine di conseguire maggiore espressività e poter affermare qual è la fonte da cui si trae l'informazione.

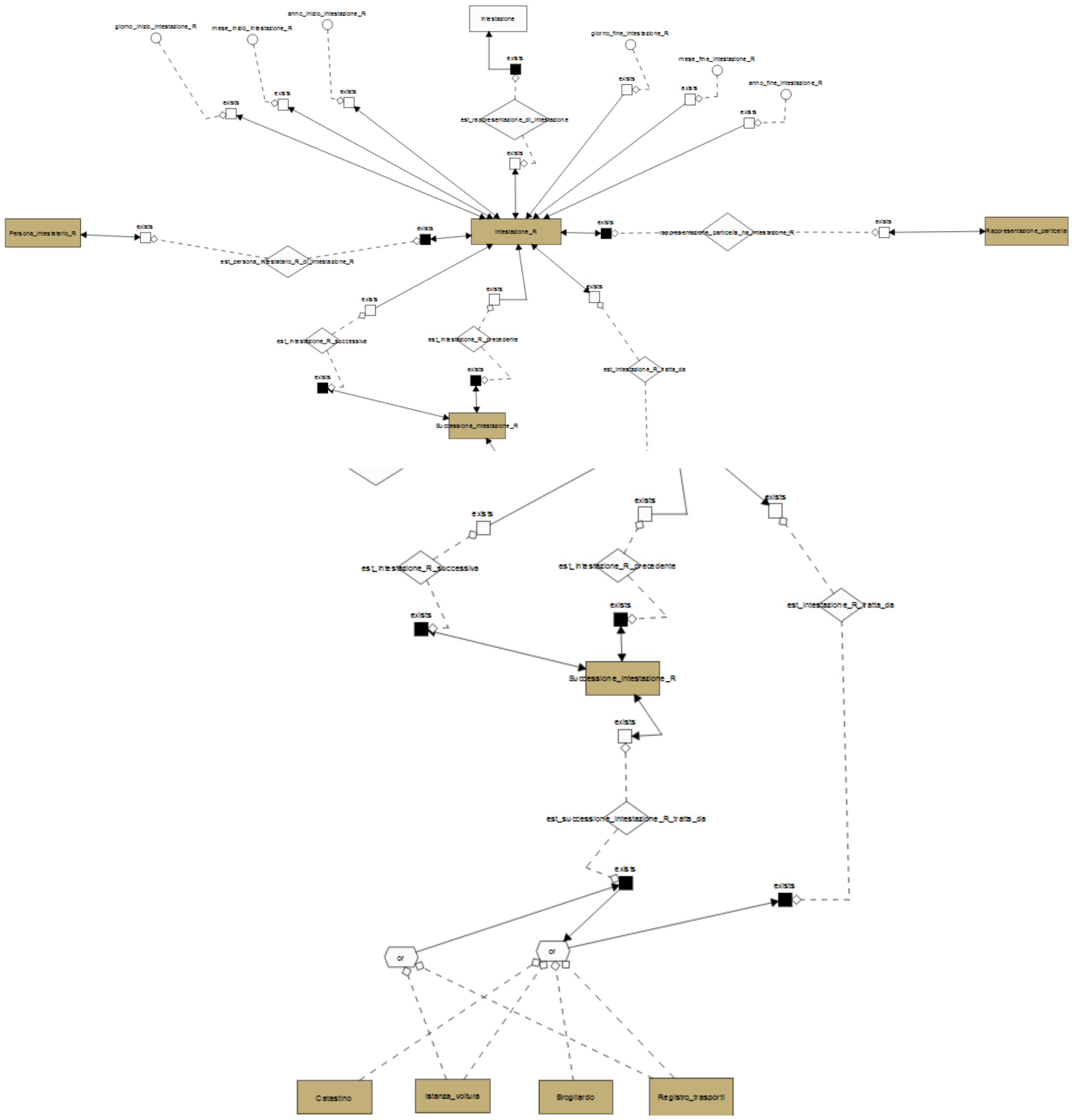

Fig. 29

Gli espedienti di modellazione cui si è fatto ricorso e le motivazioni addotte, si rinvengono anche in questa porzione dedicata alla rappresentazione dell'intestazione (*Intestazione* R), quindi del legame che associa la persona al bene - *Rappresentazione particella* ha *Intestazione* R di cui gode una *Persona Intestatario* R - tal quale emerso dall'analisi archivistica. Non tutte le intestazioni hanno relativa evidenza documentaria, come rilevato per gli intestatari (persona o ente) ma anche per i tecnici coinvolti nella produzione di mappe (Fig.32). Della classe *Intestazione* R interessano la data (giorno, mese, anno) di inizio, e laddove presente, quella di fine (giorno, mese, anno). Mentre di una *Intestazione* R possiamo trovare notizia in un catastino, o in un brogliardo, o in un'istanza di voltura o in un registro dei trasporti, se volessimo conoscere, la successione con cui un bene passa da un individuo ad un altro e quindi capire qual è - nell'ambito di una successione - l'intestazione precedente e quale quella successiva, dovremmo far appello alle sole istanze di voltura e registri dei trasporti.

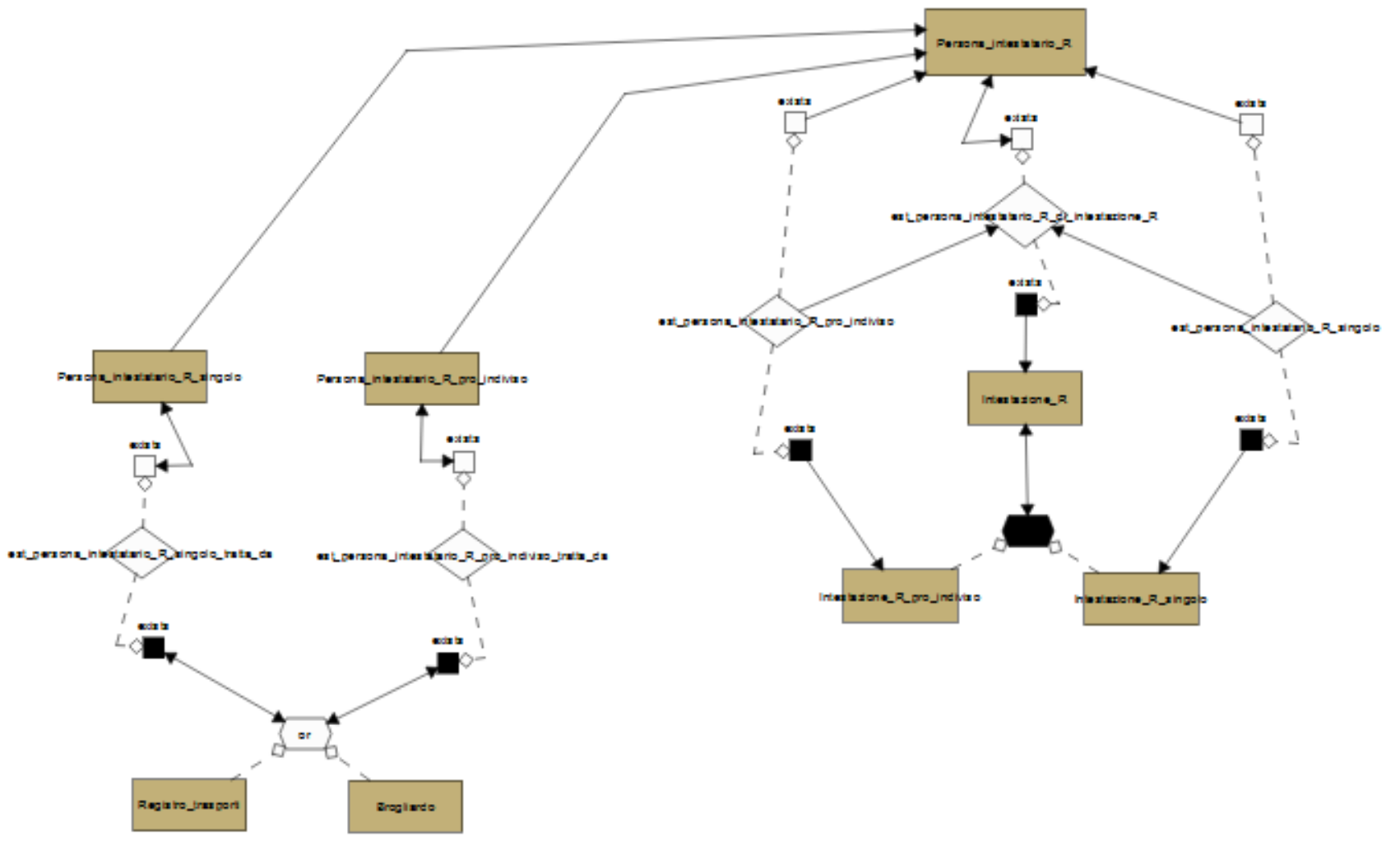

Fig. 30

Tipologicamente circoscritte risultano essere anche le fonti da cui desumere se una certa *Intestazione* R sia associata ad una *Persona Intestatario* R separatamente - *Persona Intestatario* R *padrone singolo* - o collettivamente - *Persona Intestatario* R *pro indiviso: Brogliardo* e *Registro dei trasporti.*

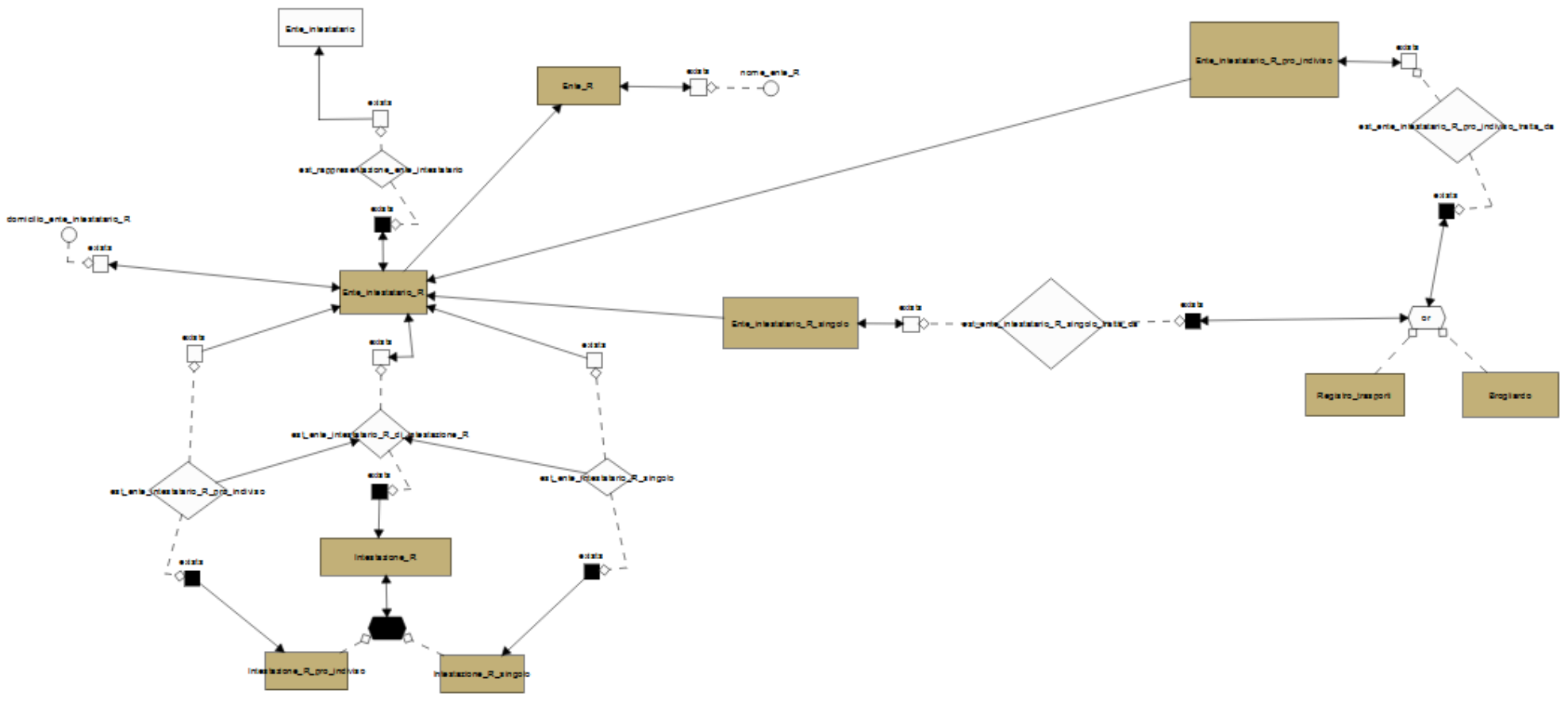

Fig.31

Il discorso fatto per la Persona Intestatario R, singolo o pro indiviso (Fig.30), risulta parimenti valido nel caso in cui l'intestatario sia un ente. La classe *Ente intestatario* R, sottoclasse di *Ente* R, ne eredita la data property relativa al nome. Per un *Ente intestatario* R avremo sempre contezza del domicilio dal momento che la denominazione e l'ubicazione vengono prescritti, e di fatto riportati nella documentazione, come identificatori. Si ribadisce, che analogamente ad altre entità, anche gli enti intestatari non hanno tutti evidenza documentaria.

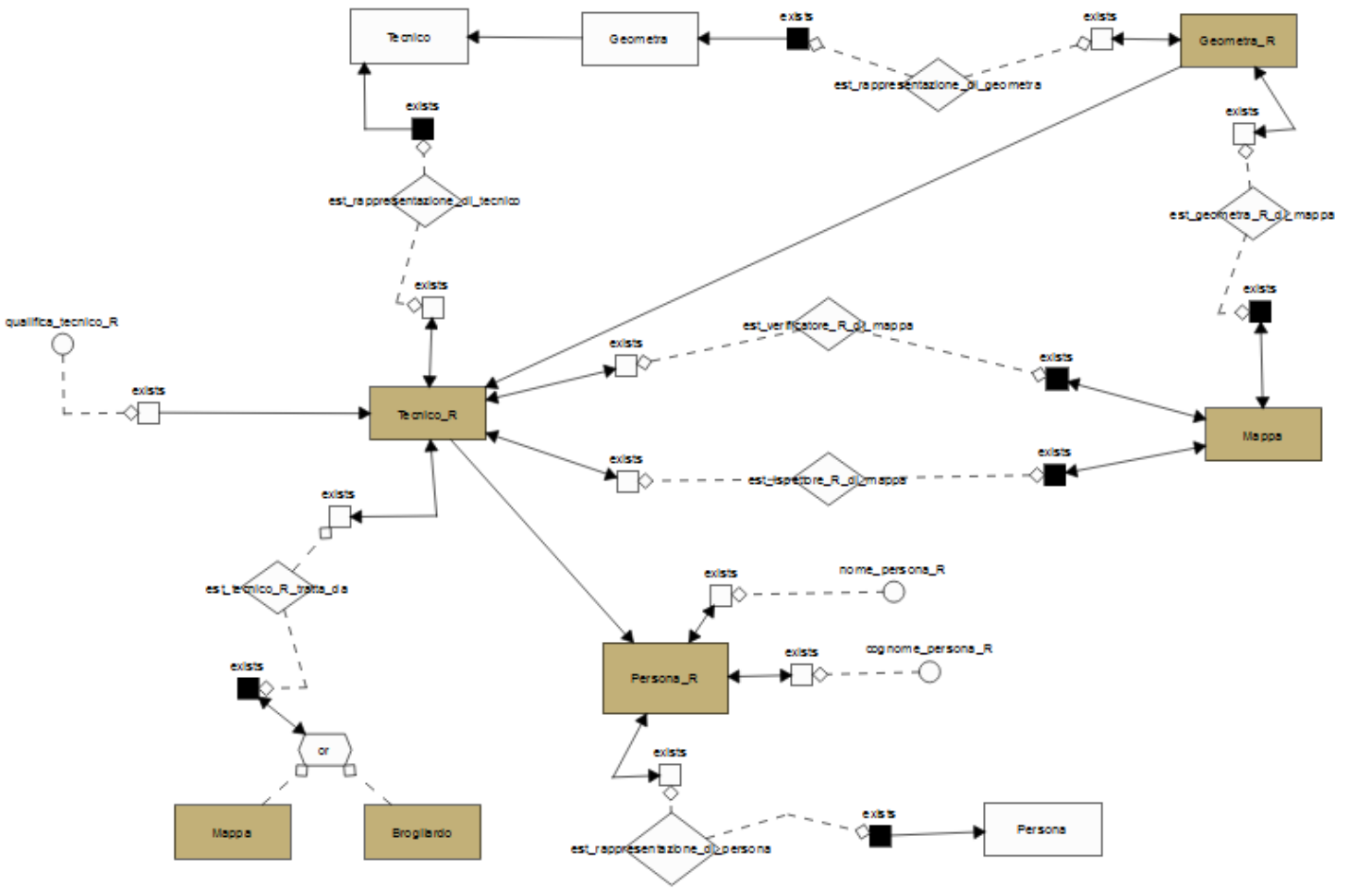

Fig. 32

La penultima porzione del diagramma, adibita alla rappresentazione dei tecnici, di coloro che concretamente procedettero all'elevazione, alla verifica e all'aggiornamento delle mappe, vede la classe *Tecnico* R nel ruolo di sottoclasse di una generica entità *Persona* R - di cui eredita le data properties - e di superclasse per *Geometra* R. La data property inerente la qualifica, è stata inserita per consentire di specificare se si è di fronte ad un ingegnere ispettore o per esempio ad un ingegnere verificatore. Nome, cognome e qualifica si desumono dalla *Mappa* - in particolare, dalle sottoscrizioni su di essa vergate - e dal corrispondente *Brogliardo*. Si è già detto della non perfetta equivalenza tra quanto presente nei documenti archivistici e il mondo reale. La porzione conclusiva (Fig.33), che si riaggancia a quella in Fig. 32, mostra coloro che coadiuvarono nella realizzazione delle rappresentazioni cartografiche. In genere, persone che risiedevano nel territorio in cui avevano luogo le operazioni di misura e poi di stima, incaricati e retribuiti dal Comune: *Assistente* R, *Indicatore* R e *Aiutante* R (sottoclasse di *Persona* R pertanto e non di *Tecnico* R) Dal momento che, come da normativa d'altronde, le mappe recano in calce anche le loro firme, autografe, si è valutata come migliore una concettualizzazione che permettesse di restituirne le identità, mediante la reificazione.

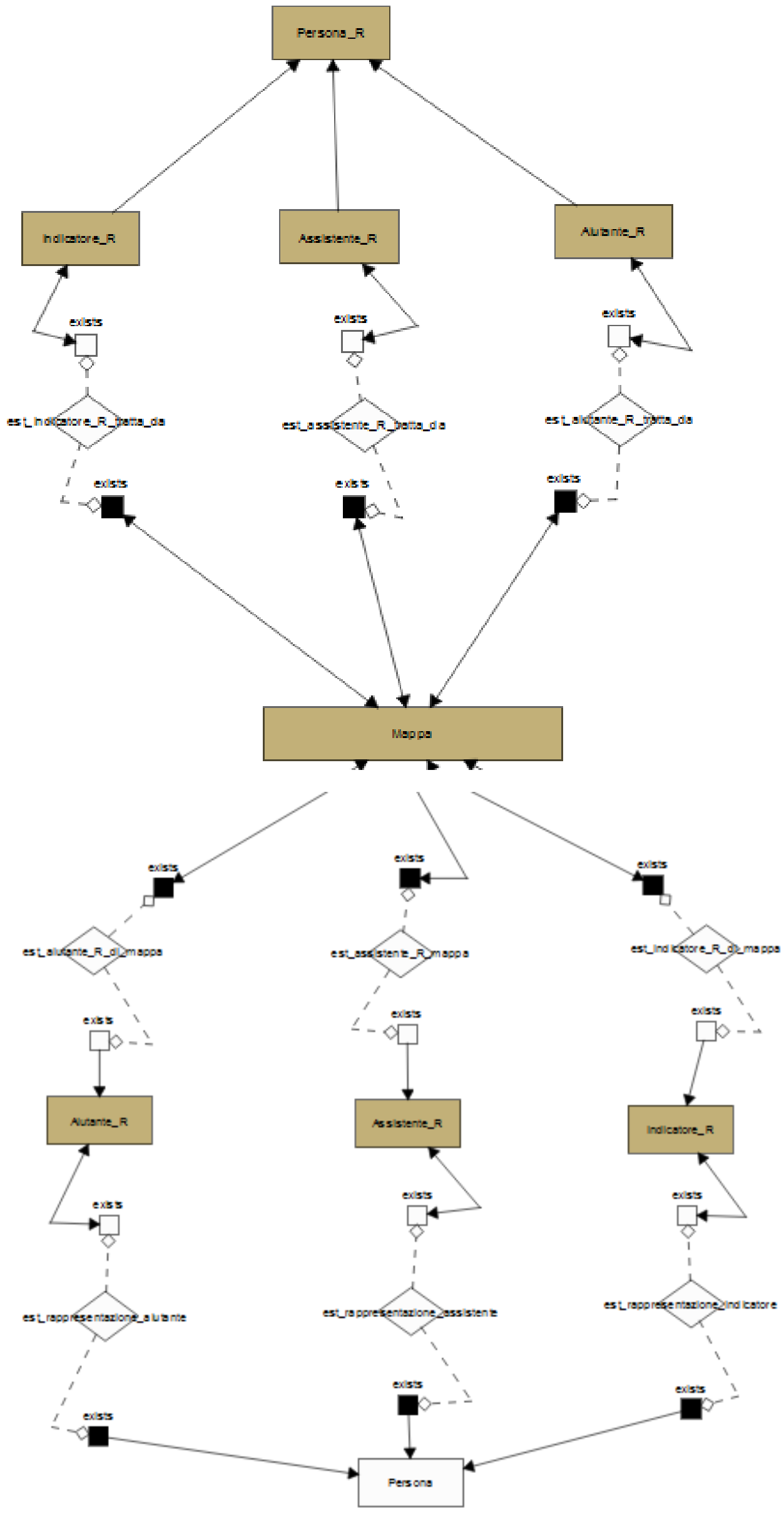

Fig. 33

## 3. Livello estensionale e possibili percorsi informativi

Prima di passare all'esperienza informativa che tenteremo di concretizzare in chiusura di capitolo, risulta opportuno concentrare l'attenzione sulle fonti archivistiche dalle quali sono stati desunti i dati immessi nel livello estensionale dell'ontologia - in corrispondenza delle corrispondenti classi e proprietà - sfruttando, uno dei molti servizi offerti dall'editor Protégé. Anzitutto, il fondo inventariato: *Catasti pontifici di Tivoli e suo territorio* o *Cancelleria del censo di Tivoli*. Dopo aver circoscritto il raggio d'azione alla sola città di Tivoli, si è attinto, per le scritture catastali, alle sottoserie:

1) *Istanze di Voltura*, 1819 - 1858 (3907 - 3948);

2) *Catastini, Rustico*1835 (3564 - 3567);
3) *Trasporti*, *Rustico*, 1835 (3568 - 3571);
4) *Catastini*, *Urbano*, 1835 (3582, 3583);
5) *Trasporti*, *Urbano*,1835 - 1871 (3584 - 3587).

In seconda battuta, per le mappe e i relativi brogliardi, alle digitalizzazioni del Progetto Imago II[14] del materiale contenuto in *Presidenza generale del censo*, *Archivio delle mappe e carte censuarie, Catasto Gregoriano, Comarca di Roma, Tivoli* (1819)[15]. In forza di quanto disposto dall'art. 125 del *Regolamento sulla misura dei terreni e formazione delle mappe* - «quando il territorio sarà molto esteso dovrà dividersi in diverse sezioni più regolari che sia possibile» -, il suddetto comune risulta ripartito in dieci sezioni catastali[16] cui si perviene, partendo dal livello indicato e scendendo fino all'unità archivistica desiderata. Da questa è possibile avere accesso alle unità documentarie contenute: mappa e corrispondente brogliardo; per la *sez. I- Città di Tivoli*, si hanno Mappa (la cui segnatura ha come specifica finale la dicitura *mappa 140*) e Brogliardo (con al termine *brogliardo 140*).[17] Dalle intestazioni ricercate e qui disposte in ordine alfabetico, si traggono sia dati di natura anagrafica relativi al possessore (tipicamente nome, cognome, paternità, domicilio, coniugio) che descrittivi e amministrativi inerenti, quindi le particelle (la mappa o sezione, il numero della particella principale e subordinato, la contrada o vocabolo in cui è ubicata, la superficie - espressa in quadrati, tavole, centesimi -, l'estimo - in scudi e baiocchi -, il tipo di coltivazione) che, nel corso del tempo, gli siano - a vario titolo - appartenute. Per favorire la comprensione delle notizie raccolte in relazione al modello ontologico (vd. § 4), e accrescerne l'utilità, preciseremo quali sono le informazioni rinvenute in ciascuna delle tipologie documentarie visionate.

- Liberatore Carlo. Dal registro dei trasporti (*Trasporti*, vol. I, 3568, nn. d'ordine della mutazione 384-385) veniamo a conoscenza del fatto che abbia venduto a Genga Francesco, figlio di Genga Pietro Paolo, una particella (mappale principale: 694 - subalterno: 1,2) adibita a *semina*[18] e situata in contrada Paterno (sezione 8ª, San Marco) con un'estensione pari a 1.0.00 (quadrati, tavole, centesimi) ed un estimo di 31 scudi e 32 baiocchi. L'istanza di voltura con cui avviene il passaggio è datata 10 aprile 1836 (numero 4176). Da questa

[14] http://www.cflr.beniculturali.it/index.html.
[15]http://ricerca.archiviodistatoroma.beniculturali.it/OpacASRoma/inventario/IT-ASROMA-AS1388 0000000#n
[16] Sezione I – Città di Tivoli; sezione II – Fosse; sezione III – Ponte de' prati; sezione IV – Martellona; sezione V – Villa Adriana; sezione VI – Quintiliolo; sezione VII – Sterpare; sezione VII – San Marco; sezione IX – Carciano; sezione X – Aurora.
[17] Il numero, nel caso specifico 140, costituisce parte della segnatura dell'unità digitalizzata. Sarà possibile consultare la mappa urbana di Tivoli, per rimanere sul nostro esempio, attraverso l'iter: *Imago*, *Catasto Gregoriano*,*Comarca*,(Imagohttp://www.cflr.beniculturali.it/comarca/sfoglia_comarca.php?Path=comarca&r=&lar=1 366&alt=768), COMARCA 140.
[18] *Regolamento sulla misura dei terreni e formazione delle mappe*, art.163: «Se il terreno sarà destinato alla coltivazione de' grani, e marzatelli, si chiamerà *Terreno seminativo*».

(*Istanze di Voltura*, 3924) riceviamo conferma di quanto scorto; in aggiunta ci si dice che Genga è domiciliato a Tivoli, che la particella - di cui è divenuto proprietario con *instrumentum* del 24 settembre 1835 rogato dal notaio Vincenzo Castrucci di Tivoli - apparteneva a Liberatore Carlo e a Geltrude Mazza già vedova Formica, coniugi, domiciliati a Tivoli. Ultimo riscontro, i *Catastini* (*Rustico*, vol. III, 3566; il numero di pagina, 1362, si deduce dal registro dei trasporti) da cui scopriamo che Carlo, figlio di Antonio, a quest'altezza cronologica risulta intestatario anche di un altro bene immobile (sezione 2ª, mappale principale: 163). Come detto in precedenza, se volessimo toccare con mano la mappa e/o il brogliardo, avremo bisogno della seguente segnatura: *Presidenza generale del censo 1816 - 1870*, *Archivio delle mappe e carte censuarie, Catasto Gregoriano, Comarca di Roma, Tivoli* (1819), *Sezione VIII - San Marco;* in alternativa, per la sola mappa, basterà accedere ai servizi digitali dal portale dell'Archivio di Stato di Roma e selezionare l'oggetto identificato dalla dicitura COMARCA 147. La Mappa di *San Marco*, «elevata dal giorno 21 Maggio al giorno 30 Giugno Anno 1819 dal geometra censuario Lucca Marconi, sotto la direzione del Signor Ingegnere Ispettore Pietro Locatelli», viene sottoscritta da Marconi e da: Raffaelle Guarmani Aiutante Censuario, Nicolao di Bendetti Assistente Comunale, Carlo Penga Indicatore.

- Masci Agostino. Dal registro dei trasporti (*Trasporti*, vol. I, 3568, nn. d'ordine della mutazione 376-377) veniamo a conoscenza del fatto che Agostino, figlio di Bernardo, abbia venduto a Coccanari Domenico, figlio di Tarquinio, una particella adibita a *orto*[19], situata in contrada Veste (sezione 1ª, mappale 809), con un'estensione pari a 56 centesimi e un estimo di 35 scudi e 35 baiocchi; l'istanza di voltura corrispondente è datata 2 aprile 1836 (numero 4171). Da questa (*Istanze di Voltura*, 3924) riceviamo conferma di quanto accertato finora; in aggiunta ci viene detto che tanto il vecchio quanto il nuovo possessore risultano domiciliati a Tivoli e che Coccanari è divenuto proprietario del terreno con *instrumentum* del 26 febbraio 1836 rogato dal notaio Vincenzo Castrucci di Tivoli. Ultimo riscontro, i *Catastini* (*Rustico*, vol. III, 3566; il numero di pagina, 1464, si deduce dal registro dei trasporti) dai quali scopriamo che Agostino è intestatario di altri tre beni immobili (due nelle sezione 1ª, mappali: 1122, 1124 e un terzo nella sezione 2ª, Fosse, numero principale di mappa: 189). Se volessimo toccare con mano la mappa e/o il brogliardo avremo bisogno della seguente segnatura: *Presidenza generale del censo 1816 - 1870*, *Archivio delle mappe e carte*

[19] *Ibid.* :«Se il terreno è coltivato a delizia in grande, si chiama *Villa*, se coltivato a delizia in piccolo, *Giardino*; se coltivato ad ortaglia, che serva al commercio, *Orto*, coll'avvertenza che trovandosi questa coltivazione lungi dai luoghi popolati, e servendo al solo uso del Colono, il terreno sarà considerato della stessa qualità di coltivazione in cui si trova il terreno attiguo, e perciò non gli si darà il nome di Orto».

*censuarie, Catasto Gregoriano, Comarca di Roma, Tivoli* (1819), *Sezione I – Città di Tivoli;* in alternativa, questa volta tanto per la mappa che per il brogliardo[20], basterà accedere, come si è fatto, ai servizi digitali dal portale dell'Archivio di Stato di Roma e selezionare l'oggetto identificato dalla dicitura COMARCA 140. Unica discrasia riportata nel brogliardo e confermata dal catastino, concerne la paternità: non Bernardo ma Bernardino. La Mappa urbana di Tivoli, «elevata dal giorno 29 Novembre 1819 al giorno 8 Marzo 1820 dal geometra censuario Cajo Marconi, sotto la direzione del Signor Ingegnere Ispettore Pietro Locatelli», viene sottoscritta da Marconi e da: Raffaele Guarmani Aiutante del Catasto e Carlo Penga Assistente e Indicatore Comunitativo.

- Paganelli Pietro, De Angelis Luigi e Paganelli Pasqua in De' Angelis. (Dal registro dei trasporti (*Trasporti*, vol. I, 3568, nn. d'ordine della mutazione 157-159) veniamo a conoscenza del fatto che Pietro, figlio di Francesco, abbia donato alla figlia e a De Angelis Luigi quattro particelle (con estimo e superficie complessivi pari a 11.91 - quadrati e tavole - e 412 scudi, 88 baiocchi) nel modo che segue: a Pasqua, una particella adibita a orto (sezione 8ª, mappale principale: 888, subalterni 1 e 2), enfiteusi di San Giovanni in Laterano, ubicata in contrada detta Strada Romana; a Luigi tre appezzamenti adibiti a vigna, oliveto e seminativo (sezione 8ª, mappale 23, Vigna, Vocabolo Paterno; sezione 10ª, mappali 247 e 361, vocabolo Pussiano). L'istanza di voltura corrispondente è datata 12 novembre 1835 (numero 4082). Da questa (*Istanze di Voltura*, 3924) riceviamo conferma di quanto accertato finora; in aggiunta ci viene indicata l'estensione e il valore di ciascun bene (al pari del registro dei trasporti), che i tre possessori risultano domiciliati a Tivoli, che Luigi è figlio di un tal Vincenzo, che l'oliveto con terreno annesso costituisce enfiteusi di Fernandez de' Velasco di Bernando o Bernardino. Infine, che con *instrumentum* del 7 febbraio 1835, rogato dal notaio Pietro Serbucci, Pietro abbia donato a Luigi «in ragion di dote» e a Pasqua «per titolo di donazione *inter vivos*». Ultimo riscontro, i *Catastini* (*Rustico*, vol. III, 3566; i numeri di pagina, 1757, 1758, 1759 si deducono dal registro dei trasporti) da cui scopriamo che Pietro è intestatario di una serie di beni: sez. I (mappale 1099), sez. V (mappali: 145, 208), sez. VIII (mappali: 17,18, 23, 702, 707, 888, 919, 927, 928, 999, 1066), sez. X (mappali: 181, 207, 208, 215, 230, 247, 261, 361). Della mappa della sez. VIII e delle modalità di consultazione - cartacea e digitale - di cartografia e brogliardi si è già detto; aggiungiamo per quella relativa alla sez. X, *Aurora* che «elevata dal giorno 1° Novembre al giorno 16 Decembre 1819 dal geometra censuario Lucca Marconi, sotto la direzione del Signor

[20] http://www.cflr.beniculturali.it/Gregoriano/brogliardi.php?lar=1366&alt=768.

Ingegnere Ispettore Pietro Locatelli», viene sottoscritta da Marconi e da: Giovanni Battista Massadaja Aiutante Censuario e Nicolao di Benedetti Assistente Comunitativo e Indicatore.

Quel che è stato fatto finora, è stato illustrare la *provenance* del contenuto informativo introdotto nella knowledge base, il "dove"; discuteremo, invece, ora - sinteticamente, e proponendo, per motivi di equilibrio e coerenza, un solo esempio (Masci Agostino) - il "come". *Conditio sine qua non* dell'accesso e dell'interrogazione dell'ontologia, risulta essere la traduzione in linguaggio formale - e qui entra in campo Protégé - degli oggetti documentari e dei dati in essi contenuti, delineati poc'anzi in linguaggio naturale. Per effettuare questa trasposizione, gli step da seguire sono, essenzialmente, tre: individuare la classe in cui introdurre l'istanza, aggiungerla al concetto di appartenenza associandole un identificatore, anche arbitrario, (per *Persona intestatario*, pers1), descrivere l'individuo in termini di object properties e data properties. Ricordando quanto specificato nel livello intensionale (vd. § 4), nello specifico in quelle porzioni adibite alle rappresentazioni, ovvero alle entità di cui si ha evidenza documentaria, avremo per Masci Agostino, ascrivibile alla classe *Persona Intestatario R padrone diretto*, anzitutto, le data properties, ereditate da *Persona R* (di cui *Persona Intestatario* è sottoclasse), "nome_persona_R" e "cognome_persona_R", cui si aggiungerà il domicilio "domicilio_persona_intestatario_R". Masci sarà legato, poi, dalla object property "persona_intestatario_padrone_diretto_R_ha_intestazione" alla *Intestazione_R*, che, a sua volta, partecipa con *Rappresentazione particella* alla relazione "rappresentazione_particella_ha_intestazione_R". La *Rappresentazione particella*, ovvero l'immobile ascrivibile a Masci - individuo della suddetta classe, identificato dalla sigla part1 - contemplerà più data properties, il "mappale_principale" (il cui valore sarà un intero: 809), ma anche il nome della contrada in cui il bene è ubicato (il valore corrispondente sarà qui un *literal*: Veste), l2 "estimo_di_particella" (il cui valore sarà una stringa: 35. 35 scudi e baiocchi) e sarà impegnata in una serie di object properties. Particolarmente importante, quella che la vincola alla classe *Sezione catastale*, vale a dire "Sezione_catastale_contiene_rappresentazione_particella". L'istanza introdotta per *Sezione catastale* - sez1, nel nostro caso - è legata ad un istanza della classe *Mappa* - mappa1 - dalla relazione "sezione_catastale_est_rappresentata_graficamente"; ogni *Mappa* rappresenta graficamente, infatti, una ed una sola *Sezione catastale*. Tra le data properties di mappa 1, rilevante "segnatura_unità_di_descrizione", ereditata dalla classe *Unità di descrizione*, di cui Mappa è *sottoclasse*, che ne consentirà il reperimento all'interno dell'Archivio di Stato di Roma. Confidando di essere stati sufficientemente chiari nell'esposizione, non ci dilunghiamo oltre sul "dove" e "come", ma lasciamo spazio al "perché": perché ipotizzare delle ricerche lato utente. Al quesito risponde, *in primis*, con acribia critica, S. Vitali, nel momento in cui afferma che[21]:

> […] L'archivista, oltre a costruire elaborati alberi, dovrebbe ogni tanto porsi domande del tipo: sto fornendo all'utente tutti gli elementi di conoscenza che possono essergli utili per orientarsi nella documentazione che sto descrivendo? Sto dando qualcosa per scontato? Sto facendo ricorso ad un linguaggio appropriato e comprensibile? Sto dando informazioni realmente pertinenti e utili per chi vuole fare ricerca nell'archivio?

Si è, poi, allargata la visuale alle problematiche «linguistiche, logico-sintattiche e di sistema»[22] riscontrabili nella restituzione nella restituzione di contenuti archivistici digitali, e, per converso, alle potenzialità, che promettono, sotto l'aspetto della qualità dell'accesso e dell'uso le tecnologie del Semantic Web, segnatamente gli strumenti ontologici. Muovendo da tali premesse, abbiamo cercato di immaginare l'effettiva esperienza di un eventuale fruitore della base di conoscenza e di valutarne l'esito. Gli obiettivi informativi che ci si è posti e per i quali sono state realizzate e valutate nel tab di Protégé "SPARQL Query" due interrogazioni - immedesimandosi, nel primo caso, nei panni di chi vi acceda essendo a conoscenza soltanto di pochi dati essenziali, per esempio nome e cognome o anche soltanto uno dei due, dell'intestatario di un bene[23], nel secondo[24] in quelli di chi avendo cognizione del mappale di una particella desideri conoscere la mappa in cui è contenuta, e di concerto, per esempio, la segnatura o l'esistenza di una versione digitalizzata per poterla consultare - sembrano non solo confermare il ruolo strategico che l'utilizzo di ontologie potrebbe giocare in ambito archivistico ma anche i benefici che derivano dall'adozione della loro espressività nel definire i termini e l'organizzazione di un dominio di conoscenza.

## 4. Appendice

*File OWL: TBox*

```
Prefix(owl:=<http://www.w3.org/2002/07/owl#>)
Prefix(rdf:=<http://www.w3.org/1999/02/22-rdf-syntax-ns#>)
Prefix(xml:=<http://www.w3.org/XML/1998/namespace>)
```

---

[21] S. Vitali, *La descrizione degli archivi nell'epoca degli standard e dei sistemi informatici*, in *Archivistica. Teorie, metodi, pratiche*, (a cura di) L. Giuva e M. Guercio, Roma, Carocci, 2014, pp. 209-210.

[22] P. Feliciati, *Dall'inventario alla descrizione degli archivi in ambiente digitale: si possono offrire agli utenti risorse efficaci?*, in "Elaborare il sapere nell'era digitale", Montevarchi (Italia), 22-23 novembre 2007, p. 10.

[23] PREFIX xsd:<http://www.w3.org/2001/XMLSchema#>
PREFIX :<http://modeus.uniroma1.it/ontology#>
SELECT ?particella WHERE {?i :est_persona_intestatario_R_padrone_diretto ?particella. ?i :cognome_persona_R 'Poggi'^^xsd:string.

[24] PREFIX : <http://modeus.uniroma1.it/ontology#>
SELECT ?mappa WHERE {?sez :sezione_catastale_contiene_rappresentazione_particella :part1. ?sez :sezione_catastale_est_rappresentata_graficamente ?mappa}

```
Prefix(xsd:=<http://www.w3.org/2001/XMLSchema#>)
Prefix(rdfs:=<http://www.w3.org/2000/01/rdf-schema#>)
Prefix(modeus:=<http://modeus.uniroma1.it/ontology#>)
Ontology(<http://modeus.uniroma1.it/ontology>
Declaration(Class(modeus:Acqua))
Declaration(Class(modeus:Agente))
Declaration(Class(modeus:Aiutante_R))
Declaration(Class(modeus:Assistente_R))
Declaration(Class(modeus:Atto_notarile))
Declaration(Class(modeus:Brogliardo))
Declaration(Class(modeus:Brogliardo_digitalizzato))
Declaration(Class(modeus:Catastino))
Declaration(Class(modeus:Circoscrizione))
Declaration(Class(modeus:Co_mune_sede_di_ufficio_di_registro))
Declaration(Class(modeus:Componente_unità_documentaria))
Declaration(Class(modeus:Comune))
Declaration(Class(modeus:Copia_brogliardo))
Declaration(Class(modeus:Delegazione))
Declaration(Class(modeus:Destinazione))
Declaration(Class(modeus:Ente))
Declaration(Class(modeus:Ente_R))
Declaration(Class(modeus:Ente_intestatario))
Declaration(Class(modeus:Ente_intestatario_R))
Declaration(Class(modeus:Ente_intestatario_R_padrone_diretto))
Declaration(Class(modeus:Ente_intestatario_R_pro_indiviso))
Declaration(Class(modeus:Ente_intestatario_R_singolo))
Declaration(Class(modeus:Ente_intestatario_R_utilista))
Declaration(Class(modeus:Famiglia))
Declaration(Class(modeus:Fondo))
Declaration(Class(modeus:Funzione))
Declaration(Class(modeus:Geometra))
Declaration(Class(modeus:Geometra_R))
Declaration(Class(modeus:Governo_distrettuale))
Declaration(Class(modeus:Indicatore_R))
Declaration(Class(modeus:Intestazione))
Declaration(Class(modeus:Intestazione_R))
Declaration(Class(modeus:Intestazione_R_pro_indiviso))
Declaration(Class(modeus:Intestazione_R_singolo))
Declaration(Class(modeus:Intestazione_pro_indiviso))
Declaration(Class(modeus:Intestazione_singolo))
Declaration(Class(modeus:Isituto_conservatore))
Declaration(Class(modeus:Istanza_voltura))
Declaration(Class(modeus:Legazione))
Declaration(Class(modeus:Livello_di_descrizione))
Declaration(Class(modeus:Mappa))
Declaration(Class(modeus:Mappa_copia_scala_originale))
Declaration(Class(modeus:Mappa_copia_scala_originale_digitalizzata))
Declaration(Class(modeus:Mappa_digitalizzata))
Declaration(Class(modeus:Mappetta_carta))
Declaration(Class(modeus:Mappetta_tela))
Declaration(Class(modeus:Notaio))
Declaration(Class(modeus:Oggetto_fisico))
Declaration(Class(modeus:Organo_periferico))
Declaration(Class(modeus:Persona))
Declaration(Class(modeus:Persona_R))
Declaration(Class(modeus:Persona_intestatario))
Declaration(Class(modeus:Persona_intestatario_R))
Declaration(Class(modeus:Persona_intestatario_R_padrone_diretto))
Declaration(Class(modeus:Persona_intestatario_R_pro_indiviso))
Declaration(Class(modeus:Persona_intestatario_R_singolo))
Declaration(Class(modeus:Persona_intestatario_R_utilista))
Declaration(Class(modeus:Persona_intestatario_moglie_R))
```

Declaration(Class(modeus:Persona_intestatario_vedova_R))
Declaration(Class(modeus:Rappresentazione_cartacea_mappa))
Declaration(Class(modeus:Rappresentazione_particella))
Declaration(Class(modeus:Registro_istanza_voltura))
Declaration(Class(modeus:Registro_trasporti))
Declaration(Class(modeus:Relazione_persona_intestatario_R))
Declaration(Class(modeus:Scrittura_impianto_primaria))
Declaration(Class(modeus:Scrittura_secondaria))
Declaration(Class(modeus:Sede))
Declaration(Class(modeus:Serie))
Declaration(Class(modeus:Sezione_catastale))
Declaration(Class(modeus:Soggetto_produttore))
Declaration(Class(modeus:Sottoserie))
Declaration(Class(modeus:Stato))
Declaration(Class(modeus:Strada))
Declaration(Class(modeus:Sub_fondo))
Declaration(Class(modeus:Successione_intestazione_R))
Declaration(Class(modeus:Tecnico))
Declaration(Class(modeus:Tecnico_R))
Declaration(Class(modeus:Unità_archivistica))
Declaration(Class(modeus:Unità_di_descrizione))
Declaration(Class(modeus:Unità_documentaria))
Declaration(Class(modeus:destinazione_produttiva))
Declaration(Class(modeus:destinazione_uso))
Declaration(ObjectProperty(modeus:agente_est_soggetto_produttore))
Declaration(ObjectProperty(modeus:atto_notarile_est_redatto))
Declaration(ObjectProperty(modeus:brogliardo_est_riprodotto))
Declaration(ObjectProperty(modeus:brogliardo_ha_riproduzione_digitale))
Declaration(ObjectProperty(modeus:componente_documentaria_est_parte_di_unità))
Declaration(ObjectProperty(modeus:comune_appartiene_a))
Declaration(ObjectProperty(modeus:comune_comprende))
Declaration(ObjectProperty(modeus:confina_con_delegazione))
Declaration(ObjectProperty(modeus:copia_brogliardo_est_destinata_a))
Declaration(ObjectProperty(modeus:delegazione_confina_con))
Declaration(ObjectProperty(modeus:delegazione_confina_con_Stato))
Declaration(ObjectProperty(modeus:delegazione_est_divisa))
Declaration(ObjectProperty(modeus:delegazione_est_inclusa))
Declaration(ObjectProperty(modeus:est_aiutante_R_di_mappa))
Declaration(ObjectProperty(modeus:est_aiutante_R_tratta_da))
Declaration(ObjectProperty(modeus:est_aiutante_di_mappa))
Declaration(ObjectProperty(modeus:est_assistente_R_mappa))
Declaration(ObjectProperty(modeus:est_assistente_R_tratta_da))
Declaration(ObjectProperty(modeus:est_assistente_di_mappa))
Declaration(ObjectProperty(modeus:est_brogliardo_di_scrittura_primaria))
Declaration(ObjectProperty(modeus:est_ente_intestatario_R_di_intestazione_R))
Declaration(ObjectProperty(modeus:est_ente_intestatario_R_padrone_diretto))
Declaration(ObjectProperty(modeus:est_ente_intestatario_R_padrone_diretto_tratta_da))
Declaration(ObjectProperty(modeus:est_ente_intestatario_R_pro_indiviso))
Declaration(ObjectProperty(modeus:est_ente_intestatario_R_pro_indiviso_tratta_da))
Declaration(ObjectProperty(modeus:est_ente_intestatario_R_singolo))
Declaration(ObjectProperty(modeus:est_ente_intestatario_R_singolo_tratta_da))
Declaration(ObjectProperty(modeus:est_ente_intestatario_R_utilista))
Declaration(ObjectProperty(modeus:est_ente_intestatario_R_utilista_tratta_da))
Declaration(ObjectProperty(modeus:est_ente_intestatario_di_intestazione))
Declaration(ObjectProperty(modeus:est_ente_intestatario_padrone_diretto))
Declaration(ObjectProperty(modeus:est_ente_intestatario_pro_indiviso))
Declaration(ObjectProperty(modeus:est_ente_intestatario_singolo))
Declaration(ObjectProperty(modeus:est_ente_intestatario_utilista))
Declaration(ObjectProperty(modeus:est_figlio))
Declaration(ObjectProperty(modeus:est_geometra_R_di_mappa))
Declaration(ObjectProperty(modeus:est_geometra_di_mappa))
Declaration(ObjectProperty(modeus:est_indicatore_R_di_mappa))

Declaration(ObjectProperty(modeus:est_indicatore_R_tratta_da))
Declaration(ObjectProperty(modeus:est_indicatore_di_mappa))
Declaration(ObjectProperty(modeus:est_intestazione_R_precedente))
Declaration(ObjectProperty(modeus:est_intestazione_R_successiva))
Declaration(ObjectProperty(modeus:est_intestazione_R_tratta_da))
Declaration(ObjectProperty(modeus:est_ispettore_R_di_mappa))
Declaration(ObjectProperty(modeus:est_ispettore_di_mappa))
Declaration(ObjectProperty(modeus:est_mappa_di_scrittura_primaria))
Declaration(ObjectProperty(modeus:est_mappetta_tela_di_scrittura_primaria))
Declaration(ObjectProperty(modeus:est_marito_di_moglie))
Declaration(ObjectProperty(modeus:est_marito_di_vedova))
Declaration(ObjectProperty(modeus:est_moglie))
Declaration(ObjectProperty(modeus:est_padre))
Declaration(ObjectProperty(modeus:est_parte_di_unità_di_descrizione))
Declaration(ObjectProperty(modeus:est_persona_intestatario_R_di_intestazione_R))
Declaration(ObjectProperty(modeus:est_persona_intestatario_R_padrone_diretto))
Declaration(ObjectProperty(modeus:est_persona_intestatario_R_padrone_diretto_tratta_da))
Declaration(ObjectProperty(modeus:est_persona_intestatario_R_pro_indiviso))
Declaration(ObjectProperty(modeus:est_persona_intestatario_R_pro_indiviso_tratta_da))
Declaration(ObjectProperty(modeus:est_persona_intestatario_R_singolo))
Declaration(ObjectProperty(modeus:est_persona_intestatario_R_singolo_tratta_da))
Declaration(ObjectProperty(modeus:est_persona_intestatario_R_utilista))
Declaration(ObjectProperty(modeus:est_persona_intestatario_R_utilista_tratta_da))
Declaration(ObjectProperty(modeus:est_persona_intestatario_di_intestazione))
Declaration(ObjectProperty(modeus:est_persona_intestatario_padrone_diretto))
Declaration(ObjectProperty(modeus:est_persona_intestatario_pro_indiviso))
Declaration(ObjectProperty(modeus:est_persona_intestatario_singolo))
Declaration(ObjectProperty(modeus:est_persona_intestatario_utilista))
Declaration(ObjectProperty(modeus:est_rappresentazione_aiutante))
Declaration(ObjectProperty(modeus:est_rappresentazione_assistente))
Declaration(ObjectProperty(modeus:est_rappresentazione_di_geometra))
Declaration(ObjectProperty(modeus:est_rappresentazione_di_intestazione))
Declaration(ObjectProperty(modeus:est_rappresentazione_di_persona))
Declaration(ObjectProperty(modeus:est_rappresentazione_di_persona_intestatario))
Declaration(ObjectProperty(modeus:est_rappresentazione_di_tecnico))
Declaration(ObjectProperty(modeus:est_rappresentazione_ente_intestatario))
Declaration(ObjectProperty(modeus:est_rappresentazione_indicatore))
Declaration(ObjectProperty(modeus:est_relazione_persona_intestatario_R_tratta_da))
Declaration(ObjectProperty(modeus:est_soggetto_produttore_di))
Declaration(ObjectProperty(modeus:est_successione_intestazione_R_tratta_da))
Declaration(ObjectProperty(modeus:est_tecnico_R_tratta_da))
Declaration(ObjectProperty(modeus:est_unità_di_descrizione_successiva))
Declaration(ObjectProperty(modeus:est_vedova))
Declaration(ObjectProperty(modeus:est_verificatore_R_di_mappa))
Declaration(ObjectProperty(modeus:est_verificatore_di_mappa))
Declaration(ObjectProperty(modeus:funzione_est_documentata))
Declaration(ObjectProperty(modeus:istanza_est_registrata_in))
Declaration(ObjectProperty(modeus:istanza_voltura_est_associata))
Declaration(ObjectProperty(modeus:istituto_conservatore_est_ubicato_fisicamente))
Declaration(ObjectProperty(modeus:mappa_copia_scala_originale_digitalizzata_est_legata_a_mappa_digitalizzata))
Declaration(ObjectProperty(modeus:mappa_copia_scala_originale_est_destinata_a))
Declaration(ObjectProperty(modeus:mappa_copia_scala_originale_ha_riproduzione_digitale))
Declaration(ObjectProperty(modeus:mappa_digitalizzata_est_descritta_da_brogliardo_digitalizzato))
Declaration(ObjectProperty(modeus:mappa_est_descritta_da_brogliardo))
Declaration(ObjectProperty(modeus:mappa_est_relativa))
Declaration(ObjectProperty(modeus:mappa_est_riprodotta_in))
Declaration(ObjectProperty(modeus:mappa_ha_riproduzione_digitale))
Declaration(ObjectProperty(modeus:mappetta_carta_est_destinata_a))
Declaration(ObjectProperty(modeus:organo_periferico_ha_competenza))
Declaration(ObjectProperty(modeus:organo_periferico_ha_sede))
Declaration(ObjectProperty(modeus:rappresentazione_particella_confina_con))
Declaration(ObjectProperty(modeus:rappresentazione_particella_confina_con_oggetto_fisico))

Declaration(ObjectProperty(modeus:rappresentazione_particella_est_descritta))
Declaration(ObjectProperty(modeus:rappresentazione_particella_est_frazionata_in))
Declaration(ObjectProperty(modeus:rappresentazione_particella_est_fusa_in))
Declaration(ObjectProperty(modeus:rappresentazione_particella_ha_destinazione))
Declaration(ObjectProperty(modeus:rappresentazione_particella_ha_intestazione))
Declaration(ObjectProperty(modeus:rappresentazione_particella_ha_intestazione_R))
Declaration(ObjectProperty(modeus:relazione_coniugio))
Declaration(ObjectProperty(modeus:relazione_paternità))
Declaration(ObjectProperty(modeus:sezione_catastale_contiene_oggetto_fisico))
Declaration(ObjectProperty(modeus:sezione_catastale_contiene_rappresentazione_particella))
Declaration(ObjectProperty(modeus:sezione_catastale_est_rappresentata_graficamente))
Declaration(ObjectProperty(modeus:soggetto_produttore_adempie_funzione))
Declaration(ObjectProperty(modeus:successione_intestazione))
Declaration(ObjectProperty(modeus:unità_di_descrizione_est_conservata))
Declaration(ObjectProperty(modeus:unità_di_descrizione_ha_livello))
Declaration(DataProperty(modeus:anno_aggiornamento_mappa_copia))
Declaration(DataProperty(modeus:anno_di_morte))
Declaration(DataProperty(modeus:anno_di_nascita))
Declaration(DataProperty(modeus:anno_fine_intestazione))
Declaration(DataProperty(modeus:anno_fine_intestazione_R))
Declaration(DataProperty(modeus:anno_fine_range_unità))
Declaration(DataProperty(modeus:anno_formazione_unità))
Declaration(DataProperty(modeus:anno_inizio_intestazione))
Declaration(DataProperty(modeus:anno_inizio_intestazione_R))
Declaration(DataProperty(modeus:anno_inizio_range_unità))
Declaration(DataProperty(modeus:anno_istituzione_ente))
Declaration(DataProperty(modeus:anno_morte_rappresentazione_particella))
Declaration(DataProperty(modeus:anno_nascita_rappresentazione_particella))
Declaration(DataProperty(modeus:anno_redazione_atto))
Declaration(DataProperty(modeus:anno_registrazione_atto))
Declaration(DataProperty(modeus:anno_soppressione_ente))
Declaration(DataProperty(modeus:città_sede))
Declaration(DataProperty(modeus:codice_identificativo_conservatore))
Declaration(DataProperty(modeus:codice_identificativo_descrizione_funzione))
Declaration(DataProperty(modeus:codice_identificativo_record_autorità))
Declaration(DataProperty(modeus:codice_identificativo_unità_descrizione))
Declaration(DataProperty(modeus:cognome_persona))
Declaration(DataProperty(modeus:cognome_persona_R))
Declaration(DataProperty(modeus:consistenza_unità_descrizione))
Declaration(DataProperty(modeus:domicilio_ente_intestatario))
Declaration(DataProperty(modeus:domicilio_ente_intestatario_R))
Declaration(DataProperty(modeus:domicilio_persona_intestatario))
Declaration(DataProperty(modeus:domicilio_persona_intestatario_R))
Declaration(DataProperty(modeus:estensione_di_particella))
Declaration(DataProperty(modeus:estimo_di_particella))
Declaration(DataProperty(modeus:forma_autorizzata_nome_funzione))
Declaration(DataProperty(modeus:forma_autorizzata_nome_istituto_conservatore))
Declaration(DataProperty(modeus:forma_autorizzata_nome_soggetto_produttore))
Declaration(DataProperty(modeus:giorno_aggiornamento_mappa_copia))
Declaration(DataProperty(modeus:giorno_di_morte))
Declaration(DataProperty(modeus:giorno_di_nascita))
Declaration(DataProperty(modeus:giorno_fine_intestazione))
Declaration(DataProperty(modeus:giorno_fine_intestazione_R))
Declaration(DataProperty(modeus:giorno_fine_range_unità))
Declaration(DataProperty(modeus:giorno_formazione_unità))
Declaration(DataProperty(modeus:giorno_inizio_intestazione))
Declaration(DataProperty(modeus:giorno_inizio_intestazione_R))
Declaration(DataProperty(modeus:giorno_inizio_range_unità))
Declaration(DataProperty(modeus:giorno_istituzione_ente))
Declaration(DataProperty(modeus:giorno_morte_rappresentazione_particella))
Declaration(DataProperty(modeus:giorno_nascita_rappresentazione_particella))
Declaration(DataProperty(modeus:giorno_redazione_atto))

Declaration(DataProperty(modeus:giorno_registrazione_atto))
Declaration(DataProperty(modeus:giorno_soppressione_ente))
Declaration(DataProperty(modeus:indirizzo_sede))
Declaration(DataProperty(modeus:indirizzo_sede_principale))
Declaration(DataProperty(modeus:indirizzo_sede_succursale))
Declaration(DataProperty(modeus:luogo_di_morte))
Declaration(DataProperty(modeus:luogo_di_nascita))
Declaration(DataProperty(modeus:mappale_principale_di_particella))
Declaration(DataProperty(modeus:mappale_subordinato_di_particella))
Declaration(DataProperty(modeus:mese_aggiornamento_mappa_copia))
Declaration(DataProperty(modeus:mese_di_morte))
Declaration(DataProperty(modeus:mese_di_nascita))
Declaration(DataProperty(modeus:mese_fine_intestazione))
Declaration(DataProperty(modeus:mese_fine_intestazione_R))
Declaration(DataProperty(modeus:mese_fine_range_unità))
Declaration(DataProperty(modeus:mese_formazione_unità))
Declaration(DataProperty(modeus:mese_inizio_intestazione))
Declaration(DataProperty(modeus:mese_inizio_intestazione_R))
Declaration(DataProperty(modeus:mese_inizio_range_unità))
Declaration(DataProperty(modeus:mese_istituzione_ente))
Declaration(DataProperty(modeus:mese_morte_rappresentazione_particella))
Declaration(DataProperty(modeus:mese_nascita_rappresentazione_particella))
Declaration(DataProperty(modeus:mese_redazione_atto))
Declaration(DataProperty(modeus:mese_registrazione_atto))
Declaration(DataProperty(modeus:mese_soppressione_ente))
Declaration(DataProperty(modeus:nome_acqua))
Declaration(DataProperty(modeus:nome_circoscrizione))
Declaration(DataProperty(modeus:nome_comune))
Declaration(DataProperty(modeus:nome_contrada_di_particella))
Declaration(DataProperty(modeus:nome_delegazione))
Declaration(DataProperty(modeus:nome_destinazione_produttiva))
Declaration(DataProperty(modeus:nome_destinazione_uso))
Declaration(DataProperty(modeus:nome_ente))
Declaration(DataProperty(modeus:nome_ente_R))
Declaration(DataProperty(modeus:nome_governo_distrettuale))
Declaration(DataProperty(modeus:nome_persona))
Declaration(DataProperty(modeus:nome_persona_R))
Declaration(DataProperty(modeus:nome_sezione_catastale))
Declaration(DataProperty(modeus:nome_stato))
Declaration(DataProperty(modeus:nome_strada))
Declaration(DataProperty(modeus:numero_mutazione_registro_trasporti))
Declaration(DataProperty(modeus:numero_pagina_catastino))
Declaration(DataProperty(modeus:numero_pagina_registro_trasporti))
Declaration(DataProperty(modeus:numero_progressivo_istanza_voltura))
Declaration(DataProperty(modeus:numero_protocollo_registrazione_istanza_voltura))
Declaration(DataProperty(modeus:numero_sezione_catastale))
Declaration(DataProperty(modeus:qualifica_tecnico_R))
Declaration(DataProperty(modeus:quantità_componente_documentaria))
Declaration(DataProperty(modeus:scala_mappa))
Declaration(DataProperty(modeus:scala_mappetta_carta))
Declaration(DataProperty(modeus:scala_mappetta_tela))
Declaration(DataProperty(modeus:segnatura_mappa_copia_originale_digitalizzata))
Declaration(DataProperty(modeus:segnatura_mappa_digitalizzata))
Declaration(DataProperty(modeus:segnatura_unità_descrizione))
Declaration(DataProperty(modeus:sito_web_istituto_conservatore))
Declaration(DataProperty(modeus:stato_unità_documentaria))
Declaration(DataProperty(modeus:supporto_unità_descrizione))
Declaration(DataProperty(modeus:tipo_acqua))
Declaration(DataProperty(modeus:tipo_strada))
Declaration(DataProperty(modeus:tipologia_componente_documentaria))
Declaration(DataProperty(modeus:tipologia_documentaria_unità_descrizione))
Declaration(DataProperty(modeus:tipologia_funzione))

Declaration(DataProperty(modeus:titolo_attribuito_unità_descrizione))
Declaration(DataProperty(modeus:titolo_originale_unità_descrizione))
Declaration(DataProperty(modeus:titolo_unità_descrizione))
Declaration(NamedIndividual(modeus:Archivio_di_Stato_Roma))
Declaration(NamedIndividual(modeus:Cancelleria_censo_Tivoli))
Declaration(NamedIndividual(modeus:Direzione_generale_catasti))
Declaration(NamedIndividual(modeus:Fondo))
Declaration(NamedIndividual(modeus:Presidenza_del_censo))
Declaration(NamedIndividual(modeus:Serie))
Declaration(NamedIndividual(modeus:Sottoserie))
Declaration(NamedIndividual(modeus:Sub_fondo))
Declaration(NamedIndividual(modeus:Unità_archivistica))
Declaration(NamedIndividual(modeus:Unità_documentaria))
ObjectPropertyDomain(modeus:agente_est_soggetto_produttore modeus:Agente)
ObjectPropertyRange(modeus:agente_est_soggetto_produttore modeus:Soggetto_produttore)
ObjectPropertyDomain(modeus:atto_notarile_est_redatto modeus:Atto_notarile)
ObjectPropertyRange(modeus:atto_notarile_est_redatto modeus:Notaio)
ObjectPropertyDomain(modeus:brogliardo_est_riprodotto modeus:Brogliardo)
ObjectPropertyRange(modeus:brogliardo_est_riprodotto modeus:Copia_brogliardo)
ObjectPropertyDomain(modeus:brogliardo_ha_riproduzione_digitale modeus:Brogliardo)
ObjectPropertyRange(modeus:brogliardo_ha_riproduzione_digitale modeus:Brogliardo_digitalizzato)
ObjectPropertyDomain(modeus:componente_documentaria_est_parte_di_unità
modeus:Componente_unità_documentaria)
ObjectPropertyRange(modeus:componente_documentaria_est_parte_di_unità modeus:Unità_documentaria)
ObjectPropertyDomain(modeus:comune_appartiene_a modeus:Comune)
ObjectPropertyRange(modeus:comune_appartiene_a modeus:Delegazione)
ObjectPropertyDomain(modeus:comune_comprende modeus:Comune)
ObjectPropertyRange(modeus:comune_comprende modeus:Circoscrizione)
SubObjectPropertyOf(modeus:confina_con_delegazione modeus:delegazione_confina_con)
SubObjectPropertyOf(modeus:confina_con_delegazione ObjectInverseOf(modeus:confina_con_delegazione))
ObjectPropertyRange(modeus:confina_con_delegazione modeus:Delegazione)
ObjectPropertyDomain(modeus:copia_brogliardo_est_destinata_a modeus:Copia_brogliardo)
ObjectPropertyRange(modeus:copia_brogliardo_est_destinata_a modeus:Organo_periferico)
ObjectPropertyDomain(modeus:delegazione_confina_con modeus:Delegazione)
ObjectPropertyRange(modeus:delegazione_confina_con ObjectUnionOf(modeus:Delegazione modeus:Stato))
ObjectPropertyDomain(modeus:delegazione_est_divisa modeus:Delegazione)
ObjectPropertyRange(modeus:delegazione_est_divisa modeus:Governo_distrettuale)
ObjectPropertyDomain(modeus:delegazione_est_inclusa modeus:Delegazione)
ObjectPropertyRange(modeus:delegazione_est_inclusa modeus:Stato)
ObjectPropertyDomain(modeus:est_aiutante_R_di_mappa modeus:Aiutante_R)
ObjectPropertyRange(modeus:est_aiutante_R_di_mappa modeus:Mappa)
ObjectPropertyDomain(modeus:est_aiutante_R_tratta_da modeus:Aiutante_R)
ObjectPropertyRange(modeus:est_aiutante_R_tratta_da modeus:Mappa)
ObjectPropertyDomain(modeus:est_aiutante_di_mappa modeus:Persona)
ObjectPropertyRange(modeus:est_aiutante_di_mappa modeus:Mappa)
ObjectPropertyDomain(modeus:est_assistente_R_mappa modeus:Assistente_R)
ObjectPropertyRange(modeus:est_assistente_R_mappa modeus:Mappa)
ObjectPropertyDomain(modeus:est_assistente_R_tratta_da modeus:Assistente_R)
ObjectPropertyRange(modeus:est_assistente_R_tratta_da modeus:Mappa)
ObjectPropertyDomain(modeus:est_assistente_di_mappa modeus:Persona)
ObjectPropertyRange(modeus:est_assistente_di_mappa modeus:Mappa)
ObjectPropertyDomain(modeus:est_brogliardo_di_scrittura_primaria modeus:Brogliardo)
ObjectPropertyRange(modeus:est_brogliardo_di_scrittura_primaria modeus:Scrittura_impianto_primaria)
ObjectPropertyDomain(modeus:est_ente_intestatario_R_di_intestazione_R modeus:Ente_intestatario_R)
ObjectPropertyRange(modeus:est_ente_intestatario_R_di_intestazione_R modeus:Intestazione_R)
ObjectPropertyDomain(modeus:est_ente_intestatario_R_padrone_diretto modeus:Ente_intestatario_R)
ObjectPropertyRange(modeus:est_ente_intestatario_R_padrone_diretto modeus:Rappresentazione_particella)
ObjectPropertyDomain(modeus:est_ente_intestatario_R_padrone_diretto_tratta_da
modeus:Ente_intestatario_R_padrone_diretto)
ObjectPropertyRange(modeus:est_ente_intestatario_R_padrone_diretto_tratta_da
ObjectUnionOf(modeus:Brogliardo modeus:Catastino modeus:Istanza_voltura modeus:Registro_trasporti))

SubObjectPropertyOf(modeus:est_ente_intestatario_R_pro_indiviso modeus:est_ente_intestatario_R_di_intestazione_R)
ObjectPropertyDomain(modeus:est_ente_intestatario_R_pro_indiviso modeus:Ente_intestatario_R)
ObjectPropertyRange(modeus:est_ente_intestatario_R_pro_indiviso modeus:Intestazione_R_pro_indiviso)
ObjectPropertyDomain(modeus:est_ente_intestatario_R_pro_indiviso_tratta_da modeus:Ente_intestatario_R_pro_indiviso)
ObjectPropertyRange(modeus:est_ente_intestatario_R_pro_indiviso_tratta_da ObjectUnionOf(modeus:Brogliardo modeus:Registro_trasporti))
SubObjectPropertyOf(modeus:est_ente_intestatario_R_singolo modeus:est_ente_intestatario_R_di_intestazione_R)
ObjectPropertyDomain(modeus:est_ente_intestatario_R_singolo modeus:Ente_intestatario_R)
ObjectPropertyRange(modeus:est_ente_intestatario_R_singolo modeus:Intestazione_R_singolo)
ObjectPropertyDomain(modeus:est_ente_intestatario_R_singolo_tratta_da modeus:Ente_intestatario_R_singolo)
ObjectPropertyRange(modeus:est_ente_intestatario_R_singolo_tratta_da ObjectUnionOf(modeus:Brogliardo modeus:Registro_trasporti))
ObjectPropertyDomain(modeus:est_ente_intestatario_R_utilista modeus:Ente_intestatario_R)
ObjectPropertyRange(modeus:est_ente_intestatario_R_utilista modeus:Rappresentazione_particella)
ObjectPropertyDomain(modeus:est_ente_intestatario_R_utilista_tratta_da modeus:Ente_intestatario_R_utilista)
ObjectPropertyRange(modeus:est_ente_intestatario_R_utilista_tratta_da ObjectUnionOf(modeus:Brogliardo modeus:Catastino modeus:Istanza_voltura modeus:Registro_trasporti))
ObjectPropertyDomain(modeus:est_ente_intestatario_di_intestazione modeus:Ente_intestatario)
ObjectPropertyRange(modeus:est_ente_intestatario_di_intestazione modeus:Intestazione)
ObjectPropertyDomain(modeus:est_ente_intestatario_padrone_diretto modeus:Ente_intestatario)
ObjectPropertyRange(modeus:est_ente_intestatario_padrone_diretto modeus:Rappresentazione_particella)
SubObjectPropertyOf(modeus:est_ente_intestatario_pro_indiviso modeus:est_ente_intestatario_di_intestazione)
ObjectPropertyDomain(modeus:est_ente_intestatario_pro_indiviso modeus:Ente_intestatario)
ObjectPropertyRange(modeus:est_ente_intestatario_pro_indiviso modeus:Intestazione_pro_indiviso)
SubObjectPropertyOf(modeus:est_ente_intestatario_singolo modeus:est_ente_intestatario_di_intestazione)
ObjectPropertyDomain(modeus:est_ente_intestatario_singolo modeus:Ente_intestatario)
ObjectPropertyRange(modeus:est_ente_intestatario_singolo modeus:Intestazione_singolo)
ObjectPropertyDomain(modeus:est_ente_intestatario_utilista modeus:Ente_intestatario)
ObjectPropertyRange(modeus:est_ente_intestatario_utilista modeus:Rappresentazione_particella)
ObjectPropertyDomain(modeus:est_figlio modeus:Persona_intestatario_R)
ObjectPropertyRange(modeus:est_figlio modeus:Relazione_persona_intestatario_R)
ObjectPropertyDomain(modeus:est_geometra_R_di_mappa modeus:Geometra_R)
ObjectPropertyRange(modeus:est_geometra_R_di_mappa modeus:Mappa)
ObjectPropertyDomain(modeus:est_geometra_di_mappa modeus:Geometra)
ObjectPropertyRange(modeus:est_geometra_di_mappa modeus:Mappa)
ObjectPropertyDomain(modeus:est_indicatore_R_di_mappa modeus:Indicatore_R)
ObjectPropertyRange(modeus:est_indicatore_R_di_mappa modeus:Mappa)
ObjectPropertyDomain(modeus:est_indicatore_R_tratta_da modeus:Indicatore_R)
ObjectPropertyRange(modeus:est_indicatore_R_tratta_da modeus:Mappa)
ObjectPropertyDomain(modeus:est_indicatore_di_mappa modeus:Persona)
ObjectPropertyRange(modeus:est_indicatore_di_mappa modeus:Mappa)
ObjectPropertyDomain(modeus:est_intestazione_R_precedente modeus:Intestazione_R)
ObjectPropertyRange(modeus:est_intestazione_R_precedente modeus:Successione_intestazione_R)
ObjectPropertyDomain(modeus:est_intestazione_R_successiva modeus:Intestazione_R)
ObjectPropertyRange(modeus:est_intestazione_R_successiva modeus:Successione_intestazione_R)
ObjectPropertyDomain(modeus:est_intestazione_R_tratta_da modeus:Intestazione_R)
ObjectPropertyDomain(modeus:est_ispettore_R_di_mappa modeus:Tecnico_R)
ObjectPropertyRange(modeus:est_ispettore_R_di_mappa modeus:Mappa)
ObjectPropertyDomain(modeus:est_ispettore_di_mappa modeus:Tecnico)
ObjectPropertyRange(modeus:est_ispettore_di_mappa modeus:Mappa)
ObjectPropertyDomain(modeus:est_mappa_di_scrittura_primaria modeus:Mappa)
ObjectPropertyRange(modeus:est_mappa_di_scrittura_primaria modeus:Scrittura_impianto_primaria)
ObjectPropertyDomain(modeus:est_mappetta_tela_di_scrittura_primaria modeus:Mappetta_tela)
ObjectPropertyRange(modeus:est_mappetta_tela_di_scrittura_primaria modeus:Scrittura_impianto_primaria)
ObjectPropertyDomain(modeus:est_marito_di_moglie modeus:Persona_R)
ObjectPropertyRange(modeus:est_marito_di_moglie modeus:Relazione_persona_intestatario_R)
ObjectPropertyDomain(modeus:est_marito_di_vedova modeus:Persona_R)
ObjectPropertyRange(modeus:est_marito_di_vedova modeus:Relazione_persona_intestatario_R)
ObjectPropertyDomain(modeus:est_moglie modeus:Persona_intestatario_moglie_R)
ObjectPropertyRange(modeus:est_moglie modeus:Relazione_persona_intestatario_R)

ObjectPropertyDomain(modeus:est_padre modeus:Persona_R)
ObjectPropertyRange(modeus:est_padre modeus:Relazione_persona_intestatario_R)
ObjectPropertyDomain(modeus:est_parte_di_unità_di_descrizione modeus:Unità_di_descrizione)
ObjectPropertyRange(modeus:est_parte_di_unità_di_descrizione modeus:Unità_di_descrizione)
ObjectPropertyDomain(modeus:est_persona_intestatario_R_di_intestazione_R modeus:Persona_intestatario_R)
ObjectPropertyRange(modeus:est_persona_intestatario_R_di_intestazione_R modeus:Intestazione_R)
ObjectPropertyDomain(modeus:est_persona_intestatario_R_padrone_diretto modeus:Persona_intestatario_R)
ObjectPropertyRange(modeus:est_persona_intestatario_R_padrone_diretto modeus:Rappresentazione_particella)
ObjectPropertyDomain(modeus:est_persona_intestatario_R_padrone_diretto_tratta_da modeus:Persona_intestatario_R_padrone_diretto)
ObjectPropertyRange(modeus:est_persona_intestatario_R_padrone_diretto_tratta_da ObjectUnionOf(modeus:Brogliardo modeus:Catastino modeus:Istanza_voltura modeus:Registro_trasporti))
SubObjectPropertyOf(modeus:est_persona_intestatario_R_pro_indiviso modeus:est_persona_intestatario_R_di_intestazione_R)
ObjectPropertyDomain(modeus:est_persona_intestatario_R_pro_indiviso modeus:Persona_intestatario_R)
ObjectPropertyRange(modeus:est_persona_intestatario_R_pro_indiviso modeus:Intestazione_R_pro_indiviso)
ObjectPropertyDomain(modeus:est_persona_intestatario_R_pro_indiviso_tratta_da modeus:Persona_intestatario_R_pro_indiviso)
ObjectPropertyRange(modeus:est_persona_intestatario_R_pro_indiviso_tratta_da ObjectUnionOf(modeus:Brogliardo modeus:Registro_trasporti))
SubObjectPropertyOf(modeus:est_persona_intestatario_R_singolo modeus:est_persona_intestatario_R_di_intestazione_R)
ObjectPropertyDomain(modeus:est_persona_intestatario_R_singolo modeus:Persona_intestatario_R)
ObjectPropertyRange(modeus:est_persona_intestatario_R_singolo modeus:Intestazione_R_singolo)
ObjectPropertyDomain(modeus:est_persona_intestatario_R_singolo_tratta_da modeus:Persona_intestatario_R_singolo)
ObjectPropertyRange(modeus:est_persona_intestatario_R_singolo_tratta_da ObjectUnionOf(modeus:Brogliardo modeus:Registro_trasporti))
ObjectPropertyDomain(modeus:est_persona_intestatario_R_utilista modeus:Persona_intestatario_R)
ObjectPropertyRange(modeus:est_persona_intestatario_R_utilista modeus:Rappresentazione_particella)
ObjectPropertyDomain(modeus:est_persona_intestatario_R_utilista_tratta_da modeus:Persona_intestatario_R_utilista)
ObjectPropertyRange(modeus:est_persona_intestatario_R_utilista_tratta_da ObjectUnionOf(modeus:Brogliardo modeus:Catastino modeus:Istanza_voltura modeus:Registro_trasporti))
ObjectPropertyDomain(modeus:est_persona_intestatario_di_intestazione modeus:Persona_intestatario)
ObjectPropertyRange(modeus:est_persona_intestatario_di_intestazione modeus:Intestazione)
ObjectPropertyDomain(modeus:est_persona_intestatario_padrone_diretto modeus:Persona_intestatario)
ObjectPropertyRange(modeus:est_persona_intestatario_padrone_diretto modeus:Rappresentazione_particella)
SubObjectPropertyOf(modeus:est_persona_intestatario_pro_indiviso modeus:est_persona_intestatario_di_intestazione)
ObjectPropertyDomain(modeus:est_persona_intestatario_pro_indiviso modeus:Persona_intestatario)
ObjectPropertyRange(modeus:est_persona_intestatario_pro_indiviso modeus:Intestazione_pro_indiviso)
SubObjectPropertyOf(modeus:est_persona_intestatario_singolo modeus:est_persona_intestatario_di_intestazione)
ObjectPropertyDomain(modeus:est_persona_intestatario_singolo modeus:Persona_intestatario)
ObjectPropertyRange(modeus:est_persona_intestatario_singolo modeus:Intestazione_singolo)
ObjectPropertyDomain(modeus:est_persona_intestatario_utilista modeus:Persona_intestatario)
ObjectPropertyRange(modeus:est_persona_intestatario_utilista modeus:Rappresentazione_particella)
ObjectPropertyDomain(modeus:est_rappresentazione_aiutante modeus:Aiutante_R)
ObjectPropertyRange(modeus:est_rappresentazione_aiutante modeus:Persona)
ObjectPropertyDomain(modeus:est_rappresentazione_assistente modeus:Assistente_R)
ObjectPropertyRange(modeus:est_rappresentazione_assistente modeus:Persona)
ObjectPropertyDomain(modeus:est_rappresentazione_di_geometra modeus:Geometra_R)
ObjectPropertyRange(modeus:est_rappresentazione_di_geometra modeus:Geometra)
ObjectPropertyDomain(modeus:est_rappresentazione_di_intestazione modeus:Intestazione_R)
ObjectPropertyRange(modeus:est_rappresentazione_di_intestazione modeus:Intestazione)
ObjectPropertyDomain(modeus:est_rappresentazione_di_persona modeus:Persona_R)
ObjectPropertyRange(modeus:est_rappresentazione_di_persona modeus:Persona)
ObjectPropertyDomain(modeus:est_rappresentazione_di_persona_intestatario modeus:Persona_intestatario_R)
ObjectPropertyRange(modeus:est_rappresentazione_di_persona_intestatario modeus:Persona_intestatario)
ObjectPropertyDomain(modeus:est_rappresentazione_di_tecnico modeus:Tecnico_R)
ObjectPropertyRange(modeus:est_rappresentazione_di_tecnico modeus:Tecnico)
ObjectPropertyDomain(modeus:est_rappresentazione_ente_intestatario modeus:Ente_intestatario)

ObjectPropertyRange(modeus:est_rappresentazione_ente_intestatario modeus:Ente_intestatario_R)
ObjectPropertyDomain(modeus:est_rappresentazione_indicatore modeus:Indicatore_R)
ObjectPropertyRange(modeus:est_rappresentazione_indicatore modeus:Persona)
ObjectPropertyDomain(modeus:est_relazione_persona_intestatario_R_tratta_da modeus:Relazione_persona_intestatario_R)
ObjectPropertyRange(modeus:est_relazione_persona_intestatario_R_tratta_da ObjectUnionOf(modeus:Brogliardo modeus:Catastino modeus:Istanza_voltura modeus:Registro_trasporti))
ObjectPropertyDomain(modeus:est_soggetto_produttore_di modeus:Soggetto_produttore)
ObjectPropertyRange(modeus:est_soggetto_produttore_di modeus:Unità_di_descrizione)
ObjectPropertyDomain(modeus:est_successione_intestazione_R_tratta_da modeus:Successione_intestazione_R)
ObjectPropertyRange(modeus:est_successione_intestazione_R_tratta_da ObjectUnionOf(modeus:Brogliardo modeus:Catastino modeus:Istanza_voltura modeus:Registro_trasporti))
ObjectPropertyDomain(modeus:est_tecnico_R_tratta_da modeus:Tecnico_R)
ObjectPropertyRange(modeus:est_tecnico_R_tratta_da ObjectUnionOf(modeus:Brogliardo modeus:Mappa))
ObjectPropertyDomain(modeus:est_unità_di_descrizione_successiva modeus:Unità_di_descrizione)
ObjectPropertyRange(modeus:est_unità_di_descrizione_successiva modeus:Unità_di_descrizione)
ObjectPropertyDomain(modeus:est_vedova modeus:Persona_intestatario_vedova_R)
ObjectPropertyRange(modeus:est_vedova modeus:Relazione_persona_intestatario_R)
ObjectPropertyDomain(modeus:est_verificatore_R_di_mappa modeus:Tecnico_R)
ObjectPropertyRange(modeus:est_verificatore_R_di_mappa modeus:Mappa)
ObjectPropertyDomain(modeus:est_verificatore_di_mappa modeus:Tecnico)
ObjectPropertyRange(modeus:est_verificatore_di_mappa modeus:Mappa)
ObjectPropertyDomain(modeus:funzione_est_documentata modeus:Funzione)
ObjectPropertyRange(modeus:funzione_est_documentata modeus:Unità_di_descrizione)
ObjectPropertyDomain(modeus:istanza_est_registrata_in modeus:Istanza_voltura)
ObjectPropertyRange(modeus:istanza_est_registrata_in modeus:Registro_istanza_voltura)
ObjectPropertyDomain(modeus:istanza_voltura_est_associata modeus:Istanza_voltura)
ObjectPropertyRange(modeus:istanza_voltura_est_associata modeus:Atto_notarile)
ObjectPropertyDomain(modeus:istituto_conservatore_est_ubicato_fisicamente modeus:Isituto_conservatore)
ObjectPropertyRange(modeus:istituto_conservatore_est_ubicato_fisicamente modeus:Sede)
ObjectPropertyDomain(modeus:mappa_copia_scala_originale_digitalizzata_est_legata_a_mappa_digitalizzata modeus:Mappa_copia_scala_originale_digitalizzata)
ObjectPropertyRange(modeus:mappa_copia_scala_originale_digitalizzata_est_legata_a_mappa_digitalizzata modeus:Mappa_digitalizzata)
ObjectPropertyDomain(modeus:mappa_copia_scala_originale_est_destinata_a modeus:Mappa_copia_scala_originale)
ObjectPropertyRange(modeus:mappa_copia_scala_originale_est_destinata_a modeus:Organo_periferico)
ObjectPropertyDomain(modeus:mappa_copia_scala_originale_ha_riproduzione_digitale modeus:Mappa_copia_scala_originale)
ObjectPropertyRange(modeus:mappa_copia_scala_originale_ha_riproduzione_digitale modeus:Mappa_copia_scala_originale_digitalizzata)
ObjectPropertyDomain(modeus:mappa_digitalizzata_est_descritta_da_brogliardo_digitalizzato modeus:Mappa_digitalizzata)
ObjectPropertyRange(modeus:mappa_digitalizzata_est_descritta_da_brogliardo_digitalizzato modeus:Brogliardo_digitalizzato)
ObjectPropertyDomain(modeus:mappa_est_descritta_da_brogliardo modeus:Mappa)
ObjectPropertyRange(modeus:mappa_est_descritta_da_brogliardo modeus:Brogliardo)
ObjectPropertyDomain(modeus:mappa_est_relativa modeus:Mappa)
ObjectPropertyRange(modeus:mappa_est_relativa modeus:Comune)
ObjectPropertyDomain(modeus:mappa_est_riprodotta_in modeus:Mappa)
ObjectPropertyRange(modeus:mappa_est_riprodotta_in modeus:Rappresentazione_cartacea_mappa)
ObjectPropertyDomain(modeus:mappa_ha_riproduzione_digitale modeus:Mappa)
ObjectPropertyRange(modeus:mappa_ha_riproduzione_digitale modeus:Mappa_digitalizzata)
ObjectPropertyDomain(modeus:mappetta_carta_est_destinata_a modeus:Mappetta_carta)
ObjectPropertyRange(modeus:mappetta_carta_est_destinata_a modeus:Organo_periferico)
ObjectPropertyDomain(modeus:organo_periferico_ha_competenza modeus:Organo_periferico)
ObjectPropertyRange(modeus:organo_periferico_ha_competenza modeus:Circoscrizione)
ObjectPropertyDomain(modeus:organo_periferico_ha_sede modeus:Organo_periferico)
ObjectPropertyRange(modeus:organo_periferico_ha_sede modeus:Co_mune_sede_di_ufficio_di_registro)
SubObjectPropertyOf(modeus:rappresentazione_particella_confina_con ObjectInverseOf(modeus:rappresentazione_particella_confina_con))
ObjectPropertyDomain(modeus:rappresentazione_particella_confina_con modeus:Rappresentazione_particella)

ObjectPropertyRange(modeus:rappresentazione_particella_confina_con modeus:Rappresentazione_particella)
ObjectPropertyDomain(modeus:rappresentazione_particella_confina_con_oggetto_fisico modeus:Rappresentazione_particella)
ObjectPropertyRange(modeus:rappresentazione_particella_confina_con_oggetto_fisico modeus:Oggetto_fisico)
ObjectPropertyDomain(modeus:rappresentazione_particella_est_descritta modeus:Rappresentazione_particella)
ObjectPropertyRange(modeus:rappresentazione_particella_est_descritta modeus:Brogliardo)
ObjectPropertyDomain(modeus:rappresentazione_particella_est_frazionata_in modeus:Rappresentazione_particella)
ObjectPropertyRange(modeus:rappresentazione_particella_est_frazionata_in modeus:Rappresentazione_particella)
ObjectPropertyDomain(modeus:rappresentazione_particella_est_fusa_in modeus:Rappresentazione_particella)
ObjectPropertyRange(modeus:rappresentazione_particella_est_fusa_in modeus:Rappresentazione_particella)
ObjectPropertyDomain(modeus:rappresentazione_particella_ha_destinazione modeus:Rappresentazione_particella)
ObjectPropertyRange(modeus:rappresentazione_particella_ha_destinazione modeus:Destinazione)
ObjectPropertyDomain(modeus:rappresentazione_particella_ha_intestazione modeus:Rappresentazione_particella)
ObjectPropertyRange(modeus:rappresentazione_particella_ha_intestazione modeus:Intestazione)
ObjectPropertyDomain(modeus:rappresentazione_particella_ha_intestazione_R modeus:Rappresentazione_particella)
ObjectPropertyRange(modeus:rappresentazione_particella_ha_intestazione_R modeus:Intestazione_R)
ObjectPropertyDomain(modeus:relazione_coniugio modeus:Persona)
ObjectPropertyRange(modeus:relazione_coniugio modeus:Persona)
ObjectPropertyDomain(modeus:relazione_paternità modeus:Persona)
ObjectPropertyRange(modeus:relazione_paternità modeus:Persona)
ObjectPropertyDomain(modeus:sezione_catastale_contiene_oggetto_fisico modeus:Sezione_catastale)
ObjectPropertyRange(modeus:sezione_catastale_contiene_oggetto_fisico modeus:Oggetto_fisico)
SubObjectPropertyOf(modeus:sezione_catastale_contiene_rappresentazione_particella modeus:sezione_catastale_contiene_oggetto_fisico)
ObjectPropertyDomain(modeus:sezione_catastale_contiene_rappresentazione_particella modeus:Sezione_catastale)
ObjectPropertyRange(modeus:sezione_catastale_contiene_rappresentazione_particella modeus:Rappresentazione_particella)
ObjectPropertyDomain(modeus:sezione_catastale_est_rappresentata_graficamente modeus:Sezione_catastale)
ObjectPropertyRange(modeus:sezione_catastale_est_rappresentata_graficamente modeus:Mappa)
ObjectPropertyDomain(modeus:soggetto_produttore_adempie_funzione modeus:Soggetto_produttore)
ObjectPropertyRange(modeus:soggetto_produttore_adempie_funzione modeus:Funzione)
ObjectPropertyDomain(modeus:successione_intestazione modeus:Intestazione)
ObjectPropertyRange(modeus:successione_intestazione modeus:Intestazione)
ObjectPropertyDomain(modeus:unità_di_descrizione_est_conservata modeus:Unità_di_descrizione)
ObjectPropertyRange(modeus:unità_di_descrizione_est_conservata modeus:Isituto_conservatore)
ObjectPropertyDomain(modeus:unità_di_descrizione_ha_livello modeus:Unità_di_descrizione)
ObjectPropertyRange(modeus:unità_di_descrizione_ha_livello modeus:Livello_di_descrizione)
DataPropertyDomain(modeus:anno_aggiornamento_mappa_copia modeus:Mappa_copia_scala_originale)
DataPropertyDomain(modeus:anno_di_morte modeus:Persona)
DataPropertyDomain(modeus:anno_di_nascita modeus:Persona)
DataPropertyDomain(modeus:anno_fine_intestazione modeus:Intestazione)
DataPropertyDomain(modeus:anno_fine_intestazione_R modeus:Intestazione_R)
DataPropertyDomain(modeus:anno_fine_range_unità modeus:Unità_di_descrizione)
DataPropertyDomain(modeus:anno_formazione_unità modeus:Unità_di_descrizione)
DataPropertyDomain(modeus:anno_inizio_intestazione modeus:Intestazione)
DataPropertyDomain(modeus:anno_inizio_intestazione_R modeus:Intestazione_R)
DataPropertyDomain(modeus:anno_inizio_range_unità modeus:Unità_di_descrizione)
DataPropertyDomain(modeus:anno_istituzione_ente modeus:Ente)
DataPropertyDomain(modeus:anno_morte_rappresentazione_particella modeus:Rappresentazione_particella)
DataPropertyDomain(modeus:anno_morte_rappresentazione_particella ObjectUnionOf(ObjectSomeValuesFrom(modeus:rappresentazione_particella_est_frazionata_in owl:Thing) ObjectSomeValuesFrom(modeus:rappresentazione_particella_est_fusa_in owl:Thing)))
DataPropertyDomain(modeus:anno_nascita_rappresentazione_particella modeus:Rappresentazione_particella)
DataPropertyDomain(modeus:anno_redazione_atto modeus:Atto_notarile)
DataPropertyDomain(modeus:anno_registrazione_atto modeus:Atto_notarile)
DataPropertyDomain(modeus:anno_soppressione_ente modeus:Ente)
DataPropertyDomain(modeus:città_sede modeus:Sede)
DataPropertyDomain(modeus:codice_identificativo_conservatore modeus:Isituto_conservatore)
DataPropertyDomain(modeus:codice_identificativo_descrizione_funzione modeus:Funzione)
DataPropertyDomain(modeus:codice_identificativo_record_autorità modeus:Soggetto_produttore)
DataPropertyDomain(modeus:codice_identificativo_unità_descrizione modeus:Unità_di_descrizione)

DataPropertyDomain(modeus:cognome_persona modeus:Persona)
DataPropertyDomain(modeus:cognome_persona_R modeus:Persona_R)
DataPropertyDomain(modeus:consistenza_unità_descrizione modeus:Unità_di_descrizione)
DataPropertyDomain(modeus:domicilio_ente_intestatario modeus:Ente_intestatario)
DataPropertyDomain(modeus:domicilio_ente_intestatario_R modeus:Ente_intestatario_R)
DataPropertyDomain(modeus:domicilio_persona_intestatario modeus:Persona_intestatario)
DataPropertyDomain(modeus:domicilio_persona_intestatario_R modeus:Persona_intestatario_R)
DataPropertyDomain(modeus:estensione_di_particella modeus:Rappresentazione_particella)
DataPropertyDomain(modeus:estimo_di_particella modeus:Rappresentazione_particella)
DataPropertyDomain(modeus:forma_autorizzata_nome_funzione modeus:Funzione)
DataPropertyDomain(modeus:forma_autorizzata_nome_istituto_conservatore modeus:Isituto_conservatore)
DataPropertyDomain(modeus:forma_autorizzata_nome_soggetto_produttore modeus:Soggetto_produttore)
DataPropertyDomain(modeus:giorno_aggiornamento_mappa_copia modeus:Mappa_copia_scala_originale)
DataPropertyDomain(modeus:giorno_di_morte modeus:Persona)
DataPropertyDomain(modeus:giorno_di_nascita modeus:Persona)
DataPropertyDomain(modeus:giorno_fine_intestazione modeus:Intestazione)
DataPropertyDomain(modeus:giorno_fine_intestazione_R modeus:Intestazione_R)
DataPropertyDomain(modeus:giorno_fine_range_unità modeus:Unità_di_descrizione)
DataPropertyDomain(modeus:giorno_formazione_unità modeus:Unità_di_descrizione)
DataPropertyDomain(modeus:giorno_inizio_intestazione modeus:Intestazione)
DataPropertyDomain(modeus:giorno_inizio_intestazione_R modeus:Intestazione_R)
DataPropertyDomain(modeus:giorno_inizio_range_unità modeus:Unità_di_descrizione)
DataPropertyDomain(modeus:giorno_istituzione_ente modeus:Ente)
DataPropertyDomain(modeus:giorno_morte_rappresentazione_particella modeus:Rappresentazione_particella)
DataPropertyDomain(modeus:giorno_morte_rappresentazione_particella
ObjectUnionOf(ObjectSomeValuesFrom(modeus:rappresentazione_particella_est_frazionata_in owl:Thing)
ObjectSomeValuesFrom(modeus:rappresentazione_particella_est_fusa_in owl:Thing)))
DataPropertyDomain(modeus:giorno_nascita_rappresentazione_particella modeus:Rappresentazione_particella)
DataPropertyDomain(modeus:giorno_redazione_atto modeus:Atto_notarile)
DataPropertyDomain(modeus:giorno_registrazione_atto modeus:Atto_notarile)
DataPropertyDomain(modeus:giorno_soppressione_ente modeus:Ente)
DataPropertyDomain(modeus:indirizzo_sede modeus:Sede)
SubDataPropertyOf(modeus:indirizzo_sede_principale modeus:indirizzo_sede)
DisjointDataProperties(modeus:indirizzo_sede_principale modeus:indirizzo_sede_succursale)
SubDataPropertyOf(modeus:indirizzo_sede_succursale modeus:indirizzo_sede)
DataPropertyDomain(modeus:luogo_di_morte modeus:Persona)
DataPropertyDomain(modeus:luogo_di_nascita modeus:Persona)
DataPropertyDomain(modeus:mappale_principale_di_particella modeus:Rappresentazione_particella)
DataPropertyDomain(modeus:mappale_subordinato_di_particella modeus:Rappresentazione_particella)
DataPropertyDomain(modeus:mese_aggiornamento_mappa_copia modeus:Mappa_copia_scala_originale)
DataPropertyDomain(modeus:mese_di_morte modeus:Persona)
DataPropertyDomain(modeus:mese_di_nascita modeus:Persona)
DataPropertyDomain(modeus:mese_fine_intestazione modeus:Intestazione)
DataPropertyDomain(modeus:mese_fine_intestazione_R modeus:Intestazione_R)
DataPropertyDomain(modeus:mese_fine_range_unità modeus:Unità_di_descrizione)
DataPropertyDomain(modeus:mese_formazione_unità modeus:Unità_di_descrizione)
DataPropertyDomain(modeus:mese_inizio_intestazione modeus:Intestazione)
DataPropertyDomain(modeus:mese_inizio_intestazione_R modeus:Intestazione_R)
DataPropertyDomain(modeus:mese_inizio_range_unità modeus:Unità_di_descrizione)
DataPropertyDomain(modeus:mese_istituzione_ente modeus:Ente)
DataPropertyDomain(modeus:mese_morte_rappresentazione_particella modeus:Rappresentazione_particella)
DataPropertyDomain(modeus:mese_morte_rappresentazione_particella
ObjectUnionOf(ObjectSomeValuesFrom(modeus:rappresentazione_particella_est_frazionata_in owl:Thing)
ObjectSomeValuesFrom(modeus:rappresentazione_particella_est_fusa_in owl:Thing)))
DataPropertyDomain(modeus:mese_nascita_rappresentazione_particella modeus:Rappresentazione_particella)
DataPropertyDomain(modeus:mese_redazione_atto modeus:Atto_notarile)
DataPropertyDomain(modeus:mese_registrazione_atto modeus:Atto_notarile)
DataPropertyDomain(modeus:mese_soppressione_ente modeus:Ente)
DataPropertyDomain(modeus:nome_acqua modeus:Acqua)
DataPropertyDomain(modeus:nome_circoscrizione modeus:Circoscrizione)
DataPropertyDomain(modeus:nome_comune modeus:Comune)
DataPropertyDomain(modeus:nome_contrada_di_particella modeus:Rappresentazione_particella)

DataPropertyDomain(modeus:nome_delegazione modeus:Delegazione)
DataPropertyDomain(modeus:nome_destinazione_produttiva modeus:destinazione_produttiva)
DataPropertyDomain(modeus:nome_destinazione_uso modeus:destinazione_uso)
DataPropertyDomain(modeus:nome_ente modeus:Ente)
DataPropertyDomain(modeus:nome_ente_R modeus:Ente_R)
DataPropertyDomain(modeus:nome_governo_distrettuale modeus:Governo_distrettuale)
DataPropertyDomain(modeus:nome_persona modeus:Persona)
DataPropertyDomain(modeus:nome_persona_R modeus:Persona_R)
DataPropertyDomain(modeus:nome_sezione_catastale modeus:Sezione_catastale)
DataPropertyDomain(modeus:nome_stato modeus:Stato)
DataPropertyDomain(modeus:nome_strada modeus:Strada)
DataPropertyDomain(modeus:numero_mutazione_registro_trasporti modeus:Registro_trasporti)
DataPropertyDomain(modeus:numero_pagina_catastino modeus:Catastino)
DataPropertyDomain(modeus:numero_pagina_registro_trasporti modeus:Registro_trasporti)
EquivalentDataProperties(modeus:numero_progressivo_istanza_voltura
modeus:numero_protocollo_registrazione_istanza_voltura)
DataPropertyDomain(modeus:numero_progressivo_istanza_voltura modeus:Istanza_voltura)
DataPropertyDomain(modeus:numero_sezione_catastale modeus:Sezione_catastale)
DataPropertyDomain(modeus:qualifica_tecnico_R modeus:Tecnico_R)
DataPropertyDomain(modeus:quantità_componente_documentaria modeus:Componente_unità_documentaria)
DataPropertyDomain(modeus:scala_mappa modeus:Mappa)
DataPropertyDomain(modeus:scala_mappetta_carta modeus:Mappetta_carta)
DataPropertyDomain(modeus:scala_mappetta_tela modeus:Mappetta_tela)
DataPropertyDomain(modeus:segnatura_mappa_copia_originale_digitalizzata
modeus:Mappa_copia_scala_originale_digitalizzata)
DataPropertyDomain(modeus:segnatura_mappa_digitalizzata modeus:Mappa_digitalizzata)
DataPropertyDomain(modeus:segnatura_unità_descrizione modeus:Unità_di_descrizione)
DataPropertyDomain(modeus:sito_web_istituto_conservatore modeus:Isituto_conservatore)
DataPropertyDomain(modeus:stato_unità_documentaria modeus:Unità_documentaria)
DataPropertyDomain(modeus:supporto_unità_descrizione modeus:Unità_di_descrizione)
DataPropertyDomain(modeus:tipo_acqua modeus:Acqua)
DataPropertyDomain(modeus:tipo_strada modeus:Strada)
DataPropertyDomain(modeus:tipologia_componente_documentaria modeus:Componente_unità_documentaria)
DataPropertyDomain(modeus:tipologia_documentaria_unità_descrizione modeus:Unità_di_descrizione)
DataPropertyDomain(modeus:tipologia_funzione modeus:Funzione)
SubDataPropertyOf(modeus:titolo_attribuito_unità_descrizione modeus:titolo_unità_descrizione)
SubDataPropertyOf(modeus:titolo_originale_unità_descrizione modeus:titolo_unità_descrizione)
DataPropertyDomain(modeus:titolo_unità_descrizione modeus:Unità_di_descrizione)
EquivalentClasses(modeus:Acqua DataSomeValuesFrom(modeus:tipo_acqua rdfs:Literal))
DisjointClasses(modeus:Acqua modeus:Strada)
EquivalentClasses(modeus:Agente ObjectSomeValuesFrom(modeus:agente_est_soggetto_produttore owl:Thing))
EquivalentClasses(modeus:Aiutante_R ObjectSomeValuesFrom(modeus:est_aiutante_R_tratta_da owl:Thing))
EquivalentClasses(modeus:Aiutante_R ObjectSomeValuesFrom(modeus:est_rappresentazione_aiutante owl:Thing))
SubClassOf(modeus:Aiutante_R modeus:Persona_R)
EquivalentClasses(modeus:Assistente_R ObjectSomeValuesFrom(modeus:est_assistente_R_tratta_da owl:Thing))
EquivalentClasses(modeus:Assistente_R ObjectSomeValuesFrom(modeus:est_rappresentazione_assistente
owl:Thing))
SubClassOf(modeus:Assistente_R modeus:Persona_R)
EquivalentClasses(modeus:Atto_notarile ObjectSomeValuesFrom(modeus:atto_notarile_est_redatto owl:Thing))
EquivalentClasses(modeus:Atto_notarile
ObjectSomeValuesFrom(ObjectInverseOf(modeus:istanza_voltura_est_associata) owl:Thing))
EquivalentClasses(modeus:Atto_notarile DataSomeValuesFrom(modeus:anno_redazione_atto rdfs:Literal))
EquivalentClasses(modeus:Atto_notarile DataSomeValuesFrom(modeus:anno_registrazione_atto rdfs:Literal))
EquivalentClasses(modeus:Atto_notarile DataSomeValuesFrom(modeus:giorno_redazione_atto rdfs:Literal))
EquivalentClasses(modeus:Atto_notarile DataSomeValuesFrom(modeus:giorno_registrazione_atto rdfs:Literal))
EquivalentClasses(modeus:Atto_notarile DataSomeValuesFrom(modeus:mese_redazione_atto rdfs:Literal))
EquivalentClasses(modeus:Atto_notarile DataSomeValuesFrom(modeus:mese_registrazione_atto rdfs:Literal))
EquivalentClasses(modeus:Brogliardo ObjectSomeValuesFrom(modeus:brogliardo_est_riprodotto owl:Thing))
EquivalentClasses(modeus:Brogliardo ObjectSomeValuesFrom(modeus:est_brogliardo_di_scrittura_primaria
owl:Thing))
EquivalentClasses(modeus:Brogliardo
ObjectSomeValuesFrom(ObjectInverseOf(modeus:mappa_est_descritta_da_brogliardo) owl:Thing))

EquivalentClasses(modeus:Brogliardo
ObjectSomeValuesFrom(ObjectInverseOf(modeus:rappresentazione_particella_est_descritta) owl:Thing))
SubClassOf(modeus:Brogliardo modeus:Scrittura_impianto_primaria)
EquivalentClasses(modeus:Catastino DataSomeValuesFrom(modeus:numero_pagina_catastino rdfs:Literal))
SubClassOf(modeus:Catastino modeus:Scrittura_secondaria)
EquivalentClasses(modeus:Circoscrizione ObjectSomeValuesFrom(ObjectInverseOf(modeus:comune_comprende) owl:Thing))
EquivalentClasses(modeus:Circoscrizione
ObjectSomeValuesFrom(ObjectInverseOf(modeus:organo_periferico_ha_competenza) owl:Thing))
EquivalentClasses(modeus:Circoscrizione DataSomeValuesFrom(modeus:nome_circoscrizione rdfs:Literal))
EquivalentClasses(modeus:Co_mune_sede_di_ufficio_di_registro
ObjectSomeValuesFrom(ObjectInverseOf(modeus:organo_periferico_ha_sede) owl:Thing))
SubClassOf(modeus:Co_mune_sede_di_ufficio_di_registro modeus:Comune)
EquivalentClasses(modeus:Componente_unità_documentaria
ObjectSomeValuesFrom(modeus:componente_documentaria_est_parte_di_unità owl:Thing))
EquivalentClasses(modeus:Componente_unità_documentaria
DataSomeValuesFrom(modeus:quantità_componente_documentaria rdfs:Literal))
EquivalentClasses(modeus:Componente_unità_documentaria
DataSomeValuesFrom(modeus:tipologia_componente_documentaria rdfs:Literal))
EquivalentClasses(modeus:Comune ObjectSomeValuesFrom(modeus:comune_appartiene_a owl:Thing))
EquivalentClasses(modeus:Comune ObjectSomeValuesFrom(modeus:comune_comprende owl:Thing))
EquivalentClasses(modeus:Comune ObjectSomeValuesFrom(ObjectInverseOf(modeus:mappa_est_relativa) owl:Thing))
EquivalentClasses(modeus:Comune DataSomeValuesFrom(modeus:nome_comune rdfs:Literal))
EquivalentClasses(modeus:Copia_brogliardo ObjectSomeValuesFrom(modeus:copia_brogliardo_est_destinata_a owl:Thing))
EquivalentClasses(modeus:Copia_brogliardo
ObjectSomeValuesFrom(ObjectInverseOf(modeus:brogliardo_est_riprodotto) owl:Thing))
SubClassOf(modeus:Copia_brogliardo modeus:Scrittura_secondaria)
EquivalentClasses(modeus:Delegazione ObjectSomeValuesFrom(modeus:delegazione_confina_con owl:Thing))
EquivalentClasses(modeus:Delegazione ObjectSomeValuesFrom(modeus:delegazione_est_divisa owl:Thing))
EquivalentClasses(modeus:Delegazione ObjectSomeValuesFrom(modeus:delegazione_est_inclusa owl:Thing))
EquivalentClasses(modeus:Delegazione ObjectSomeValuesFrom(ObjectInverseOf(modeus:comune_appartiene_a) owl:Thing))
EquivalentClasses(modeus:Delegazione DataSomeValuesFrom(modeus:nome_delegazione rdfs:Literal))
SubClassOf(modeus:Delegazione ObjectSomeValuesFrom(modeus:delegazione_confina_con_Stato owl:Thing))
EquivalentClasses(modeus:Destinazione ObjectUnionOf(modeus:destinazione_produttiva modeus:destinazione_uso))
EquivalentClasses(modeus:Destinazione
ObjectSomeValuesFrom(ObjectInverseOf(modeus:rappresentazione_particella_ha_destinazione) owl:Thing))
EquivalentClasses(modeus:Ente DataSomeValuesFrom(modeus:anno_istituzione_ente rdfs:Literal))
EquivalentClasses(modeus:Ente DataSomeValuesFrom(modeus:giorno_istituzione_ente rdfs:Literal))
EquivalentClasses(modeus:Ente DataSomeValuesFrom(modeus:mese_istituzione_ente rdfs:Literal))
EquivalentClasses(modeus:Ente DataSomeValuesFrom(modeus:nome_ente rdfs:Literal))
EquivalentClasses(modeus:Ente_R DataSomeValuesFrom(modeus:nome_ente_R rdfs:Literal))
EquivalentClasses(modeus:Ente_intestatario ObjectSomeValuesFrom(modeus:est_ente_intestatario_di_intestazione owl:Thing))
EquivalentClasses(modeus:Ente_intestatario DataSomeValuesFrom(modeus:domicilio_ente_intestatario rdfs:Literal))
SubClassOf(modeus:Ente_intestatario modeus:Ente)
SubClassOf(modeus:Ente_intestatario
ObjectUnionOf(ObjectSomeValuesFrom(modeus:est_ente_intestatario_padrone_diretto owl:Thing)
ObjectSomeValuesFrom(modeus:est_ente_intestatario_utilista owl:Thing)))
EquivalentClasses(modeus:Ente_intestatario_R
ObjectSomeValuesFrom(modeus:est_ente_intestatario_R_di_intestazione_R owl:Thing))
EquivalentClasses(modeus:Ente_intestatario_R
ObjectSomeValuesFrom(ObjectInverseOf(modeus:est_rappresentazione_ente_intestatario) owl:Thing))
EquivalentClasses(modeus:Ente_intestatario_R DataSomeValuesFrom(modeus:domicilio_ente_intestatario_R rdfs:Literal))
SubClassOf(modeus:Ente_intestatario_R modeus:Ente_R)

SubClassOf(modeus:Ente_intestatario_R
ObjectUnionOf(ObjectSomeValuesFrom(modeus:est_ente_intestatario_R_padrone_diretto owl:Thing)
ObjectSomeValuesFrom(modeus:est_ente_intestatario_R_utilista owl:Thing)))
EquivalentClasses(modeus:Ente_intestatario_R_padrone_diretto
ObjectSomeValuesFrom(modeus:est_ente_intestatario_R_padrone_diretto_tratta_da owl:Thing))
SubClassOf(modeus:Ente_intestatario_R_padrone_diretto modeus:Ente_intestatario_R)
EquivalentClasses(modeus:Ente_intestatario_R_pro_indiviso
ObjectSomeValuesFrom(modeus:est_ente_intestatario_R_pro_indiviso_tratta_da owl:Thing))
SubClassOf(modeus:Ente_intestatario_R_pro_indiviso modeus:Ente_intestatario_R)
EquivalentClasses(modeus:Ente_intestatario_R_singolo
ObjectSomeValuesFrom(modeus:est_ente_intestatario_R_singolo_tratta_da owl:Thing))
SubClassOf(modeus:Ente_intestatario_R_singolo modeus:Ente_intestatario_R)
EquivalentClasses(modeus:Ente_intestatario_R_utilista
ObjectSomeValuesFrom(modeus:est_ente_intestatario_R_utilista_tratta_da owl:Thing))
SubClassOf(modeus:Ente_intestatario_R_utilista modeus:Ente_intestatario_R)
EquivalentClasses(modeus:Funzione ObjectSomeValuesFrom(modeus:funzione_est_documentata owl:Thing))
EquivalentClasses(modeus:Funzione
ObjectSomeValuesFrom(ObjectInverseOf(modeus:soggetto_produttore_adempie_funzione) owl:Thing))
EquivalentClasses(modeus:Funzione DataSomeValuesFrom(modeus:forma_autorizzata_nome_funzione
rdfs:Literal))
EquivalentClasses(modeus:Funzione DataSomeValuesFrom(modeus:tipologia_funzione rdfs:Literal))
EquivalentClasses(modeus:Geometra ObjectSomeValuesFrom(modeus:est_geometra_di_mappa owl:Thing))
SubClassOf(modeus:Geometra modeus:Tecnico)
EquivalentClasses(modeus:Geometra_R ObjectSomeValuesFrom(modeus:est_geometra_R_di_mappa owl:Thing))
EquivalentClasses(modeus:Geometra_R ObjectSomeValuesFrom(modeus:est_rappresentazione_di_geometra
owl:Thing))
SubClassOf(modeus:Geometra_R modeus:Tecnico_R)
EquivalentClasses(modeus:Governo_distrettuale
ObjectSomeValuesFrom(ObjectInverseOf(modeus:delegazione_est_divisa) owl:Thing))
EquivalentClasses(modeus:Governo_distrettuale DataSomeValuesFrom(modeus:nome_governo_distrettuale
rdfs:Literal))
EquivalentClasses(modeus:Indicatore_R ObjectSomeValuesFrom(modeus:est_indicatore_R_tratta_da owl:Thing))
EquivalentClasses(modeus:Indicatore_R ObjectSomeValuesFrom(modeus:est_rappresentazione_indicatore
owl:Thing))
SubClassOf(modeus:Indicatore_R modeus:Persona_R)
EquivalentClasses(modeus:Intestazione ObjectUnionOf(modeus:Intestazione_pro_indiviso
modeus:Intestazione_singolo))
EquivalentClasses(modeus:Intestazione
ObjectSomeValuesFrom(ObjectInverseOf(modeus:rappresentazione_particella_ha_intestazione) owl:Thing))
EquivalentClasses(modeus:Intestazione DataSomeValuesFrom(modeus:anno_inizio_intestazione rdfs:Literal))
EquivalentClasses(modeus:Intestazione DataSomeValuesFrom(modeus:giorno_inizio_intestazione rdfs:Literal))
EquivalentClasses(modeus:Intestazione DataSomeValuesFrom(modeus:mese_inizio_intestazione rdfs:Literal))
EquivalentClasses(modeus:Intestazione_R ObjectUnionOf(modeus:Intestazione_R_pro_indiviso
modeus:Intestazione_R_singolo))
EquivalentClasses(modeus:Intestazione_R ObjectSomeValuesFrom(modeus:est_intestazione_R_tratta_da
owl:Thing))
EquivalentClasses(modeus:Intestazione_R ObjectSomeValuesFrom(modeus:est_rappresentazione_di_intestazione
owl:Thing))
EquivalentClasses(modeus:Intestazione_R
ObjectSomeValuesFrom(ObjectInverseOf(modeus:est_persona_intestatario_R_di_intestazione_R) owl:Thing))
EquivalentClasses(modeus:Intestazione_R
ObjectSomeValuesFrom(ObjectInverseOf(modeus:rappresentazione_particella_ha_intestazione_R) owl:Thing))
EquivalentClasses(modeus:Intestazione_R DataSomeValuesFrom(modeus:anno_inizio_intestazione_R rdfs:Literal))
EquivalentClasses(modeus:Intestazione_R DataSomeValuesFrom(modeus:giorno_inizio_intestazione_R
rdfs:Literal))
EquivalentClasses(modeus:Intestazione_R DataSomeValuesFrom(modeus:mese_inizio_intestazione_R rdfs:Literal))
DisjointClasses(modeus:Intestazione_R_pro_indiviso modeus:Intestazione_R_singolo)
DisjointClasses(modeus:Intestazione_pro_indiviso modeus:Intestazione_singolo)
EquivalentClasses(modeus:Isituto_conservatore
ObjectSomeValuesFrom(modeus:istituto_conservatore_est_ubicato_fisicamente owl:Thing))
EquivalentClasses(modeus:Isituto_conservatore
ObjectSomeValuesFrom(ObjectInverseOf(modeus:unità_di_descrizione_est_conservata) owl:Thing))

EquivalentClasses(modeus:Isituto_conservatore DataSomeValuesFrom(modeus:forma_autorizzata_nome_istituto_conservatore rdfs:Literal))
EquivalentClasses(modeus:Isituto_conservatore DataSomeValuesFrom(modeus:sito_web_istituto_conservatore rdfs:Literal))
EquivalentClasses(modeus:Istanza_voltura ObjectSomeValuesFrom(modeus:istanza_est_registrata_in owl:Thing))
EquivalentClasses(modeus:Istanza_voltura ObjectSomeValuesFrom(modeus:istanza_voltura_est_associata owl:Thing))
EquivalentClasses(modeus:Istanza_voltura DataSomeValuesFrom(modeus:numero_progressivo_istanza_voltura rdfs:Literal))
SubClassOf(modeus:Istanza_voltura modeus:Scrittura_secondaria)
SubClassOf(modeus:Legazione modeus:Delegazione)
EquivalentClasses(modeus:Mappa ObjectSomeValuesFrom(modeus:est_mappa_di_scrittura_primaria owl:Thing))
EquivalentClasses(modeus:Mappa ObjectSomeValuesFrom(modeus:mappa_est_descritta_da_brogliardo owl:Thing))
EquivalentClasses(modeus:Mappa ObjectSomeValuesFrom(modeus:mappa_est_relativa owl:Thing))
EquivalentClasses(modeus:Mappa ObjectSomeValuesFrom(ObjectInverseOf(modeus:est_aiutante_R_di_mappa) owl:Thing))
EquivalentClasses(modeus:Mappa ObjectSomeValuesFrom(ObjectInverseOf(modeus:est_aiutante_R_tratta_da) owl:Thing))
EquivalentClasses(modeus:Mappa ObjectSomeValuesFrom(ObjectInverseOf(modeus:est_aiutante_di_mappa) owl:Thing))
EquivalentClasses(modeus:Mappa ObjectSomeValuesFrom(ObjectInverseOf(modeus:est_assistente_R_mappa) owl:Thing))
EquivalentClasses(modeus:Mappa ObjectSomeValuesFrom(ObjectInverseOf(modeus:est_assistente_R_tratta_da) owl:Thing))
EquivalentClasses(modeus:Mappa ObjectSomeValuesFrom(ObjectInverseOf(modeus:est_assistente_di_mappa) owl:Thing))
EquivalentClasses(modeus:Mappa ObjectSomeValuesFrom(ObjectInverseOf(modeus:est_geometra_R_di_mappa) owl:Thing))
EquivalentClasses(modeus:Mappa ObjectSomeValuesFrom(ObjectInverseOf(modeus:est_geometra_di_mappa) owl:Thing))
EquivalentClasses(modeus:Mappa ObjectSomeValuesFrom(ObjectInverseOf(modeus:est_indicatore_R_di_mappa) owl:Thing))
EquivalentClasses(modeus:Mappa ObjectSomeValuesFrom(ObjectInverseOf(modeus:est_indicatore_R_tratta_da) owl:Thing))
EquivalentClasses(modeus:Mappa ObjectSomeValuesFrom(ObjectInverseOf(modeus:est_indicatore_di_mappa) owl:Thing))
EquivalentClasses(modeus:Mappa ObjectSomeValuesFrom(ObjectInverseOf(modeus:est_ispettore_R_di_mappa) owl:Thing))
EquivalentClasses(modeus:Mappa ObjectSomeValuesFrom(ObjectInverseOf(modeus:est_ispettore_di_mappa) owl:Thing))
EquivalentClasses(modeus:Mappa ObjectSomeValuesFrom(ObjectInverseOf(modeus:est_verificatore_R_di_mappa) owl:Thing))
EquivalentClasses(modeus:Mappa ObjectSomeValuesFrom(ObjectInverseOf(modeus:est_verificatore_di_mappa) owl:Thing))
EquivalentClasses(modeus:Mappa ObjectSomeValuesFrom(ObjectInverseOf(modeus:sezione_catastale_est_rappresentata_graficamente) owl:Thing))
EquivalentClasses(modeus:Mappa DataSomeValuesFrom(modeus:scala_mappa rdfs:Literal))
SubClassOf(modeus:Mappa modeus:Scrittura_impianto_primaria)
SubClassOf(modeus:Mappa ObjectIntersectionOf(ObjectMinCardinality(3 modeus:mappa_est_riprodotta_in) ObjectMaxCardinality(3 modeus:mappa_est_riprodotta_in)))
EquivalentClasses(modeus:Mappa_copia_scala_originale ObjectSomeValuesFrom(modeus:mappa_copia_scala_originale_est_destinata_a owl:Thing))
SubClassOf(modeus:Mappa_copia_scala_originale modeus:Scrittura_secondaria)
EquivalentClasses(modeus:Mappa_copia_scala_originale_digitalizzata DataSomeValuesFrom(modeus:segnatura_mappa_copia_originale_digitalizzata rdfs:Literal))
EquivalentClasses(modeus:Mappa_digitalizzata DataSomeValuesFrom(modeus:segnatura_mappa_digitalizzata rdfs:Literal))
EquivalentClasses(modeus:Mappetta_carta ObjectSomeValuesFrom(modeus:mappetta_carta_est_destinata_a owl:Thing))
EquivalentClasses(modeus:Mappetta_carta DataSomeValuesFrom(modeus:scala_mappetta_carta rdfs:Literal))
SubClassOf(modeus:Mappetta_carta modeus:Scrittura_secondaria)

EquivalentClasses(modeus:Mappetta_tela ObjectSomeValuesFrom(modeus:est_mappetta_tela_di_scrittura_primaria owl:Thing))
EquivalentClasses(modeus:Mappetta_tela DataSomeValuesFrom(modeus:scala_mappetta_tela rdfs:Literal))
SubClassOf(modeus:Mappetta_tela modeus:Scrittura_impianto_primaria)
EquivalentClasses(modeus:Notaio ObjectSomeValuesFrom(ObjectInverseOf(modeus:atto_notarile_est_redatto) owl:Thing))
SubClassOf(modeus:Notaio modeus:Persona)
EquivalentClasses(modeus:Oggetto_fisico ObjectSomeValuesFrom(ObjectInverseOf(modeus:sezione_catastale_contiene_oggetto_fisico) owl:Thing))
EquivalentClasses(modeus:Organo_periferico ObjectSomeValuesFrom(modeus:organo_periferico_ha_competenza owl:Thing))
EquivalentClasses(modeus:Organo_periferico ObjectSomeValuesFrom(modeus:organo_periferico_ha_sede owl:Thing))
EquivalentClasses(modeus:Organo_periferico ObjectSomeValuesFrom(ObjectInverseOf(modeus:copia_brogliardo_est_destinata_a) owl:Thing))
EquivalentClasses(modeus:Organo_periferico ObjectSomeValuesFrom(ObjectInverseOf(modeus:mappa_copia_scala_originale_est_destinata_a) owl:Thing))
EquivalentClasses(modeus:Organo_periferico ObjectSomeValuesFrom(ObjectInverseOf(modeus:mappetta_carta_est_destinata_a) owl:Thing))
SubClassOf(modeus:Organo_periferico modeus:Ente)
EquivalentClasses(modeus:Persona ObjectSomeValuesFrom(ObjectInverseOf(modeus:relazione_paternità) owl:Thing))
EquivalentClasses(modeus:Persona DataSomeValuesFrom(modeus:anno_di_nascita rdfs:Literal))
EquivalentClasses(modeus:Persona DataSomeValuesFrom(modeus:cognome_persona rdfs:Literal))
EquivalentClasses(modeus:Persona DataSomeValuesFrom(modeus:giorno_di_nascita rdfs:Literal))
EquivalentClasses(modeus:Persona DataSomeValuesFrom(modeus:luogo_di_nascita rdfs:Literal))
EquivalentClasses(modeus:Persona DataSomeValuesFrom(modeus:mese_di_nascita rdfs:Literal))
EquivalentClasses(modeus:Persona DataSomeValuesFrom(modeus:nome_persona rdfs:Literal))
EquivalentClasses(modeus:Persona_R ObjectSomeValuesFrom(modeus:est_rappresentazione_di_persona owl:Thing))
EquivalentClasses(modeus:Persona_R DataSomeValuesFrom(modeus:cognome_persona_R rdfs:Literal))
EquivalentClasses(modeus:Persona_R DataSomeValuesFrom(modeus:nome_persona_R rdfs:Literal))
EquivalentClasses(modeus:Persona_intestatario ObjectSomeValuesFrom(modeus:est_persona_intestatario_di_intestazione owl:Thing))
EquivalentClasses(modeus:Persona_intestatario DataSomeValuesFrom(modeus:domicilio_persona_intestatario rdfs:Literal))
SubClassOf(modeus:Persona_intestatario modeus:Persona)
SubClassOf(modeus:Persona_intestatario ObjectUnionOf(ObjectSomeValuesFrom(modeus:est_persona_intestatario_padrone_diretto owl:Thing) ObjectSomeValuesFrom(modeus:est_persona_intestatario_utilista owl:Thing)))
EquivalentClasses(modeus:Persona_intestatario_R ObjectSomeValuesFrom(modeus:est_figlio owl:Thing))
EquivalentClasses(modeus:Persona_intestatario_R ObjectSomeValuesFrom(modeus:est_persona_intestatario_R_di_intestazione_R owl:Thing))
EquivalentClasses(modeus:Persona_intestatario_R ObjectSomeValuesFrom(modeus:est_rappresentazione_di_persona_intestatario owl:Thing))
EquivalentClasses(modeus:Persona_intestatario_R DataSomeValuesFrom(modeus:domicilio_persona_intestatario_R rdfs:Literal))
SubClassOf(modeus:Persona_intestatario_R modeus:Persona_R)
SubClassOf(modeus:Persona_intestatario_R ObjectUnionOf(ObjectSomeValuesFrom(modeus:est_persona_intestatario_R_padrone_diretto owl:Thing) ObjectSomeValuesFrom(modeus:est_persona_intestatario_R_utilista owl:Thing)))
EquivalentClasses(modeus:Persona_intestatario_R_padrone_diretto ObjectSomeValuesFrom(modeus:est_persona_intestatario_R_padrone_diretto_tratta_da owl:Thing))
SubClassOf(modeus:Persona_intestatario_R_padrone_diretto modeus:Persona_intestatario_R)
EquivalentClasses(modeus:Persona_intestatario_R_pro_indiviso ObjectSomeValuesFrom(modeus:est_persona_intestatario_R_pro_indiviso_tratta_da owl:Thing))
SubClassOf(modeus:Persona_intestatario_R_pro_indiviso modeus:Persona_intestatario_R)
EquivalentClasses(modeus:Persona_intestatario_R_singolo ObjectSomeValuesFrom(modeus:est_persona_intestatario_R_singolo_tratta_da owl:Thing))
SubClassOf(modeus:Persona_intestatario_R_singolo modeus:Persona_intestatario_R)
EquivalentClasses(modeus:Persona_intestatario_R_utilista ObjectSomeValuesFrom(modeus:est_persona_intestatario_R_utilista_tratta_da owl:Thing))

SubClassOf(modeus:Persona_intestatario_R_utilista modeus:Persona_intestatario_R)
EquivalentClasses(modeus:Persona_intestatario_moglie_R ObjectSomeValuesFrom(modeus:est_moglie owl:Thing))
SubClassOf(modeus:Persona_intestatario_moglie_R modeus:Persona_intestatario_R)
EquivalentClasses(modeus:Persona_intestatario_vedova_R ObjectSomeValuesFrom(modeus:est_vedova owl:Thing))
SubClassOf(modeus:Persona_intestatario_vedova_R modeus:Persona_intestatario_R)
EquivalentClasses(modeus:Rappresentazione_cartacea_mappa
ObjectUnionOf(modeus:Mappa_copia_scala_originale modeus:Mappetta_carta modeus:Mappetta_tela))
EquivalentClasses(modeus:Rappresentazione_cartacea_mappa
ObjectSomeValuesFrom(ObjectInverseOf(modeus:mappa_est_riprodotta_in) owl:Thing))
EquivalentClasses(modeus:Rappresentazione_particella
ObjectSomeValuesFrom(modeus:rappresentazione_particella_est_descritta owl:Thing))
EquivalentClasses(modeus:Rappresentazione_particella
ObjectSomeValuesFrom(modeus:rappresentazione_particella_ha_destinazione owl:Thing))
EquivalentClasses(modeus:Rappresentazione_particella
ObjectSomeValuesFrom(modeus:rappresentazione_particella_ha_intestazione owl:Thing))
EquivalentClasses(modeus:Rappresentazione_particella
ObjectSomeValuesFrom(modeus:rappresentazione_particella_ha_intestazione_R owl:Thing))
EquivalentClasses(modeus:Rappresentazione_particella
ObjectSomeValuesFrom(ObjectInverseOf(modeus:sezione_catastale_contiene_rappresentazione_particella)
owl:Thing))
EquivalentClasses(modeus:Rappresentazione_particella
DataSomeValuesFrom(modeus:anno_nascita_rappresentazione_particella rdfs:Literal))
EquivalentClasses(modeus:Rappresentazione_particella DataSomeValuesFrom(modeus:estensione_di_particella
rdfs:Literal))
EquivalentClasses(modeus:Rappresentazione_particella DataSomeValuesFrom(modeus:estimo_di_particella
rdfs:Literal))
EquivalentClasses(modeus:Rappresentazione_particella
DataSomeValuesFrom(modeus:giorno_nascita_rappresentazione_particella rdfs:Literal))
EquivalentClasses(modeus:Rappresentazione_particella
DataSomeValuesFrom(modeus:mappale_principale_di_particella rdfs:Literal))
EquivalentClasses(modeus:Rappresentazione_particella
DataSomeValuesFrom(modeus:mese_nascita_rappresentazione_particella rdfs:Literal))
EquivalentClasses(modeus:Rappresentazione_particella DataSomeValuesFrom(modeus:nome_contrada_di_particella
rdfs:Literal))
SubClassOf(modeus:Rappresentazione_particella modeus:Oggetto_fisico)
EquivalentClasses(modeus:Registro_istanza_voltura
ObjectSomeValuesFrom(ObjectInverseOf(modeus:istanza_est_registrata_in) owl:Thing))
SubClassOf(modeus:Registro_istanza_voltura modeus:Scrittura_secondaria)
EquivalentClasses(modeus:Registro_trasporti DataSomeValuesFrom(modeus:numero_mutazione_registro_trasporti
rdfs:Literal))
EquivalentClasses(modeus:Registro_trasporti DataSomeValuesFrom(modeus:numero_pagina_registro_trasporti
rdfs:Literal))
SubClassOf(modeus:Registro_trasporti modeus:Scrittura_secondaria)
EquivalentClasses(modeus:Relazione_persona_intestatario_R
ObjectSomeValuesFrom(modeus:est_relazione_persona_intestatario_R_tratta_da owl:Thing))
SubClassOf(modeus:Relazione_persona_intestatario_R
ObjectUnionOf(ObjectSomeValuesFrom(ObjectInverseOf(modeus:est_figlio) owl:Thing)
ObjectSomeValuesFrom(ObjectInverseOf(modeus:est_moglie) owl:Thing)
ObjectSomeValuesFrom(ObjectInverseOf(modeus:est_vedova) owl:Thing)))
SubClassOf(modeus:Relazione_persona_intestatario_R
ObjectUnionOf(ObjectSomeValuesFrom(ObjectInverseOf(modeus:est_marito_di_vedova) owl:Thing)
ObjectSomeValuesFrom(ObjectInverseOf(modeus:est_padre) owl:Thing)))
EquivalentClasses(modeus:Scrittura_impianto_primaria
ObjectSomeValuesFrom(ObjectInverseOf(modeus:est_brogliardo_di_scrittura_primaria) owl:Thing))
EquivalentClasses(modeus:Scrittura_impianto_primaria
ObjectSomeValuesFrom(ObjectInverseOf(modeus:est_mappa_di_scrittura_primaria) owl:Thing))
EquivalentClasses(modeus:Scrittura_impianto_primaria
ObjectSomeValuesFrom(ObjectInverseOf(modeus:est_mappetta_tela_di_scrittura_primaria) owl:Thing))
SubClassOf(modeus:Scrittura_impianto_primaria modeus:Unità_di_descrizione)
SubClassOf(modeus:Scrittura_secondaria modeus:Unità_di_descrizione)
EquivalentClasses(modeus:Sede
ObjectSomeValuesFrom(ObjectInverseOf(modeus:istituto_conservatore_est_ubicato_fisicamente) owl:Thing))

EquivalentClasses(modeus:Sede DataSomeValuesFrom(modeus:città_sede rdfs:Literal))
EquivalentClasses(modeus:Sede DataSomeValuesFrom(modeus:indirizzo_sede rdfs:Literal))
EquivalentClasses(modeus:Sezione_catastale
ObjectSomeValuesFrom(modeus:sezione_catastale_contiene_oggetto_fisico owl:Thing))
EquivalentClasses(modeus:Sezione_catastale
ObjectSomeValuesFrom(modeus:sezione_catastale_contiene_rappresentazione_particella owl:Thing))
EquivalentClasses(modeus:Sezione_catastale
ObjectSomeValuesFrom(modeus:sezione_catastale_est_rappresentata_graficamente owl:Thing))
EquivalentClasses(modeus:Sezione_catastale DataSomeValuesFrom(modeus:nome_sezione_catastale rdfs:Literal))
EquivalentClasses(modeus:Soggetto_produttore ObjectSomeValuesFrom(modeus:est_soggetto_produttore_di owl:Thing))
EquivalentClasses(modeus:Soggetto_produttore
ObjectSomeValuesFrom(modeus:soggetto_produttore_adempie_funzione owl:Thing))
EquivalentClasses(modeus:Soggetto_produttore
ObjectSomeValuesFrom(ObjectInverseOf(modeus:agente_est_soggetto_produttore) owl:Thing))
EquivalentClasses(modeus:Soggetto_produttore
DataSomeValuesFrom(modeus:forma_autorizzata_nome_soggetto_produttore rdfs:Literal))
EquivalentClasses(modeus:Stato ObjectSomeValuesFrom(ObjectInverseOf(modeus:delegazione_est_inclusa) owl:Thing))
EquivalentClasses(modeus:Stato DataSomeValuesFrom(modeus:nome_stato rdfs:Literal))
SubClassOf(modeus:Stato ObjectSomeValuesFrom(ObjectInverseOf(modeus:delegazione_confina_con_Stato) owl:Thing))
EquivalentClasses(modeus:Strada DataSomeValuesFrom(modeus:tipo_strada rdfs:Literal))
EquivalentClasses(modeus:Successione_intestazione_R
ObjectSomeValuesFrom(modeus:est_successione_intestazione_R_tratta_da owl:Thing))
EquivalentClasses(modeus:Successione_intestazione_R
ObjectSomeValuesFrom(ObjectInverseOf(modeus:est_intestazione_R_precedente) owl:Thing))
EquivalentClasses(modeus:Successione_intestazione_R
ObjectSomeValuesFrom(ObjectInverseOf(modeus:est_intestazione_R_successiva) owl:Thing))
EquivalentClasses(modeus:Tecnico ObjectSomeValuesFrom(modeus:est_ispettore_di_mappa owl:Thing))
EquivalentClasses(modeus:Tecnico ObjectSomeValuesFrom(modeus:est_verificatore_di_mappa owl:Thing))
SubClassOf(modeus:Tecnico modeus:Persona)
EquivalentClasses(modeus:Tecnico_R ObjectSomeValuesFrom(modeus:est_ispettore_R_di_mappa owl:Thing))
EquivalentClasses(modeus:Tecnico_R ObjectSomeValuesFrom(modeus:est_rappresentazione_di_tecnico owl:Thing))
EquivalentClasses(modeus:Tecnico_R ObjectSomeValuesFrom(modeus:est_tecnico_R_tratta_da owl:Thing))
EquivalentClasses(modeus:Tecnico_R ObjectSomeValuesFrom(modeus:est_verificatore_R_di_mappa owl:Thing))
SubClassOf(modeus:Tecnico_R modeus:Persona_R)
EquivalentClasses(modeus:Unità_di_descrizione
ObjectSomeValuesFrom(modeus:unità_di_descrizione_est_conservata owl:Thing))
EquivalentClasses(modeus:Unità_di_descrizione ObjectSomeValuesFrom(modeus:unità_di_descrizione_ha_livello owl:Thing))
EquivalentClasses(modeus:Unità_di_descrizione
ObjectSomeValuesFrom(ObjectInverseOf(modeus:est_soggetto_produttore_di) owl:Thing))
EquivalentClasses(modeus:Unità_di_descrizione
ObjectSomeValuesFrom(ObjectInverseOf(modeus:funzione_est_documentata) owl:Thing))
EquivalentClasses(modeus:Unità_di_descrizione DataSomeValuesFrom(modeus:anno_formazione_unità rdfs:Literal))
EquivalentClasses(modeus:Unità_di_descrizione DataSomeValuesFrom(modeus:anno_inizio_range_unità rdfs:Literal))
EquivalentClasses(modeus:Unità_di_descrizione DataSomeValuesFrom(modeus:consistenza_unità_descrizione rdfs:Literal))
EquivalentClasses(modeus:Unità_di_descrizione DataSomeValuesFrom(modeus:giorno_formazione_unità rdfs:Literal))
EquivalentClasses(modeus:Unità_di_descrizione DataSomeValuesFrom(modeus:giorno_inizio_range_unità rdfs:Literal))
EquivalentClasses(modeus:Unità_di_descrizione DataSomeValuesFrom(modeus:mese_formazione_unità rdfs:Literal))
EquivalentClasses(modeus:Unità_di_descrizione DataSomeValuesFrom(modeus:mese_inizio_range_unità rdfs:Literal))
EquivalentClasses(modeus:Unità_di_descrizione DataSomeValuesFrom(modeus:segnatura_unità_descrizione rdfs:Literal))

EquivalentClasses(modeus:Unità_di_descrizione DataSomeValuesFrom(modeus:supporto_unità_descrizione rdfs:Literal))
EquivalentClasses(modeus:Unità_di_descrizione DataSomeValuesFrom(modeus:tipologia_documentaria_unità_descrizione rdfs:Literal))
EquivalentClasses(modeus:Unità_di_descrizione DataSomeValuesFrom(modeus:titolo_unità_descrizione rdfs:Literal))
EquivalentClasses(modeus:Unità_documentaria DataSomeValuesFrom(modeus:stato_unità_documentaria rdfs:Literal))
EquivalentClasses(modeus:destinazione_produttiva DataSomeValuesFrom(modeus:nome_destinazione_produttiva rdfs:Literal))
DisjointClasses(modeus:destinazione_produttiva modeus:destinazione_uso)
EquivalentClasses(modeus:destinazione_uso DataSomeValuesFrom(modeus:nome_destinazione_uso rdfs:Literal))
ClassAssertion(modeus:Isituto_conservatore modeus:Archivio_di_Stato_Roma)
ClassAssertion(modeus:Organo_periferico modeus:Cancelleria_censo_Tivoli)
ClassAssertion(modeus:Soggetto_produttore modeus:Cancelleria_censo_Tivoli)
ClassAssertion(modeus:Soggetto_produttore modeus:Direzione_generale_catasti)
ClassAssertion(modeus:Livello_di_descrizione modeus:Fondo)
ClassAssertion(modeus:Soggetto_produttore modeus:Presidenza_del_censo)
ClassAssertion(modeus:Livello_di_descrizione modeus:Serie)
ClassAssertion(modeus:Livello_di_descrizione modeus:Sottoserie)
ClassAssertion(modeus:Livello_di_descrizione modeus:Sub_fondo)
ClassAssertion(modeus:Livello_di_descrizione modeus:Unità_archivistica)
ClassAssertion(modeus:Livello_di_descrizione modeus:Unità_documentaria)
EquivalentClasses(ObjectUnionOf(modeus:Brogliardo modeus:Catastino modeus:Istanza_voltura modeus:Registro_trasporti)
ObjectSomeValuesFrom(ObjectInverseOf(modeus:est_ente_intestatario_R_padrone_diretto_tratta_da) owl:Thing))
EquivalentClasses(ObjectUnionOf(modeus:Brogliardo modeus:Catastino modeus:Istanza_voltura modeus:Registro_trasporti)
ObjectSomeValuesFrom(ObjectInverseOf(modeus:est_ente_intestatario_R_utilista_tratta_da) owl:Thing))
EquivalentClasses(ObjectUnionOf(modeus:Brogliardo modeus:Catastino modeus:Istanza_voltura modeus:Registro_trasporti)
ObjectSomeValuesFrom(ObjectInverseOf(modeus:est_persona_intestatario_R_padrone_diretto_tratta_da) owl:Thing))
EquivalentClasses(ObjectUnionOf(modeus:Brogliardo modeus:Catastino modeus:Istanza_voltura modeus:Registro_trasporti)
ObjectSomeValuesFrom(ObjectInverseOf(modeus:est_persona_intestatario_R_utilista_tratta_da) owl:Thing))
EquivalentClasses(ObjectUnionOf(modeus:Brogliardo modeus:Catastino modeus:Istanza_voltura modeus:Registro_trasporti)
ObjectSomeValuesFrom(ObjectInverseOf(modeus:est_relazione_persona_intestatario_R_tratta_da) owl:Thing))
EquivalentClasses(ObjectUnionOf(modeus:Brogliardo modeus:Mappa)
ObjectSomeValuesFrom(ObjectInverseOf(modeus:est_tecnico_R_tratta_da) owl:Thing))
EquivalentClasses(ObjectUnionOf(modeus:Brogliardo modeus:Registro_trasporti)
ObjectSomeValuesFrom(ObjectInverseOf(modeus:est_ente_intestatario_R_pro_indiviso_tratta_da) owl:Thing))
EquivalentClasses(ObjectUnionOf(modeus:Brogliardo modeus:Registro_trasporti)
ObjectSomeValuesFrom(ObjectInverseOf(modeus:est_ente_intestatario_R_singolo_tratta_da) owl:Thing))
EquivalentClasses(ObjectUnionOf(modeus:Brogliardo modeus:Registro_trasporti)
ObjectSomeValuesFrom(ObjectInverseOf(modeus:est_persona_intestatario_R_pro_indiviso_tratta_da) owl:Thing))
EquivalentClasses(ObjectUnionOf(modeus:Brogliardo modeus:Registro_trasporti)
ObjectSomeValuesFrom(ObjectInverseOf(modeus:est_persona_intestatario_R_singolo_tratta_da) owl:Thing))
EquivalentClasses(ObjectUnionOf(ObjectSomeValuesFrom(modeus:rappresentazione_particella_est_frazionata_in owl:Thing) ObjectSomeValuesFrom(modeus:rappresentazione_particella_est_fusa_in owl:Thing))
DataSomeValuesFrom(modeus:anno_morte_rappresentazione_particella rdfs:Literal))
EquivalentClasses(ObjectUnionOf(ObjectSomeValuesFrom(modeus:rappresentazione_particella_est_frazionata_in owl:Thing) ObjectSomeValuesFrom(modeus:rappresentazione_particella_est_fusa_in owl:Thing))
DataSomeValuesFrom(modeus:giorno_morte_rappresentazione_particella rdfs:Literal))
EquivalentClasses(ObjectUnionOf(ObjectSomeValuesFrom(modeus:rappresentazione_particella_est_frazionata_in owl:Thing) ObjectSomeValuesFrom(modeus:rappresentazione_particella_est_fusa_in owl:Thing))
DataSomeValuesFrom(modeus:mese_morte_rappresentazione_particella rdfs:Literal))
SubClassOf(ObjectUnionOf(modeus:Acqua modeus:Strada) modeus:Oggetto_fisico)
SubClassOf(ObjectUnionOf(modeus:Brogliardo modeus:Catastino modeus:Istanza_voltura modeus:Registro_trasporti) ObjectSomeValuesFrom(ObjectInverseOf(modeus:est_intestazione_R_tratta_da) owl:Thing))

```
SubClassOf(ObjectUnionOf(modeus:Catastino modeus:Copia_brogliardo modeus:Istanza_voltura
modeus:Mappa_copia_scala_originale modeus:Mappetta_carta modeus:Registro_istanza_voltura
modeus:Registro_trasporti) modeus:Scrittura_secondaria)
SubClassOf(ObjectUnionOf(modeus:Ente modeus:Famiglia modeus:Persona) modeus:Agente)
SubClassOf(ObjectUnionOf(modeus:Fondo modeus:Serie modeus:Sottoserie modeus:Sub_fondo
modeus:Unità_archivistica modeus:Unità_documentaria) modeus:Unità_di_descrizione)
SubClassOf(ObjectUnionOf(modeus:Istanza_voltura modeus:Registro_trasporti)
ObjectSomeValuesFrom(ObjectInverseOf(modeus:est_successione_intestazione_R_tratta_da) owl:Thing))
DisjointClasses(modeus:Catastino modeus:Copia_brogliardo modeus:Istanza_voltura
modeus:Mappa_copia_scala_originale modeus:Mappetta_carta modeus:Registro_istanza_voltura
modeus:Registro_trasporti)
DisjointClasses(modeus:Ente modeus:Famiglia modeus:Persona)
DisjointClasses(modeus:Fondo modeus:Serie modeus:Sottoserie modeus:Sub_fondo modeus:Unità_archivistica
modeus:Unità_documentaria)
DisjointClasses(modeus:Mappa_copia_scala_originale modeus:Mappetta_carta modeus:Mappetta_tela)
)
```

*File OWL: ABox*

```
Prefix(:=<http://modeus.uniroma1.it/abox#>)
Prefix(owl:=<http://www.w3.org/2002/07/owl#>)
Prefix(rdf:=<http://www.w3.org/1999/02/22-rdf-syntax-ns#>)
Prefix(xml:=<http://www.w3.org/XML/1998/namespace>)
Prefix(xsd:=<http://www.w3.org/2001/XMLSchema#>)
Prefix(rdfs:=<http://www.w3.org/2000/01/rdf-schema#>)
Ontology(<http://modeus.uniroma1.it/abox>
Declaration(Class(<http://modeus.uniroma1.it/ontology#Acqua>))
Declaration(Class(<http://modeus.uniroma1.it/ontology#Agente>))
Declaration(Class(<http://modeus.uniroma1.it/ontology#Aiutante_R>))
Declaration(Class(<http://modeus.uniroma1.it/ontology#Assistente_R>))
Declaration(Class(<http://modeus.uniroma1.it/ontology#Atto_notarile>))
Declaration(Class(<http://modeus.uniroma1.it/ontology#Brogliardo>))
Declaration(Class(<http://modeus.uniroma1.it/ontology#Brogliardo_digitalizzato>))
Declaration(Class(<http://modeus.uniroma1.it/ontology#Catastino>))
Declaration(Class(<http://modeus.uniroma1.it/ontology#Circoscrizione>))
Declaration(Class(<http://modeus.uniroma1.it/ontology#Co_mune_sede_di_ufficio_di_registro>))
Declaration(Class(<http://modeus.uniroma1.it/ontology#Componente_unità_documentaria>))
Declaration(Class(<http://modeus.uniroma1.it/ontology#Comune>))
Declaration(Class(<http://modeus.uniroma1.it/ontology#Copia_brogliardo>))
Declaration(Class(<http://modeus.uniroma1.it/ontology#Delegazione>))
Declaration(Class(<http://modeus.uniroma1.it/ontology#Destinazione>))
Declaration(Class(<http://modeus.uniroma1.it/ontology#Ente>))
Declaration(Class(<http://modeus.uniroma1.it/ontology#Ente_intestatario>))
Declaration(Class(<http://modeus.uniroma1.it/ontology#Ente_intestatario_R>))
Declaration(Class(<http://modeus.uniroma1.it/ontology#Ente_intestatario_R_padrone_diretto>))
Declaration(Class(<http://modeus.uniroma1.it/ontology#Ente_intestatario_R_pro_indiviso>))
Declaration(Class(<http://modeus.uniroma1.it/ontology#Ente_intestatario_R_singolo>))
Declaration(Class(<http://modeus.uniroma1.it/ontology#Ente_intestatario_R_utilista>))
Declaration(Class(<http://modeus.uniroma1.it/ontology#Famiglia>))
Declaration(Class(<http://modeus.uniroma1.it/ontology#Fondo>))
Declaration(Class(<http://modeus.uniroma1.it/ontology#Funzione>))
Declaration(Class(<http://modeus.uniroma1.it/ontology#Geometra>))
Declaration(Class(<http://modeus.uniroma1.it/ontology#Geometra_R>))
Declaration(Class(<http://modeus.uniroma1.it/ontology#Governo_distrettuale>))
Declaration(Class(<http://modeus.uniroma1.it/ontology#Indicatore_R>))
Declaration(Class(<http://modeus.uniroma1.it/ontology#Intestazione>))
Declaration(Class(<http://modeus.uniroma1.it/ontology#Intestazione_R>))
Declaration(Class(<http://modeus.uniroma1.it/ontology#Intestazione_R_pro_indiviso>))
Declaration(Class(<http://modeus.uniroma1.it/ontology#Intestazione_R_singolo>))
Declaration(Class(<http://modeus.uniroma1.it/ontology#Intestazione_pro_indiviso>))
Declaration(Class(<http://modeus.uniroma1.it/ontology#Intestazione_singolo>))
```

Declaration(Class(<http://modeus.uniroma1.it/ontology#Isituto_conservatore>))
Declaration(Class(<http://modeus.uniroma1.it/ontology#Istanza_voltura>))
Declaration(Class(<http://modeus.uniroma1.it/ontology#Legazione>))
Declaration(Class(<http://modeus.uniroma1.it/ontology#Livello_di_descrizione>))
Declaration(Class(<http://modeus.uniroma1.it/ontology#Mappa>))
Declaration(Class(<http://modeus.uniroma1.it/ontology#Mappa_copia_scala_originale>))
Declaration(Class(<http://modeus.uniroma1.it/ontology#Mappa_copia_scala_originale_digitalizzata>))
Declaration(Class(<http://modeus.uniroma1.it/ontology#Mappa_digitalizzata>))
Declaration(Class(<http://modeus.uniroma1.it/ontology#Mappetta_carta>))
Declaration(Class(<http://modeus.uniroma1.it/ontology#Mappetta_tela>))
Declaration(Class(<http://modeus.uniroma1.it/ontology#Notaio>))
Declaration(Class(<http://modeus.uniroma1.it/ontology#Oggetto_fisico>))
Declaration(Class(<http://modeus.uniroma1.it/ontology#Organo_periferico>))
Declaration(Class(<http://modeus.uniroma1.it/ontology#Persona>))
Declaration(Class(<http://modeus.uniroma1.it/ontology#Persona_R>))
Declaration(Class(<http://modeus.uniroma1.it/ontology#Persona_intestatario>))
Declaration(Class(<http://modeus.uniroma1.it/ontology#Persona_intestatario_R>))
Declaration(Class(<http://modeus.uniroma1.it/ontology#Persona_intestatario_R_padrone_diretto>))
Declaration(Class(<http://modeus.uniroma1.it/ontology#Persona_intestatario_R_pro_indiviso>))
Declaration(Class(<http://modeus.uniroma1.it/ontology#Persona_intestatario_R_singolo>))
Declaration(Class(<http://modeus.uniroma1.it/ontology#Persona_intestatario_R_utilista>))
Declaration(Class(<http://modeus.uniroma1.it/ontology#Persona_intestatario_moglie_R>))
Declaration(Class(<http://modeus.uniroma1.it/ontology#Persona_intestatario_vedova_R>))
Declaration(Class(<http://modeus.uniroma1.it/ontology#Rappresentazione_cartacea_mappa>))
Declaration(Class(<http://modeus.uniroma1.it/ontology#Rappresentazione_particella>))
Declaration(Class(<http://modeus.uniroma1.it/ontology#Registro_istanza_voltura>))
Declaration(Class(<http://modeus.uniroma1.it/ontology#Registro_trasporti>))
Declaration(Class(<http://modeus.uniroma1.it/ontology#Relazione_persona_intestatario_R>))
Declaration(Class(<http://modeus.uniroma1.it/ontology#Scrittura_impianto_primaria>))
Declaration(Class(<http://modeus.uniroma1.it/ontology#Scrittura_secondaria>))
Declaration(Class(<http://modeus.uniroma1.it/ontology#Sede>))
Declaration(Class(<http://modeus.uniroma1.it/ontology#Serie>))
Declaration(Class(<http://modeus.uniroma1.it/ontology#Sezione_catastale>))
Declaration(Class(<http://modeus.uniroma1.it/ontology#Soggetto_produttore>))
Declaration(Class(<http://modeus.uniroma1.it/ontology#Sottoserie>))
Declaration(Class(<http://modeus.uniroma1.it/ontology#Stato>))
Declaration(Class(<http://modeus.uniroma1.it/ontology#Strada>))
Declaration(Class(<http://modeus.uniroma1.it/ontology#Sub_fondo>))
Declaration(Class(<http://modeus.uniroma1.it/ontology#Successione_intestazione_R>))
Declaration(Class(<http://modeus.uniroma1.it/ontology#Successione_livello_descrizione>))
Declaration(Class(<http://modeus.uniroma1.it/ontology#Tecnico>))
Declaration(Class(<http://modeus.uniroma1.it/ontology#Tecnico_R>))
Declaration(Class(<http://modeus.uniroma1.it/ontology#Unità_archivistica>))
Declaration(Class(<http://modeus.uniroma1.it/ontology#Unità_di_descrizione>))
Declaration(Class(<http://modeus.uniroma1.it/ontology#Unità_documentaria>))
Declaration(Class(<http://modeus.uniroma1.it/ontology#destinazione_produttiva>))
Declaration(Class(<http://modeus.uniroma1.it/ontology#destinazione_uso>))
Declaration(ObjectProperty(<http://modeus.uniroma1.it/ontology#agente_est_soggetto_produttore>))
Declaration(ObjectProperty(<http://modeus.uniroma1.it/ontology#atto_notarile_est_redatto>))
Declaration(ObjectProperty(<http://modeus.uniroma1.it/ontology#brogliardo_est_riprodotto>))
Declaration(ObjectProperty(<http://modeus.uniroma1.it/ontology#brogliardo_ha_riproduzione_digitale>))
Declaration(ObjectProperty(<http://modeus.uniroma1.it/ontology#componente_documentaria_est_parte_di_unità
>))
Declaration(ObjectProperty(<http://modeus.uniroma1.it/ontology#comune_appartiene_a>))
Declaration(ObjectProperty(<http://modeus.uniroma1.it/ontology#comune_comprende>))
Declaration(ObjectProperty(<http://modeus.uniroma1.it/ontology#confina_con_delegazione>))
Declaration(ObjectProperty(<http://modeus.uniroma1.it/ontology#copia_brogliardo_est_destinata_a>))
Declaration(ObjectProperty(<http://modeus.uniroma1.it/ontology#delegazione_confina_con>))
Declaration(ObjectProperty(<http://modeus.uniroma1.it/ontology#delegazione_confina_con_Stato>))
Declaration(ObjectProperty(<http://modeus.uniroma1.it/ontology#delegazione_est_divisa>))
Declaration(ObjectProperty(<http://modeus.uniroma1.it/ontology#delegazione_est_inclusa>))
Declaration(ObjectProperty(<http://modeus.uniroma1.it/ontology#est_aiutante_R_di_mappa>))

```
Declaration(ObjectProperty(<http://modeus.uniroma1.it/ontology#est_aiutante_R_tratta_da>))
Declaration(ObjectProperty(<http://modeus.uniroma1.it/ontology#est_aiutante_di_mappa>))
Declaration(ObjectProperty(<http://modeus.uniroma1.it/ontology#est_assistente_R_mappa>))
Declaration(ObjectProperty(<http://modeus.uniroma1.it/ontology#est_assistente_R_tratta_da>))
Declaration(ObjectProperty(<http://modeus.uniroma1.it/ontology#est_assistente_di_mappa>))
Declaration(ObjectProperty(<http://modeus.uniroma1.it/ontology#est_brogliardo_di_scrittura_primaria>))
Declaration(ObjectProperty(<http://modeus.uniroma1.it/ontology#est_ente_intestatario_R_di_intestazione_R>))
Declaration(ObjectProperty(<http://modeus.uniroma1.it/ontology#est_ente_intestatario_R_padrone_diretto>))
Declaration(ObjectProperty(<http://modeus.uniroma1.it/ontology#est_ente_intestatario_R_padrone_diretto_tratta
_da>))
Declaration(ObjectProperty(<http://modeus.uniroma1.it/ontology#est_ente_intestatario_R_pro_indiviso>))
Declaration(ObjectProperty(<http://modeus.uniroma1.it/ontology#est_ente_intestatario_R_pro_indiviso_tratta_d
a>))
Declaration(ObjectProperty(<http://modeus.uniroma1.it/ontology#est_ente_intestatario_R_singolo>))
Declaration(ObjectProperty(<http://modeus.uniroma1.it/ontology#est_ente_intestatario_R_singolo_tratta_da>))
Declaration(ObjectProperty(<http://modeus.uniroma1.it/ontology#est_ente_intestatario_R_utilista>))
Declaration(ObjectProperty(<http://modeus.uniroma1.it/ontology#est_ente_intestatario_R_utilista_tratta_da>))
Declaration(ObjectProperty(<http://modeus.uniroma1.it/ontology#est_ente_intestatario_di_intestazione>))
Declaration(ObjectProperty(<http://modeus.uniroma1.it/ontology#est_ente_intestatario_padrone_diretto>))
Declaration(ObjectProperty(<http://modeus.uniroma1.it/ontology#est_ente_intestatario_pro_indiviso>))
Declaration(ObjectProperty(<http://modeus.uniroma1.it/ontology#est_ente_intestatario_singolo>))
Declaration(ObjectProperty(<http://modeus.uniroma1.it/ontology#est_ente_intestatario_utilista>))
Declaration(ObjectProperty(<http://modeus.uniroma1.it/ontology#est_figlio>))
Declaration(ObjectProperty(<http://modeus.uniroma1.it/ontology#est_geometra_R_di_mappa>))
Declaration(ObjectProperty(<http://modeus.uniroma1.it/ontology#est_geometra_R_tratta_da>))
Declaration(ObjectProperty(<http://modeus.uniroma1.it/ontology#est_geometra_di_mappa>))
Declaration(ObjectProperty(<http://modeus.uniroma1.it/ontology#est_indicatore_R_di_mappa>))
Declaration(ObjectProperty(<http://modeus.uniroma1.it/ontology#est_indicatore_R_tratta_da>))
Declaration(ObjectProperty(<http://modeus.uniroma1.it/ontology#est_indicatore_di_mappa>))
Declaration(ObjectProperty(<http://modeus.uniroma1.it/ontology#est_intestazione_R_precedente>))
Declaration(ObjectProperty(<http://modeus.uniroma1.it/ontology#est_intestazione_R_successiva>))
Declaration(ObjectProperty(<http://modeus.uniroma1.it/ontology#est_intestazione_R_tratta_da>))
Declaration(ObjectProperty(<http://modeus.uniroma1.it/ontology#est_ispettore_R_di_mappa>))
Declaration(ObjectProperty(<http://modeus.uniroma1.it/ontology#est_ispettore_di_mappa>))
Declaration(ObjectProperty(<http://modeus.uniroma1.it/ontology#est_livello_descrizione_precedente>))
Declaration(ObjectProperty(<http://modeus.uniroma1.it/ontology#est_livello_descrizione_successivo>))
Declaration(ObjectProperty(<http://modeus.uniroma1.it/ontology#est_mappa_di_scrittura_primaria>))
Declaration(ObjectProperty(<http://modeus.uniroma1.it/ontology#est_mappetta_tela_di_scrittura_primaria>))
Declaration(ObjectProperty(<http://modeus.uniroma1.it/ontology#est_marito_di_moglie>))
Declaration(ObjectProperty(<http://modeus.uniroma1.it/ontology#est_marito_di_vedova>))
Declaration(ObjectProperty(<http://modeus.uniroma1.it/ontology#est_moglie>))
Declaration(ObjectProperty(<http://modeus.uniroma1.it/ontology#est_padre>))
Declaration(ObjectProperty(<http://modeus.uniroma1.it/ontology#est_parte_di_livello_di_descrizione>))
Declaration(ObjectProperty(<http://modeus.uniroma1.it/ontology#est_persona_intestatario_R_di_intestazione_R
>))
Declaration(ObjectProperty(<http://modeus.uniroma1.it/ontology#est_persona_intestatario_R_padrone_diretto>)
)
Declaration(ObjectProperty(<http://modeus.uniroma1.it/ontology#est_persona_intestatario_R_padrone_diretto_tr
atta_da>))
Declaration(ObjectProperty(<http://modeus.uniroma1.it/ontology#est_persona_intestatario_R_pro_indiviso>))
Declaration(ObjectProperty(<http://modeus.uniroma1.it/ontology#est_persona_intestatario_R_pro_indiviso_tratt
a_da>))
Declaration(ObjectProperty(<http://modeus.uniroma1.it/ontology#est_persona_intestatario_R_singolo>))
Declaration(ObjectProperty(<http://modeus.uniroma1.it/ontology#est_persona_intestatario_R_singolo_tratta_da
>))
Declaration(ObjectProperty(<http://modeus.uniroma1.it/ontology#est_persona_intestatario_R_utilista>))
Declaration(ObjectProperty(<http://modeus.uniroma1.it/ontology#est_persona_intestatario_R_utilista_tratta_da>
))
Declaration(ObjectProperty(<http://modeus.uniroma1.it/ontology#est_persona_intestatario_di_intestazione>))
Declaration(ObjectProperty(<http://modeus.uniroma1.it/ontology#est_persona_intestatario_padrone_diretto>))
Declaration(ObjectProperty(<http://modeus.uniroma1.it/ontology#est_persona_intestatario_pro_indiviso>))
Declaration(ObjectProperty(<http://modeus.uniroma1.it/ontology#est_persona_intestatario_singolo>))
```

Declaration(ObjectProperty(<http://modeus.uniroma1.it/ontology#est_persona_intestatario_utilista>))
Declaration(ObjectProperty(<http://modeus.uniroma1.it/ontology#est_rappresentazione_aiutante>))
Declaration(ObjectProperty(<http://modeus.uniroma1.it/ontology#est_rappresentazione_assistente>))
Declaration(ObjectProperty(<http://modeus.uniroma1.it/ontology#est_rappresentazione_di_geometra>))
Declaration(ObjectProperty(<http://modeus.uniroma1.it/ontology#est_rappresentazione_di_intestazione>))
Declaration(ObjectProperty(<http://modeus.uniroma1.it/ontology#est_rappresentazione_di_persona>))
Declaration(ObjectProperty(<http://modeus.uniroma1.it/ontology#est_rappresentazione_di_persona_intestatario>))
Declaration(ObjectProperty(<http://modeus.uniroma1.it/ontology#est_rappresentazione_di_tecnico>))
Declaration(ObjectProperty(<http://modeus.uniroma1.it/ontology#est_rappresentazione_ente_intestatario>))
Declaration(ObjectProperty(<http://modeus.uniroma1.it/ontology#est_rappresentazione_indicatore>))
Declaration(ObjectProperty(<http://modeus.uniroma1.it/ontology#est_relazione_persona_intestatario_R_tratta_da>))
Declaration(ObjectProperty(<http://modeus.uniroma1.it/ontology#est_soggetto_produttore_di>))
Declaration(ObjectProperty(<http://modeus.uniroma1.it/ontology#est_successione_intestazione_R_tratta_da>))
Declaration(ObjectProperty(<http://modeus.uniroma1.it/ontology#est_tecnico_R_tratta_da>))
Declaration(ObjectProperty(<http://modeus.uniroma1.it/ontology#est_vedova>))
Declaration(ObjectProperty(<http://modeus.uniroma1.it/ontology#est_verificatore_R_di_mappa>))
Declaration(ObjectProperty(<http://modeus.uniroma1.it/ontology#est_verificatore_di_mappa>))
Declaration(ObjectProperty(<http://modeus.uniroma1.it/ontology#funzione_est_documentata>))
Declaration(ObjectProperty(<http://modeus.uniroma1.it/ontology#istanza_est_registrata_in>))
Declaration(ObjectProperty(<http://modeus.uniroma1.it/ontology#istanza_voltura_est_associata>))
Declaration(ObjectProperty(<http://modeus.uniroma1.it/ontology#istituto_conservatore_est_ubicato_fisicamente>))
Declaration(ObjectProperty(<http://modeus.uniroma1.it/ontology#mappa_copia_scala_originale_digitalizzata_est_legata_a_mappa_digitalizzata>))
Declaration(ObjectProperty(<http://modeus.uniroma1.it/ontology#mappa_copia_scala_originale_est_destinata_a>))
Declaration(ObjectProperty(<http://modeus.uniroma1.it/ontology#mappa_copia_scala_originale_ha_riproduzione_digitale>))
Declaration(ObjectProperty(<http://modeus.uniroma1.it/ontology#mappa_digitalizzata_est_descritta_da_brogliardo_digitalizzato>))
Declaration(ObjectProperty(<http://modeus.uniroma1.it/ontology#mappa_est_descritta_da_brogliardo>))
Declaration(ObjectProperty(<http://modeus.uniroma1.it/ontology#mappa_est_relativa>))
Declaration(ObjectProperty(<http://modeus.uniroma1.it/ontology#mappa_est_riprodotta_in>))
Declaration(ObjectProperty(<http://modeus.uniroma1.it/ontology#mappa_ha_riproduzione_digitale>))
Declaration(ObjectProperty(<http://modeus.uniroma1.it/ontology#mappetta_carta_est_destinata_a>))
Declaration(ObjectProperty(<http://modeus.uniroma1.it/ontology#organo_periferico_ha_competenza>))
Declaration(ObjectProperty(<http://modeus.uniroma1.it/ontology#organo_periferico_ha_sede>))
Declaration(ObjectProperty(<http://modeus.uniroma1.it/ontology#rappresentazione_particella_confina_con>))
Declaration(ObjectProperty(<http://modeus.uniroma1.it/ontology#rappresentazione_particella_confina_con_oggetto_fisico>))
Declaration(ObjectProperty(<http://modeus.uniroma1.it/ontology#rappresentazione_particella_est_descritta>))
Declaration(ObjectProperty(<http://modeus.uniroma1.it/ontology#rappresentazione_particella_est_frazionata_in>))
Declaration(ObjectProperty(<http://modeus.uniroma1.it/ontology#rappresentazione_particella_est_fusa_in>))
Declaration(ObjectProperty(<http://modeus.uniroma1.it/ontology#rappresentazione_particella_ha_destinazione>))
Declaration(ObjectProperty(<http://modeus.uniroma1.it/ontology#rappresentazione_particella_ha_intestazione>))
Declaration(ObjectProperty(<http://modeus.uniroma1.it/ontology#rappresentazione_particella_ha_intestazione_R>))
Declaration(ObjectProperty(<http://modeus.uniroma1.it/ontology#relazione_coniugio>))
Declaration(ObjectProperty(<http://modeus.uniroma1.it/ontology#relazione_paternità>))
Declaration(ObjectProperty(<http://modeus.uniroma1.it/ontology#sezione_catastale_contiene_oggetto_fisico>))
Declaration(ObjectProperty(<http://modeus.uniroma1.it/ontology#sezione_catastale_contiene_rappresentazione_particella>))
Declaration(ObjectProperty(<http://modeus.uniroma1.it/ontology#sezione_catastale_est_rappresentata_graficamente>))
Declaration(ObjectProperty(<http://modeus.uniroma1.it/ontology#soggetto_produttore_adempie_funzione>))
Declaration(ObjectProperty(<http://modeus.uniroma1.it/ontology#successione_intestazione>))
Declaration(ObjectProperty(<http://modeus.uniroma1.it/ontology#unità_di_descrizione_est_conservata>))
Declaration(ObjectProperty(<http://modeus.uniroma1.it/ontology#unità_di_descrizione_ha_livello>))

```
Declaration(DataProperty(<http://modeus.uniroma1.it/ontology#anno_aggiornamento_mappa_copia>))
Declaration(DataProperty(<http://modeus.uniroma1.it/ontology#anno_di_morte>))
Declaration(DataProperty(<http://modeus.uniroma1.it/ontology#anno_di_nascita>))
Declaration(DataProperty(<http://modeus.uniroma1.it/ontology#anno_fine_intestazione>))
Declaration(DataProperty(<http://modeus.uniroma1.it/ontology#anno_fine_intestazione_R>))
Declaration(DataProperty(<http://modeus.uniroma1.it/ontology#anno_fine_range_unità>))
Declaration(DataProperty(<http://modeus.uniroma1.it/ontology#anno_formazione_unità>))
Declaration(DataProperty(<http://modeus.uniroma1.it/ontology#anno_inizio_intestazione>))
Declaration(DataProperty(<http://modeus.uniroma1.it/ontology#anno_inizio_intestazione_R>))
Declaration(DataProperty(<http://modeus.uniroma1.it/ontology#anno_inizio_range_unità>))
Declaration(DataProperty(<http://modeus.uniroma1.it/ontology#anno_istituzione_ente>))
Declaration(DataProperty(<http://modeus.uniroma1.it/ontology#anno_morte_rappresentazione_particella>))
Declaration(DataProperty(<http://modeus.uniroma1.it/ontology#anno_nascita_rappresentazione_particella>))
Declaration(DataProperty(<http://modeus.uniroma1.it/ontology#anno_redazione_atto>))
Declaration(DataProperty(<http://modeus.uniroma1.it/ontology#anno_registrazione_atto>))
Declaration(DataProperty(<http://modeus.uniroma1.it/ontology#anno_soppressione_ente>))
Declaration(DataProperty(<http://modeus.uniroma1.it/ontology#città_sede>))
Declaration(DataProperty(<http://modeus.uniroma1.it/ontology#codice_identificativo_conservatore>))
Declaration(DataProperty(<http://modeus.uniroma1.it/ontology#codice_identificativo_descrizione_funzione>))
Declaration(DataProperty(<http://modeus.uniroma1.it/ontology#codice_identificativo_record_autorità>))
Declaration(DataProperty(<http://modeus.uniroma1.it/ontology#codice_identificativo_unità_descrizione>))
Declaration(DataProperty(<http://modeus.uniroma1.it/ontology#cognome_persona>))
Declaration(DataProperty(<http://modeus.uniroma1.it/ontology#cognome_persona_R>))
Declaration(DataProperty(<http://modeus.uniroma1.it/ontology#consistenza_unità_descrizione>))
Declaration(DataProperty(<http://modeus.uniroma1.it/ontology#domicilio_ente_intestatario>))
Declaration(DataProperty(<http://modeus.uniroma1.it/ontology#domicilio_ente_intestatario_R>))
Declaration(DataProperty(<http://modeus.uniroma1.it/ontology#domicilio_persona_intestatario>))
Declaration(DataProperty(<http://modeus.uniroma1.it/ontology#domicilio_persona_intestatario_R>))
Declaration(DataProperty(<http://modeus.uniroma1.it/ontology#estensione_di_particella>))
Declaration(DataProperty(<http://modeus.uniroma1.it/ontology#estimo_di_particella>))
Declaration(DataProperty(<http://modeus.uniroma1.it/ontology#forma_autorizzata_nome_funzione>))
Declaration(DataProperty(<http://modeus.uniroma1.it/ontology#forma_autorizzata_nome_istituto_conservatore>
))
Declaration(DataProperty(<http://modeus.uniroma1.it/ontology#forma_autorizzata_nome_soggetto_produttore>
))
Declaration(DataProperty(<http://modeus.uniroma1.it/ontology#giorno_aggiornamento_mappa_copia>))
Declaration(DataProperty(<http://modeus.uniroma1.it/ontology#giorno_di_morte>))
Declaration(DataProperty(<http://modeus.uniroma1.it/ontology#giorno_di_nascita>))
Declaration(DataProperty(<http://modeus.uniroma1.it/ontology#giorno_fine_intestazione>))
Declaration(DataProperty(<http://modeus.uniroma1.it/ontology#giorno_fine_intestazione_R>))
Declaration(DataProperty(<http://modeus.uniroma1.it/ontology#giorno_fine_range_unità>))
Declaration(DataProperty(<http://modeus.uniroma1.it/ontology#giorno_formazione_unità>))
Declaration(DataProperty(<http://modeus.uniroma1.it/ontology#giorno_inizio_intestazione>))
Declaration(DataProperty(<http://modeus.uniroma1.it/ontology#giorno_inizio_intestazione_R>))
Declaration(DataProperty(<http://modeus.uniroma1.it/ontology#giorno_inizio_range_unità>))
Declaration(DataProperty(<http://modeus.uniroma1.it/ontology#giorno_istituzione_ente>))
Declaration(DataProperty(<http://modeus.uniroma1.it/ontology#giorno_morte_rappresentazione_particella>))
Declaration(DataProperty(<http://modeus.uniroma1.it/ontology#giorno_nascita_rappresentazione_particella>))
Declaration(DataProperty(<http://modeus.uniroma1.it/ontology#giorno_redazione_atto>))
Declaration(DataProperty(<http://modeus.uniroma1.it/ontology#giorno_registrazione_atto>))
Declaration(DataProperty(<http://modeus.uniroma1.it/ontology#giorno_soppressione_ente>))
Declaration(DataProperty(<http://modeus.uniroma1.it/ontology#indirizzo_sede>))
Declaration(DataProperty(<http://modeus.uniroma1.it/ontology#indirizzo_sede_principale>))
Declaration(DataProperty(<http://modeus.uniroma1.it/ontology#indirizzo_sede_succursale>))
Declaration(DataProperty(<http://modeus.uniroma1.it/ontology#luogo_di_morte>))
Declaration(DataProperty(<http://modeus.uniroma1.it/ontology#luogo_di_nascita>))
Declaration(DataProperty(<http://modeus.uniroma1.it/ontology#mappale_principale_di_particella>))
Declaration(DataProperty(<http://modeus.uniroma1.it/ontology#mappale_subordinato_di_particella>))
Declaration(DataProperty(<http://modeus.uniroma1.it/ontology#mese_aggiornamento_mappa_copia>))
Declaration(DataProperty(<http://modeus.uniroma1.it/ontology#mese_di_morte>))
Declaration(DataProperty(<http://modeus.uniroma1.it/ontology#mese_di_nascita>))
Declaration(DataProperty(<http://modeus.uniroma1.it/ontology#mese_fine_intestazione>))
```

Declaration(DataProperty(<http://modeus.uniroma1.it/ontology#mese_fine_intestazione_R>))
Declaration(DataProperty(<http://modeus.uniroma1.it/ontology#mese_fine_range_unità>))
Declaration(DataProperty(<http://modeus.uniroma1.it/ontology#mese_formazione_unità>))
Declaration(DataProperty(<http://modeus.uniroma1.it/ontology#mese_inizio_intestazione>))
Declaration(DataProperty(<http://modeus.uniroma1.it/ontology#mese_inizio_intestazione_R>))
Declaration(DataProperty(<http://modeus.uniroma1.it/ontology#mese_inizio_range_unità>))
Declaration(DataProperty(<http://modeus.uniroma1.it/ontology#mese_istituzione_ente>))
Declaration(DataProperty(<http://modeus.uniroma1.it/ontology#mese_morte_rappresentazione_particella>))
Declaration(DataProperty(<http://modeus.uniroma1.it/ontology#mese_nascita_rappresentazione_particella>))
Declaration(DataProperty(<http://modeus.uniroma1.it/ontology#mese_redazione_atto>))
Declaration(DataProperty(<http://modeus.uniroma1.it/ontology#mese_registrazione_atto>))
Declaration(DataProperty(<http://modeus.uniroma1.it/ontology#mese_soppressione_ente>))
Declaration(DataProperty(<http://modeus.uniroma1.it/ontology#nome_acqua>))
Declaration(DataProperty(<http://modeus.uniroma1.it/ontology#nome_circoscrizione>))
Declaration(DataProperty(<http://modeus.uniroma1.it/ontology#nome_comune>))
Declaration(DataProperty(<http://modeus.uniroma1.it/ontology#nome_contrada_di_particella>))
Declaration(DataProperty(<http://modeus.uniroma1.it/ontology#nome_delegazione>))
Declaration(DataProperty(<http://modeus.uniroma1.it/ontology#nome_destinazione_produttiva>))
Declaration(DataProperty(<http://modeus.uniroma1.it/ontology#nome_destinazione_uso>))
Declaration(DataProperty(<http://modeus.uniroma1.it/ontology#nome_ente>))
Declaration(DataProperty(<http://modeus.uniroma1.it/ontology#nome_ente_intestatario_R>))
Declaration(DataProperty(<http://modeus.uniroma1.it/ontology#nome_governo_distrettuale>))
Declaration(DataProperty(<http://modeus.uniroma1.it/ontology#nome_persona>))
Declaration(DataProperty(<http://modeus.uniroma1.it/ontology#nome_persona_R>))
Declaration(DataProperty(<http://modeus.uniroma1.it/ontology#nome_sezione_catastale>))
Declaration(DataProperty(<http://modeus.uniroma1.it/ontology#nome_stato>))
Declaration(DataProperty(<http://modeus.uniroma1.it/ontology#nome_strada>))
Declaration(DataProperty(<http://modeus.uniroma1.it/ontology#numero_mutazione_registro_trasporti>))
Declaration(DataProperty(<http://modeus.uniroma1.it/ontology#numero_pagina_catastino>))
Declaration(DataProperty(<http://modeus.uniroma1.it/ontology#numero_pagina_registro_trasporti>))
Declaration(DataProperty(<http://modeus.uniroma1.it/ontology#numero_progressivo_istanza_voltura>))
Declaration(DataProperty(<http://modeus.uniroma1.it/ontology#numero_protocollo_registrazione_istanza_voltura>))
Declaration(DataProperty(<http://modeus.uniroma1.it/ontology#numero_sezione_catastale>))
Declaration(DataProperty(<http://modeus.uniroma1.it/ontology#qualifica_tecnico_R>))
Declaration(DataProperty(<http://modeus.uniroma1.it/ontology#quantità_componente_documentaria>))
Declaration(DataProperty(<http://modeus.uniroma1.it/ontology#scala_mappa>))
Declaration(DataProperty(<http://modeus.uniroma1.it/ontology#scala_mappetta_carta>))
Declaration(DataProperty(<http://modeus.uniroma1.it/ontology#scala_mappetta_tela>))
Declaration(DataProperty(<http://modeus.uniroma1.it/ontology#segnatura_mappa_copia_originale_digitalizzata>))
Declaration(DataProperty(<http://modeus.uniroma1.it/ontology#segnatura_mappa_digitalizzata>))
Declaration(DataProperty(<http://modeus.uniroma1.it/ontology#segnatura_unità_descrizione>))
Declaration(DataProperty(<http://modeus.uniroma1.it/ontology#sito_web_istituto_conservatore>))
Declaration(DataProperty(<http://modeus.uniroma1.it/ontology#stato_unità_documentaria>))
Declaration(DataProperty(<http://modeus.uniroma1.it/ontology#supporto_unità_descrizione>))
Declaration(DataProperty(<http://modeus.uniroma1.it/ontology#tipo_acqua>))
Declaration(DataProperty(<http://modeus.uniroma1.it/ontology#tipo_strada>))
Declaration(DataProperty(<http://modeus.uniroma1.it/ontology#tipologia_componente_documentaria>))
Declaration(DataProperty(<http://modeus.uniroma1.it/ontology#tipologia_documentaria_unità_descrizione>))
Declaration(DataProperty(<http://modeus.uniroma1.it/ontology#tipologia_funzione>))
Declaration(DataProperty(<http://modeus.uniroma1.it/ontology#titolo_attribuito_unità_descrizione>))
Declaration(DataProperty(<http://modeus.uniroma1.it/ontology#titolo_originale_unità_descrizione>))
Declaration(DataProperty(<http://modeus.uniroma1.it/ontology#titolo_unità_descrizione>))
Declaration(NamedIndividual(:Catastino1))
Declaration(NamedIndividual(:Catastino2))
Declaration(NamedIndividual(:IntestazioneR1))
Declaration(NamedIndividual(:IntestazioneR2))
Declaration(NamedIndividual(:IntestazioneR3))
Declaration(NamedIndividual(:bro1))
Declaration(NamedIndividual(:destprod1))
Declaration(NamedIndividual(:destprod2))

```
Declaration(NamedIndividual(:destprod3))
Declaration(NamedIndividual(:destprod4))
Declaration(NamedIndividual(:geom1))
Declaration(NamedIndividual(:mappa1))
Declaration(NamedIndividual(:mappadigit1))
Declaration(NamedIndividual(:part1))
Declaration(NamedIndividual(:part2))
Declaration(NamedIndividual(:part3))
Declaration(NamedIndividual(:part4))
Declaration(NamedIndividual(:part5))
Declaration(NamedIndividual(:part6))
Declaration(NamedIndividual(:part7))
Declaration(NamedIndividual(:pers_i1))
Declaration(NamedIndividual(:pers_i2))
Declaration(NamedIndividual(:sez1))
Declaration(NamedIndividual(:voltura1))

#############################
#   Named Individuals
#############################

# Individual: :Catastino1 (:Catastino1)
ClassAssertion(<http://modeus.uniroma1.it/ontology#Catastino> :Catastino1)
DataPropertyAssertion(<http://modeus.uniroma1.it/ontology#numero_pagina_catastino> :Catastino1
"1464"^^xsd:int)
DataPropertyAssertion(<http://modeus.uniroma1.it/ontology#segnatura_unità_descrizione> :Catastino1 "ASR,
Cancelleria del censo di Tivoli, Catastino Rustico, Vol. III, 3566"^^xsd:string)
# Individual: :Catastino2 (:Catastino2)
ClassAssertion(<http://modeus.uniroma1.it/ontology#Catastino> :Catastino2)
DataPropertyAssertion(<http://modeus.uniroma1.it/ontology#numero_pagina_catastino> :Catastino2
"1798"^^xsd:int)
DataPropertyAssertion(<http://modeus.uniroma1.it/ontology#segnatura_unità_descrizione> :Catastino2 "ASR,
Cancelleria del censo di Tivoli, Catastino Rustico, Vol. III, 3566"^^xsd:string)
# Individual: :IntestazioneR1 (:IntestazioneR1)
ClassAssertion(<http://modeus.uniroma1.it/ontology#Intestazione_R> :IntestazioneR1)
ObjectPropertyAssertion(<http://modeus.uniroma1.it/ontology#est_intestazione_R_tratta_da> :IntestazioneR1
:voltura1)
# Individual: :IntestazioneR2 (:IntestazioneR2)
ClassAssertion(<http://modeus.uniroma1.it/ontology#Intestazione_R> :IntestazioneR2)
ObjectPropertyAssertion(<http://modeus.uniroma1.it/ontology#est_intestazione_R_tratta_da> :IntestazioneR2
:Catastino1)
# Individual: :IntestazioneR3 (:IntestazioneR3)
ClassAssertion(<http://modeus.uniroma1.it/ontology#Intestazione_R> :IntestazioneR3)
ObjectPropertyAssertion(<http://modeus.uniroma1.it/ontology#est_intestazione_R_tratta_da> :IntestazioneR3
:Catastino2)
# Individual: :bro1 (:bro1)
ClassAssertion(<http://modeus.uniroma1.it/ontology#Brogliardo> :bro1)
DataPropertyAssertion(<http://modeus.uniroma1.it/ontology#segnatura_unità_descrizione> :bro1 "ASR,
Presidenza generale del censo, Archivio delle mappe e carte censuarie, Catasto Gregoriano, Comarca di Roma,
Tivoli, sezione I- Città di Tivoli, brogliardo 140"^^rdfs:Literal)
# Individual: :destprod1 (:destprod1)

ClassAssertion(<http://modeus.uniroma1.it/ontology#destinazione_produttiva> :destprod1)
DataPropertyAssertion(<http://modeus.uniroma1.it/ontology#nome_destinazione_produttiva> :destprod1
"Orto"^^rdfs:Literal)
# Individual: :destprod2 (:destprod2)
ClassAssertion(<http://modeus.uniroma1.it/ontology#destinazione_produttiva> :destprod2)
DataPropertyAssertion(<http://modeus.uniroma1.it/ontology#nome_destinazione_produttiva> :destprod2
"Seminativo"^^rdfs:Literal)
# Individual: :destprod3 (:destprod3)
ClassAssertion(<http://modeus.uniroma1.it/ontology#destinazione_produttiva> :destprod3)
```

DataPropertyAssertion(<http://modeus.uniroma1.it/ontology#nome_destinazione_produttiva> :destprod3 "Vigna"^^rdfs:Literal)
# Individual: :destprod4 (:destprod4)
ClassAssertion(<http://modeus.uniroma1.it/ontology#destinazione_produttiva> :destprod4)
DataPropertyAssertion(<http://modeus.uniroma1.it/ontology#nome_destinazione_produttiva> :destprod4 "Vigna"^^rdfs:Literal)
# Individual: :geom1 (:geom1)
ClassAssertion(<http://modeus.uniroma1.it/ontology#Geometra_R> :geom1)
ObjectPropertyAssertion(<http://modeus.uniroma1.it/ontology#est_geometra_R_di_mappa> :geom1 :mappa1)
DataPropertyAssertion(<http://modeus.uniroma1.it/ontology#cognome_persona_R> :geom1 "Marconi"^^rdfs:Literal)
DataPropertyAssertion(<http://modeus.uniroma1.it/ontology#nome_persona_R> :geom1 "Cajo"^^rdfs:Literal)
# Individual: :mappa1 (:mappa1)
ClassAssertion(<http://modeus.uniroma1.it/ontology#Mappa> :mappa1)
ObjectPropertyAssertion(<http://modeus.uniroma1.it/ontology#mappa_ha_riproduzione_digitale> :mappa1 :mappadigit1)
DataPropertyAssertion(<http://modeus.uniroma1.it/ontology#anno_fine_range_unità> :mappa1 "1820"^^xsd:int)
DataPropertyAssertion(<http://modeus.uniroma1.it/ontology#anno_inizio_range_unità> :mappa1 "1819"^^xsd:int)
DataPropertyAssertion(<http://modeus.uniroma1.it/ontology#giorno_fine_range_unità> :mappa1 "8"^^xsd:int)
DataPropertyAssertion(<http://modeus.uniroma1.it/ontology#giorno_inizio_range_unità> :mappa1 "29"^^xsd:int)
DataPropertyAssertion(<http://modeus.uniroma1.it/ontology#mese_fine_range_unità> :mappa1 "marzo"^^rdfs:Literal)
DataPropertyAssertion(<http://modeus.uniroma1.it/ontology#mese_inizio_range_unità> :mappa1 "novembre"^^rdfs:Literal)
DataPropertyAssertion(<http://modeus.uniroma1.it/ontology#segnatura_unità_descrizione> :mappa1 "ASR, Presidenza generale del censo, Archivio delle mappe e carte censuarie, Catasto Gregoriano, Comarca di Roma, Tivoli, sezione I - Città di Tivoli, mappa 140"^^rdfs:Literal)
# Individual: :mappadigit1 (:mappadigit1)
ClassAssertion(<http://modeus.uniroma1.it/ontology#Mappa_digitalizzata> :mappadigit1)
DataPropertyAssertion(<http://modeus.uniroma1.it/ontology#segnatura_mappa_digitalizzata> :mappadigit1 "COMARCA 140"^^xsd:string)
# Individual: :part1 (:part1)
ClassAssertion(<http://modeus.uniroma1.it/ontology#Rappresentazione_particella> :part1)
ObjectPropertyAssertion(<http://modeus.uniroma1.it/ontology#rappresentazione_particella_est_descritta> :part1 :bro1)
ObjectPropertyAssertion(<http://modeus.uniroma1.it/ontology#rappresentazione_particella_ha_destinazione> :part1 :destprod1)
ObjectPropertyAssertion(<http://modeus.uniroma1.it/ontology#rappresentazione_particella_ha_intestazione_R> :part1 :IntestazioneR1)
DataPropertyAssertion(<http://modeus.uniroma1.it/ontology#estensione_di_particella> :part1 "56 centesimi"^^xsd:string)
DataPropertyAssertion(<http://modeus.uniroma1.it/ontology#estimo_di_particella> :part1 "35 scudi e 35 baiocchi"^^xsd:string)
DataPropertyAssertion(<http://modeus.uniroma1.it/ontology#mappale_principale_di_particella> :part1 "809"^^xsd:int)
DataPropertyAssertion(<http://modeus.uniroma1.it/ontology#nome_contrada_di_particella> :part1 "Veste"^^rdfs:Literal)
# Individual: :part2 (:part2)
ClassAssertion(<http://modeus.uniroma1.it/ontology#Rappresentazione_particella> :part2)
ObjectPropertyAssertion(<http://modeus.uniroma1.it/ontology#rappresentazione_particella_ha_intestazione_R> :part2 :IntestazioneR2)
DataPropertyAssertion(<http://modeus.uniroma1.it/ontology#mappale_principale_di_particella> :part2 "1122"^^xsd:int)
# Individual: :part3 (:part3)
ClassAssertion(<http://modeus.uniroma1.it/ontology#Rappresentazione_particella> :part3)
DataPropertyAssertion(<http://modeus.uniroma1.it/ontology#mappale_principale_di_particella> :part3 "1124"^^xsd:int)
# Individual: :part4 (:part4)
ClassAssertion(<http://modeus.uniroma1.it/ontology#Rappresentazione_particella> :part4)
DataPropertyAssertion(<http://modeus.uniroma1.it/ontology#mappale_principale_di_particella> :part4 "189"^^xsd:int)

```
# Individual: :part5 (:part5)
ClassAssertion(<http://modeus.uniroma1.it/ontology#Rappresentazione_particella> :part5)
ObjectPropertyAssertion(<http://modeus.uniroma1.it/ontology#rappresentazione_particella_ha_destinazione>
:part5 :destprod2)
ObjectPropertyAssertion(<http://modeus.uniroma1.it/ontology#rappresentazione_particella_ha_intestazione_R>
:part5 :IntestazioneR3)
DataPropertyAssertion(<http://modeus.uniroma1.it/ontology#mappale_principale_di_particella> :part5
"255"^^xsd:int)
# Individual: :part6 (:part6)
ClassAssertion(<http://modeus.uniroma1.it/ontology#Rappresentazione_particella> :part6)
ObjectPropertyAssertion(<http://modeus.uniroma1.it/ontology#rappresentazione_particella_ha_destinazione>
:part6 :destprod3)
DataPropertyAssertion(<http://modeus.uniroma1.it/ontology#mappale_principale_di_particella> :part6
"597"^^xsd:int)
# Individual: :part7 (:part7)
ClassAssertion(<http://modeus.uniroma1.it/ontology#Rappresentazione_particella> :part7)
ObjectPropertyAssertion(<http://modeus.uniroma1.it/ontology#rappresentazione_particella_ha_destinazione>
:part7 :destprod4)
DataPropertyAssertion(<http://modeus.uniroma1.it/ontology#mappale_principale_di_particella> :part7
"229"^^xsd:int)
# Individual: :pers_i1 (:pers_i1)
ClassAssertion(<http://modeus.uniroma1.it/ontology#Persona_intestatario_R_padrone_diretto> :pers_i1)
ObjectPropertyAssertion(<http://modeus.uniroma1.it/ontology#est_persona_intestatario_R_di_intestazione_R>
:pers_i1 :IntestazioneR1)
ObjectPropertyAssertion(<http://modeus.uniroma1.it/ontology#est_persona_intestatario_R_padrone_diretto>
:pers_i1 :part1)
ObjectPropertyAssertion(<http://modeus.uniroma1.it/ontology#est_persona_intestatario_R_padrone_diretto>
:pers_i1 :part2)
ObjectPropertyAssertion(<http://modeus.uniroma1.it/ontology#est_persona_intestatario_R_padrone_diretto>
:pers_i1 :part3)
ObjectPropertyAssertion(<http://modeus.uniroma1.it/ontology#est_persona_intestatario_R_padrone_diretto>
:pers_i1 :part4)
DataPropertyAssertion(<http://modeus.uniroma1.it/ontology#cognome_persona_R> :pers_i1
"Masci"^^rdfs:Literal)
DataPropertyAssertion(<http://modeus.uniroma1.it/ontology#domicilio_persona_intestatario_R> :pers_i1
"Tivoli"^^rdfs:Literal)
DataPropertyAssertion(<http://modeus.uniroma1.it/ontology#nome_persona_R> :pers_i1
"Agostino"^^rdfs:Literal)
# Individual: :pers_i2 (:pers_i2)
ClassAssertion(<http://modeus.uniroma1.it/ontology#Persona_intestatario_R_padrone_diretto> :pers_i2)
ObjectPropertyAssertion(<http://modeus.uniroma1.it/ontology#est_persona_intestatario_R_padrone_diretto>
:pers_i2 :part5)
ObjectPropertyAssertion(<http://modeus.uniroma1.it/ontology#est_persona_intestatario_R_padrone_diretto>
:pers_i2 :part6)
ObjectPropertyAssertion(<http://modeus.uniroma1.it/ontology#est_persona_intestatario_R_utilista> :pers_i2
:part7)
DataPropertyAssertion(<http://modeus.uniroma1.it/ontology#cognome_persona_R> :pers_i2
"Poggi"^^rdfs:Literal)
DataPropertyAssertion(<http://modeus.uniroma1.it/ontology#nome_persona_R> :pers_i2
"Giuseppe"^^rdfs:Literal)
# Individual: :sez1 (:sez1)

ClassAssertion(<http://modeus.uniroma1.it/ontology#Sezione_catastale> :sez1)
ObjectPropertyAssertion(<http://modeus.uniroma1.it/ontology#sezione_catastale_contiene_rappresentazione_par
ticella> :sez1 :part1)
ObjectPropertyAssertion(<http://modeus.uniroma1.it/ontology#sezione_catastale_est_rappresentata_graficamente
> :sez1 :mappa1)
DataPropertyAssertion(<http://modeus.uniroma1.it/ontology#nome_sezione_catastale> :sez1 "Città di
Tivoli"^^rdfs:Literal)
DataPropertyAssertion(<http://modeus.uniroma1.it/ontology#numero_sezione_catastale> :sez1 "1"^^xsd:int)
# Individual: :voltura1 (:voltura1)
ClassAssertion(<http://modeus.uniroma1.it/ontology#Istanza_voltura> :voltura1)
```

DataPropertyAssertion(<http://modeus.uniroma1.it/ontology#anno_formazione_unità> :voltura1 "1836"^^rdfs:Literal)
DataPropertyAssertion(<http://modeus.uniroma1.it/ontology#giorno_formazione_unità> :voltura1 "2"^^xsd:int)
DataPropertyAssertion(<http://modeus.uniroma1.it/ontology#mese_formazione_unità> :voltura1 "aprile"^^rdfs:Literal)
DataPropertyAssertion(<http://modeus.uniroma1.it/ontology#numero_progressivo_istanza_voltura> :voltura1 "4171"^^xsd:int)
DataPropertyAssertion(<http://modeus.uniroma1.it/ontology#segnatura_unità_descrizione> :voltura1 "ASR, Cancelleria del censo di Tivoli, Istanze di Voltura, 3924"^^xsd:string)
)